%% file: main.tex
\documentclass[11pt,a4paper]{article}
\usepackage{mathtools,amssymb,amsthm,bm,mathtools,multirow,fullpage}
\usepackage{graphicx,color}
\usepackage{booktabs}
\usepackage[round,sort]{natbib}
\usepackage[hyphens]{url}

\usepackage[colorlinks=true,
linkcolor=blue,
urlcolor=blue,
citecolor=blue]{hyperref}
\usepackage[capitalize]{cleveref}
\newtheorem{theorem}{Theorem}
\newtheorem{corollary}{Corollary}
\newtheorem{proposition}{Proposition}
\newtheorem{lemma}{Lemma}
\newtheorem{assumption}{Assumption}
\newtheorem{Example}{Example}
\theoremstyle{definition}

\newtheorem{remark}{Remark}

\usepackage{graphicx,xcolor,float}
\input{jnwcommands}

\crefname{assumption}{Assumption}{Assumptions} 

\usepackage{algorithm}
\usepackage{algpseudocode}
\makeatletter
\def\BState{\State\hskip-\ALG@thistlm}
\makeatother

\newcommand{\wh}{\widehat}
\newcommand{\wt}{\widetilde}
\newcommand{\rI}{\mathrm{I}}
\newcommand{\rII}{\mathrm{II}}

\newcommand{\be}{\mathbf{e}}
\newcommand{\bI}{\mathbf{I}}

\newcommand{\rank}{\mathrm{rank}}
\newcommand{\diag}{\mathrm{diag}}
\newcommand{\T}{\top}

\newcommand{\op}{\mathrm{op}}
\newcommand{\Cov}{\mathrm{Cov}}

\def\bomega{{\bm \omega}}
\def\btheta{{\bm \theta}}
\def\balpha{{\bm \alpha}}

\def\cB{{\mathbb B}}
\def\bM{{\mathbf M}}

\def\oa{\overline{\alpha}} 
\def\ua{\underline{\alpha}}
\def\os{\overline{\sigma}} 
\def\us{\underline{\sigma}}
\def\vs{\varsigma}
\def\i{\infty}

\def\supp{{\rm supp}}

\def\bO{\mathbb{O}}
\def\bS{\mathbb{S}}
 
\def\bX{\mathbf{X}} 
\def\bmvec{\bm m}

\begin{document}

\title{The EM-algorithm and the Method of Moments in Softmax Mixture Models}

\author{
Xin Bing\thanks{Department of Statistical Sciences, University of Toronto}
\and
Florentina Bunea\thanks{Department of Statistics and Data Science, Cornell University}
\and
Jonathan Niles-Weed\thanks{Courant Institute of Mathematical Sciences and Center for Data Science, New York University}
\and
Marten Wegkamp\thanks{Department of Mathematics and Department of Statistics and Data Science, Cornell University}}
\date{\today}

\maketitle

\begin{abstract}
 Softmax Mixture Models (SMMs) are discrete $K$-mixture models for the probability of selecting one of $p$ candidate feature vectors $X_1,\ldots,X_p \in \mathbb{R}^L$ in heterogeneous populations. In addition to their classical use in econometrics and scientific applications, closely related softmax mixture mechanisms are key building blocks in modern LLM architectures. Despite their broad applicability, a focused theoretical and methodological study of SMMs is lacking. This paper takes a step in this direction by providing a comprehensive analysis of the Expectation-Maximization (EM) algorithm and the Method of Moments (MoM) for parameter estimation in SMMs.

We show that EM recovers the mixture atoms at the parametric rate, up to logarithmic factors, after only $\mathcal{O}(\log N)$ iterations, where $N$ is the sample size. This result holds when the separation between the SMM atoms is at least of order $\log K$, for any $K>1$. This condition improves upon those required by existing theoretical analyses of EM for high-dimensional mixture models, which, to date, have focused on Gaussian mixtures.

As a second contribution, we develop MoM procedures tailored to SMMs, based on a construction that, to our knowledge, has not been previously considered. These procedures can be used for both parameter and subspace estimation. Our theoretical analysis shows that MoM parameter estimates converge more slowly than EM estimates and that their finite-sample performance can deteriorate rapidly with $K$. Nevertheless, they can provably provide warm starts for EM and are recommended in practice for small $K$. For general $K$, we develop a MoM-based procedure for estimating the $K$-dimensional subspace spanned by the mixture atoms. This leads to our overall theoretical and practical recommendation for parameter estimation in SMMs: run EM from multiple random initializations within the estimated subspace. A simulation study and a data analysis provide further insight into the methods and theory.

Finally, we study the limiting behavior of SMMs as $p \to \infty$. We show that SMMs can be viewed as discrete approximations to $K$-mixtures of exponential tilts of the feature distribution, a connection that yields asymptotic identifiability of the model.
\end{abstract}

{\em Keywords:} Finite mixture models, method of moments, EM algorithm, parameter estimation, mixed multinomial logit models, latent class models, asymptotic identifiability, LLM, rates of convergence, latent space, mixing measure.

\newpage 

\tableofcontents

\bigskip

\section{Introduction}

\subsection{The softmax mixture model}\label{first-section}

``Softmax mixtures'' define a parametric class of discrete mixture models
$\pi \in \Delta^{p-1}$, where $\Delta^{p-1}$ denotes the probability simplex in $\RR^p$, supported on a known set of vectors $x_1,\ldots,x_p \in \RR^L$.
For a fixed and known $K$, let $\btheta_k\in\RR^L$, $k\in[K]:=\{1,\ldots,K\}$, denote the component parameters. Each mixture component
$A(\btheta_k):=A(\cdot;\btheta_k)$
is a probability vector in $\Delta^{p-1}$, supported on $x_1,\ldots,x_p$, and parametrized via the
\emph{softmax function} $\mathsf{softmax}:\RR^p\to\Delta^{p-1}$,
\begin{equation}\label{softmax}
A(x_j;\btheta_k)
=
\bigl[\mathsf{softmax}(x_1^\T\btheta_k,\ldots,x_p^\T\btheta_k)\bigr]_j
=
\frac{\exp(x_j^\T\btheta_k)}
{\sum_{i=1}^p \exp(x_i^\T\btheta_k)},
\end{equation}
for each $j\in[p]$. Let $\balpha:=(\alpha_1,\ldots,\alpha_K)^\T\in\Delta^{K-1}$ denote the vector of mixing weights, and write
$\bomega:=(\balpha,\btheta_1,\ldots,\btheta_K)$.
The \emph{softmax mixture model} is then defined by
\begin{equation}\label{mix}
\pi(y;\bomega)
:=
\sum_{k=1}^K \alpha_k A(y;\btheta_k),
\qquad
y\in\{x_1,\ldots,x_p\}.
\end{equation}
We use the acronym $K$-SMM to refer to a softmax mixture model of the form~\eqref{mix}, and refer to the $\btheta_k$'s as the atoms of this model.

Throughout this paper, our focus is on estimating the parameter
$\bomega^*:=(\balpha^*,\btheta_1^*,\ldots,\btheta_K^*)$
from a sample $Y_1,\ldots,Y_N$ drawn from
$\pi^*:=\pi(\cdot;\bomega^*)$.

When $K=1$, the softmax mixture model reduces to the conditional logit model, whose properties are well understood; see \cite{MF74} and the literature review in \cite[Chapter 9]{Agresti}. In contrast, the case $K>1$ has received comparatively little theoretical attention.

The goal of this work is to lay the foundation for a systematic study of softmax mixture models. To this end, we study, in the context of model~\eqref{mix}, two fundamental methods for mixture estimation: the expectation-maximization (EM) algorithm and the Method of Moments (MoM). After developing and analyzing each method, we propose a parameter estimation procedure for SMMs that combines their strengths.

Our methodological and theoretical interest in SMMs is motivated in part by their applications. SMMs are widely used in econometrics, while related softmax mechanisms are central to modern LLM architectures. We discuss two such applications below. The structure of the paper and our main contributions are presented in Section~\ref{structure}.
 
\paragraph{Basic discrete choice models.}
Softmax mixtures were introduced in the econometrics literature by \citet{boyd1980effect} and \citet{cardell1980measuring} under the name ``mixed multinomial logits'' to model the preferences of a heterogeneous population of consumers over a set of mutually exclusive goods. In this application, each vector $x_j\in\RR^L$ represents the attributes of good $j$, while the vector $\btheta$ represents a consumer's preferences for these attributes. The model posits that consumers act via \emph{random utility maximization}: a consumer chooses
\begin{equation*}
j^\star
=
\argmax_{j\in[p]}
\{x_j^\T\btheta+\epsilon_j\}.
\end{equation*}
Here $\epsilon_1,\ldots,\epsilon_p$ are independent stochastic terms that reflect idiosyncratic variation in the consumer's preferences. When $\epsilon_1,\ldots,\epsilon_p$ follow a Gumbel distribution, \citep[see, e.g.,][]{yellott1977relationship}
\begin{equation*}
\p{j^\star=j}
=
\frac{\exp(x_j^\T\btheta)}
{\sum_{i=1}^p \exp(x_i^\T\btheta)},
\qquad
j\in[p].
\end{equation*}
Thus, a consumer with preference vector $\btheta$ chooses among the observed goods $x_1,\ldots,x_p$ according to the probability vector $A(\btheta)$.

This is an appropriate model for the choices of a single consumer or, more generally, for a group of consumers with identical preferences. To model a population of consumers with heterogeneous preferences, \citet{boyd1980effect} and \citet{cardell1980measuring} proposed modeling the population as a mixture of consumers with different preference vectors. The resulting aggregate choice probabilities are then given by the softmax mixture model~\eqref{mix}.

This model has been broadly adopted throughout the management science and econometrics literature because of its flexibility and practicality; see \citep{mcfadden2000mixed,train2009discrete} and references therein. Section~\ref{sec_real_data} provides an application of our proposed estimation approach to transaction data.

\paragraph{Next-token prediction in LLMs.}
Modern LLMs generate open-ended text in response to a prompt one token at a time. Formally, the prompt is tokenized to yield $u_1,\ldots,u_m\in\RR^L$. This sequence is passed through a transformer-based model, following the architecture introduced by \citet{vaswani2017attention}, to produce contextually embedded vectors $z_1,\ldots,z_m$, from which a context-dependent representation $z\in\RR^L$ is obtained.

Given a vocabulary represented by vectors $x_1,\ldots,x_p\in\RR^L$, corresponding to the $p$ possible next tokens, the next token is drawn from a probability distribution on these $p$ tokens with respective masses
\begin{equation*}
A(x_j;z)
:=
\frac{\exp(x_j^\T z)}
{\sum_{i=1}^p \exp(x_i^\T z)},
\qquad
j\in[p].
\end{equation*}
Thus, conditional on a latent representation $Z=z$, the next-token distribution has the same form as~\eqref{softmax}.

This observation motivates the use of a softmax mixture as a stylized model for heterogeneity in the latent representations underlying next-token predictions. In particular, suppose that
\begin{equation*}
Z
\sim
\rho
:=
\sum_{k=1}^K \alpha_k\delta_{\btheta_k},
\end{equation*}
where $\btheta_1,\ldots,\btheta_K\in\RR^L$ represent a finite collection of latent directions associated with the generation of the context-dependent representation. Then, marginalizing over $Z$, the distribution of the next token $Y$ is
\begin{equation*}
\pi(y)
:=
\sum_{k=1}^K
\alpha_k
\frac{\exp(y^\T\btheta_k)}
{\sum_{j=1}^p \exp(x_j^\T\btheta_k)},
\qquad
y\in\{x_1,\ldots,x_p\},
\end{equation*}
which is precisely a softmax mixture of the form~\eqref{mix}. Estimating the atoms $\btheta_k$ and their respective weights can therefore provide a low-dimensional summary of heterogeneous latent directions associated with the LLM output distribution.

\subsection{Organization of the paper and main  contributions} \label{structure}

\subsubsection{The EM algorithm in softmax mixtures (Section \ref{EM}).}
The EM algorithm~\citep{dempster1977maximum} is commonly used to iteratively maximize the likelihood in settings where the MLE is intractable, and it has been shown to perform well across a wide range of applications.  The log-likelihood   $\ell_N(\bomega)$  under model (\ref{mix}), and given in   (\ref{llh_samp}),  is   non-concave in $\bomega = (\balpha, \btheta_1, \ldots, \btheta_K)$,  making  EM  a  natural first choice for parameter estimation.  

Section \ref{EM-algo}  gives the explicit form of the EM steps, which consist in iteratively updating $\balpha$ and $\btheta_1, \ldots, \btheta_K$,  by  maximizing a convex surrogate of $\ell_N$.

The theoretical analysis is given in Section \ref{EM-theory}. In contrast to the practical success and  popularity of the EM algorithm, its theoretical justification in a general context is scarce. 
It is often fairly easy to prove algorithmic convergence to a {\em local} optimum, but much harder  to guarantee that the limit is  a near {\em global} optimum of the sample likelihood. If the likelihood is unimodal, \cite{wu1983convergence} shows that the EM algorithm converges to the global optimum under certain regularity conditions. When the likelihood is multimodal, as is typically the case for mixture models, the theoretical understanding of the EM algorithm is largely limited to the settings of $K$-GMMs (Gaussian Mixture Models with $K$ components), and their variants. See, for instance, \cite{Xu_Hsu_Maleki,daskalakis2017ten,EM2017,CHIME2019,wu2021randomly} for $K=2$, and \cite{dasgupta2007probabilistic,yan2017convergence,Zhao2020,BingKongLi2026} for  $K\ge 2$.  To the best of our knowledge, a theoretical analysis of the EM algorithm for  $K$-SMMs has not yet been developed. As we elaborate in   \cref{rem:tricky1}  and \cref{examp_comp_2GMM} of \cref{app_sec_EM_fix}, establishing convergence to a global maximum in $K$-SMMs presents significantly greater challenges than in $K$-GMMs. We meet these challenges in 
Theorem \ref{thm_EM}, where  we give  both statistical and computational guarantees  for the EM output, when the algorithm is initialized within a neighborhood of size $\cO(1/\sqrt{L + \log p})$ of the true parameter. Theorem \ref{thm_EM} shows that EM  converges  to the true model parameters at a near-parametric rate after $\cO(\log N)$ iterations, and each iteration has computational complexity $\cO(pL)$. 
In addition to characterizing the contraction basin needed for initialization, our analysis reveals conditions on the separation  of the $K$-SMM atoms  under which EM  converges. Notably, and improving upon the sharpest known result, albeit developed only for $K$-GMMs,  we  require that the atom separation  depend  only logarithmically on the number of components and the smallest mixing weights.
Remark \ref{rem-sep}  of Section \ref{EM-theory} contains  further details on this point.



\subsubsection{The Method of Moments (MoM)  in softmax mixtures (Section \ref{sec_MoM}).}  
MoM is a classical estimation method for parametric mixture models, but the expression and properties of MoM parameter estimates in SMMs  have not been derived and studied.  It has been long known, in the classical literature on finite, parametric, mixture models \citep{Lindsay-book}, that the mixture parameters solve a specific system of equations involving a minimal set of  moments of the latent mixture \begin{equation}\label{def_rho}
\rho^* := \sum_{k = 1}^{K} \alpha_k^* \delta_{\btheta_k^*},
\end{equation} 
where $\delta$ denotes the Dirac measure on $\RR^L$. Henceforth,  we refer to any moments of $\rho^*$ as {\it latent moments}. \cref{Lindsay} in \cref{app_sec_MoM_latent} recalls  this classic result.  Furthermore, it is also known that these latent moments  equal appropriate functionals of the observable mixture distribution, henceforth called {\it observable moments}, for particular distribution classes, the so-called quadratic variance exponential families; see, for instance, \cite{Tuck63}   \cite{Bro77}, \cite{Lin89}  for earlier references,   and also  \cite{WuYan20} for Gaussian mixtures and  \cite{TiaKonVal17}, for binomial mixtures. It is noteworthy that these constructions are mostly known for univariate mixtures, $L = 1$. To the best of our knowledge, the  only exception  to the multivariate case, $L > 1$, is to  Gaussian mixtures \cite{Lindsay93,doss2023optimal}.
Relying on the equality between latent and  observable moments,    classical MoM parameter estimates in these mixture models solve the system of equations of \cref{Lindsay}, with estimated observable moments as input.


The situation is different in SMMs, which do not belong to the previously studied classes. In SMMs, it is not known how to equate  moments of $\rho^* $ with moments of $\pi^*$.
We show in \cref{randX}  of \cref{app_sec_mom_rem} that, in fact, there do not exist functionals of $\pi^*$  that can even recover the mean of $\rho^*$ along all deterministic sequences $\{x_1, \ldots, x_p\}$ of support points,   even when  $L = 1$. This underscores the main difficulty in using MoM under SMMs. 

    Fortunately, when the support points of model  (\ref{mix}) are random draws from a measure $\mu$  on $\RR^L$, we can construct functionals of $\pi^*$ that {\it approximate} latent moments of $\rho^*$, for sufficiently large $p$, and show  that  MoM-type estimation is possible in  SMMs. This is the content of  Section \ref{mom-system}. Proposition  
\ref{crux} shows how to construct moments of $\pi^*$  that approximate latent moments of $\rho^*$. Algorithm \ref{alg_MoM} solves an appropriate  system of equations with these observable moments as input, and Theorem \ref{prop:theta_gap} shows that its solution is close to the  true model parameter $\bomega^*$.
This validates the use of Algorithm \ref{alg_MoM}  for the  MoM parameter estimation procedure given   in Section \ref{MoM-estimation}. Theorem \ref{thm_mom_est} shows that MoM enjoys the usual parametric rate, up to logarithmic factors,  $\cO(\sqrt{L\log\log N/N})$, plus an error term  $\cO(\sqrt{L\log \log p/p})$ owed to the latent moment approximation step. 
Although MoM parameter estimates appear to have appealing properties, especially when $N$ and $p$ are large, the discussion following Theorems \ref{prop:theta_gap} and \ref{thm_mom_est} and  the detailed \cref{rem_rate} of \cref{app_sec_mom_rem} reveal its poor finite sample dependence on $K$. Combined with our simulation studies in \cref{sec_sims}, we do not recommend  parameter estimation via MoM when $K$ is large; empirical studies suggest caution when $K\geq 10$.  

 The method of moments can nevertheless be used reliably to estimate the $K$-dimensional subspace in $\RR^L$ spanned by the $K$ mixture atoms,  for any $K > 1$, as shown by Proposition \ref{prop_eigensp} of  \cref{latent-space}. The subspace estimation construction and properties are new in SMMs, and can be of independent  interest. 

\subsubsection{A synthesis of the EM and MoM results (Sections \ref{sec_synthesis} and \ref{sec_sims}).}

We combine  the strengths of EM and MoM  into  two overall estimation procedures. In Section \ref{sec_EM_MoM} we discuss EM with MoM initialization. Theorem \ref{thm_EM_MoM} shows that MoM verifies the initialization requirements of \cref{thm_EM}, and can therefore serve as a warm start. We recommend this procedure for small values of $K$, in light of MoM's fragility  for large $K$. 

In Section \ref{sec_EM_SS} we provide a procedure valid for all $K > 1$, which may nevertheless be more computationally demanding: EM with multiple random initializations in the estimated latent subspace. A common practical heuristic is to perform multiple random initializations and select the EM estimate that yields the highest likelihood \citep{dasgupta2007probabilistic}. However, this approach typically requires  $\cO(\exp(L))$ initializations to succeed, which quickly becomes computationally infeasible as $L$ increases. By sampling at random from the estimated  $K$-dimensional subspace spanned by the mixture atoms,  the number of  random initializations  required for   EM to succeed is reduced to $\cO(\exp(K))$; see \cref{lem_rand_init} of \cref{sec_EM_SS}. Theoretical guarantees of resulting EM procedure are given in Theorem \ref{thm_EM_SS}.  
MoM and  the EM algorithm with either of these two initialization schemes are compared numerically  in Section \ref{sec_sims}, supporting our theoretical results, which also favor EM with random subspace initialization. 

In Section \ref{sec_mom_ident} we draw concluding remarks on SMMs. Proposition \ref{prop:weak} reveals an interesting  feature of this model: discrete SMMs can be viewed as random discretizations of continuous mixtures of exponential tilts of a measure $\mu$ on $\RR^L$. When $\mu$ admits a continuous density, these mixtures are identifiable, which is  a crucial step in proving, in Theorem \ref{prop:iden},  that SMMs are  identifiable, when the number of support points $p$ is large. Further results  and open problems related to identifiability are given in Remark \ref{rem-p0}.  

The main assumptions used throughout the paper are collected in \cref{assumptions}. The proofs of all the results stated in the paper are given  in the supplement,   which also contains additional theoretical  results and simulation studies as well as a real data analysis.

\subsection{Assumptions}\label{assumptions} 
The following assumptions will be used throughout the paper. The first is a generic assumption ensuring that the model is a non-degenerate \(K\)-component mixture.
\begin{assumption}[$K$-mixture]\label{ass_alpha}
 There exist constants  $\ua,\oa \in (0,1)$ such that,
  \begin{equation}\label{def_ua_oa}
        \ua \le \min_{k\in [K]} \alpha_k^* \le \max_{k\in [K]} \alpha_k^* \le \oa.
     \end{equation}
     \end{assumption} 
The second assumption  is also generic, ensuring that the mixture components are distinct. 
\begin{assumption}[Distinct atoms]\label{separation}
There exists  $\Delta >0$ such that,
\begin{equation}\label{cond_sep}
          \min_{k\ne k'}\|\btheta_k^*-\btheta_{k'}^*\|_2  ~ \ge~  \Delta.
       \end{equation}
        \end{assumption}
The third assumption controls the scale of each atom. 
\begin{assumption}[Bounded atoms]\label{ass_theta}
        There exists some known constant $B<\i$ such that   
        $\|\btheta_k^*\|_2 \le B$ for all $k\in [K]$.
	\end{assumption}
Together with the  condition on $x_1,\ldots,x_p$ introduced below, \cref{ass_theta} ensures that the softmax scores $x_j^\T\btheta_k^*$ remain well controlled. This prevents the softmax probabilities in $A(\btheta_k^*)$ from being spiky, and is important for the softmax parametrization to remain meaningful; see \cite{arora2016latent}. 


The following is a baseline assumption on the $p$ support points of the mixture.
\begin{assumption}[Random design]\label{random-points}
  The support points $x_1, \ldots, x_p$ of the softmax mixture \eqref{mix} are realizations of $X_1, \ldots, X_p$, which are i.i.d. from a centered, sub-Gaussian probability measure   $\mu$ on $\RR^L$ with some finite sub-Gaussian constant $\os_\mu$.\footnote{A centered random vector $X\sim \mu$ is   $\sigma$ sub-Gaussian if $\EE_\mu[\exp(v^\T X)] \le \exp(\|v\|_2^2\sigma^2/2)$ for all $v\in \RR^L$.}
\end{assumption} 

The zero mean assumption is imposed for identifiability reasons: since the expression $\exp(x_j^\T \btheta_k^* + c) / \sum_{\ell=1}^p\exp(x_\ell^\T \btheta_k^* + c)$ remains the same for all $c\in \RR$, \cref{random-points} fixes this shift by setting $c = -\EE_\mu[X]^\T \btheta_k^*$. We therefore assume the columns of $\bX = (x_1^\T,\ldots,x_p^\T)^\T$ have been centered.
The sub-Gaussianity condition is used to characterize the convergence rates of the studied estimates in terms of parameters such as $p$ and $L$, and could potentially be relaxed to suitable higher-moment bounds and conditions on the existence of moment generating functions.

\section{The EM algorithm in  softmax mixtures}\label{EM}

In Section  \ref{EM-algo} we give  the description of the EM algorithm for softmax mixtures, and we analyze it theoretically in Section \ref{EM-theory}. 
Looking ahead,  the results  of this section will be complemented with an algorithm initialization construction in  Sections \ref{sec_EM_MoM} and \ref{sec_EM_SS}.

\subsection{The EM updates}\label{EM-algo}

Given  a sample $Y_1,\ldots, Y_N$ i.i.d. $\pi^*$, let 
$$
    \wh \pi_j := {1\over N}\sum_{i = 1}^{N} 1\{Y_i = x_j\},\qquad j \in [p],
$$ 
denote the observed frequency of a feature $x_j$ in the sample, for $j \in [p]$. Letting $\bomega= (\bm \alpha, \btheta_1, \ldots, \btheta_K)$ collect  all model parameters, the log-likelihood of this sample  is 
\begin{align}\label{llh_samp} \ell_N(\bomega) &=
    {1\over N}\sum_{i=1}^N \log \left(
    \sum_{k=1}^K \alpha_k  A(Y_i; \btheta_k)
    \right)  =
     \sum_{i=1}^p \wh \pi_i \log \left(
    \sum_{k=1}^K \alpha_k {\exp(x_i^\T \btheta_k) \over
\sum_{j=1}^p \exp(x_j^\T \btheta_k)}
    \right), 
\end{align}   
which is not concave in $\bomega$. Instead of maximizing $\ell_N(\bomega)$ directly, the EM algorithm optimizes the proxy introduced below. 
Let $Z$ be the random vector taking values in the set $\{\btheta_1, \cdots, \btheta_K\}$ with corresponding probabilities  $\balpha \in \Delta^{K-1}$. We define the conditional probability of $Z = \btheta_k$ given $Y = x_j$, for any $k\in [K]$ and $j\in [p]$, as
	\begin{equation}\label{distr_Z_mid_X}
		g(\btheta_k \mid x_j; \bomega) :=  {\alpha_k   A(x_j; \btheta_k) \over
  \pi(x_j; \bomega)} =
  {\alpha_k   A(x_j;\btheta_k) \over \sum_{a=1}^K \alpha_a   A(x_j; \btheta_a)}.
	\end{equation}   
Second, we define the joint probability of $Z = \btheta_k$ and $Y = x_j$ as
	\begin{align}\label{X,Z} 
	\log f(x_j, \btheta_ k; \bomega) &:=  
         \log(\alpha_k) +   x_j^\T \btheta_k  - \log\left(
        	\sum_{\ell=1}^p \exp\left({x_{\ell}^\T \btheta_k}\right)
        	\right).
	    	\end{align}
	 The EM algorithm iteratively  maximizes the  $\wh Q$-function  
\begin{align}\label{def_Q_popu}
\wh Q(\bomega \mid \bomega') 
& =   \sum_{j=1}^p \wh \pi_j \sum_{k=1}^K g(\btheta_k' \mid x_j; \bomega')  \log f(x_j, \btheta_k; \bomega) \nonumber \\ 
            & = \sum_{j=1}^p \wh \pi_j\sum_{k=1}^K {\alpha_k'A(x_j;\btheta_k') \over \pi(x_j;\bomega')}    \left[
			\log ( \alpha_k) + x_j^\T \btheta_k  - \log\left(
			\sum_{\ell=1}^p \exp\left({x_{\ell}^\T \btheta_k}\right)
			\right)
			\right]
\end{align}   over its first argument $\bomega$. 

   In the parlance of the EM algorithm literature, evaluating the $\wh Q(\bomega \mid \bomega')$ function at a given $\bomega'$ corresponds to the ``E-step'', while maximizing over $\bomega$ is the ``M-step''. Starting at some initial point $ \wh \bomega^{(0)}$, the classical 
   EM algorithm iterates as follows:  
	\begin{equation}\label{EM_iter}
 		 \wh \bomega^{(t+1)} = \argmax_{\bomega} ~  \wh Q(\bomega \mid   \wh \bomega^{(t)})\qquad \text{for} ~ t = 0, 1, 2, \ldots
	\end{equation}
	 until convergence, with $ \wh \bomega^{(t)} =: 
     (\wh {\bm \alpha}^{(t)}, \wh \btheta_{1}^{(t)}, \ldots, \wh \btheta_{K}^{(t)})$.
     
     The maximization in \eqref{EM_iter} is a concave optimization problem. Specifically,  maximizing with respect to $\balpha \in \Delta^{K-1}$ yields the closed-form solution
    	\begin{align}\label{iter_alpha}
    	 \wh \alpha_k^{(t+1)}  
      &=\sum_{j=1}^p \wh \pi_j  
      {\wh \alpha_k^{(t)}   A(x_j;  \wh \btheta_k^{(t)}) \over 
      \pi (x_j;  \wh \bomega^{(t)})}
      :=  \wh M_k( \wh \bomega^{(t)})
    	\end{align}
    for $k\in [K]$. 
    Maximization over $\btheta_1, \ldots, \btheta_K$ does not admit a closed-form solution, and we adopt a  gradient-ascent step,
         which is often  used in such circumstances. For all $ k\in [K]$, let  
         $ \nabla_{\btheta_k} \wh Q(\bomega \mid \wh \bomega^{(t)}) $ be the gradient
         of $ \wh  Q(\bomega \mid \wh \bomega^{(t)})$ with respect to $\btheta_k$ in the {\em first argument} $\bomega$.
    Given a chosen step size $\eta_k>0$,  the M-step update for maximizing over $\btheta_k$ is given by
    	\begin{align}\nonumber
    		\wh \btheta_k^{(t+1)} &=  \wh \btheta_k^{(t)} +  \eta_k ~  \nabla_{\btheta_k}  \wh Q(\bomega \mid      \wh \bomega^{(t)})   \mid_{\bomega =  \wh \bomega^{(t)}}  \\\label{iter_theta} 
            &=  \wh \btheta_k^{(t)} +  \eta_k ~  \sum_{j=1}^p  \wh \pi_j { \wh \alpha_k^{(t)} A (x_j; \wh \btheta_k^{(t)})\over \pi(x_j; \wh \bomega^{(t)})} \left( x_j - \bX ^\T A( \wh \btheta_k^{(t)})\right). 
    \end{align}  
   Thus, for all $t\ge 0$ and $k \in [K]$, the  EM updates are given by (\ref{iter_alpha}) and (\ref{iter_theta}). Since the update in (\ref{iter_alpha}) is given in closed form, whereas (\ref{iter_theta}) involves a gradient ascent step, the  EM algorithm above for softmax mixtures is viewed as a {\em hybrid} procedure.
\subsection{An analysis of the EM algorithm and rates of convergence}\label{EM-theory}
  In addition to  Assumptions \ref{ass_alpha} - \ref{random-points}, the results of this section hold under the following condition on a matrix defined below.  For any $\btheta \in \RR^L$ with $A(\btheta)\in \Delta^{p-1}$,  note that $H_{\btheta} := \bX^\T (\diag(A(\btheta)) - A(\btheta)A(\btheta)^\T) \bX$ is the $L\times L$ Fisher information matrix of  one observation $Y \sim  A(\btheta)$.  For $X_1,\ldots, X_p$ being i.i.d. copies of $X \sim \mu$,  the analogue of $H_{\btheta}$ is 
  \begin{equation}\label{def_H_theta_mu}
    H_{\btheta}^{(\mu)} := {\EE_\mu[XX^\T \exp(X^\T\btheta)] \over \EE_\mu[\exp(X^\T\btheta)]} - {\EE_\mu[X\exp(X^\T\btheta)]  \EE_\mu[X\exp(X^\T\btheta)]^\T \over (\EE_\mu[\exp(X^\T\btheta)])^2}.
 \end{equation} 
  We require some regularity conditions of this matrix  along the line segments connecting pairs of the true atoms $\btheta_1^*,\ldots,\btheta_K^*$. For  any symmetric, positive semidefinite matrix $M \in \RR^{d \times d}$, denote its  eigenvalues by $\lambda_1(M) \ge \cdots \ge \lambda_d(M)$.

\begin{assumption}\label{ass_Htheta_mu}     
For all $\btheta = u \btheta_a^* + (1-u)\btheta_b^*$ with $u\in [0,1]$ and $a,b\in [K]$,  assume that
$ 0<\us_\mu^2\le \lambda_L(H_{\btheta}^{(\mu)})\le \lambda_1(H_{\btheta}^{(\mu)}) \le \os_\mu^2<\i$.
\end{assumption} 

\cref{ass_Htheta_mu} depends on the measure  $\mu$ satisfying \cref{random-points}, but can be verified, for instance,  when $\mu=\cN(0,\Sigma)$, with $\us_\mu^2=\lambda_L(\Sigma)$ and $\os_\mu^2=\lambda_1(\Sigma)$, as shown in \cref{lem_hess} in \cref{sec:concentration-H-Gausz}.

    The following theorem presents the convergence rate of the  EM updates with respect to the distance between  any  $\bomega$ and $\bomega'$  defined as  
    	\begin{equation}\label{def_dist}
    		d(\bomega, \bomega') = \max\left\{ \os_\mu \max_{k\in [K]}\|\btheta_k - \btheta_k'\|_2 , ~  {1\over \ua} \max_{k\in[K]} | \alpha_k - \alpha_k' | \right\}
    	\end{equation}
    with $\ua$ and $\os_\mu$ given in \cref{ass_alpha,ass_Htheta_mu}, respectively.   Its proof is given in \cref{app_proof_EM_random}.
\begin{theorem}\label{thm_EM} 
Grant Assumptions \ref{ass_alpha} - \ref{ass_Htheta_mu},  
with $\Delta$ additionally satisfying  
\begin{equation}\label{cond_sep_rand}
           \Delta  \ge  \frac{C}{\us_\mu^2}\left\{
                \log K + \log {\os_\mu^2\over \us_\mu^2} + \log {\oa\over \ua}
            \right\}
        \end{equation}
for some absolute constant $C > 0$.
  Assume  that the initialization  $\wh \bomega^{(0)} $ satisfies 
 \begin{align}\label{cond_init_rand}
          d( \wh \bomega^{(0)} ,  \bomega^*)\le \delta_\mu \qquad\text{ with }\quad  \delta_\mu \le  {c_\mu\over \sqrt{L+\log p}} 
\end{align} 
with some constant $c_\mu$ depending  on $\os_\mu^2,\us_\mu^2$ and $B$ only.
Further assume that there exists some constant $C'>1$ such that 
     \begin{equation}\label{cond_N_rand}
         {\ua N\over \log N} \ge C'~  {\oa   \over \ua}{\os_\mu^4\over \us_\mu^4}  (L+\log p) ~ KL.
     \end{equation}
   
    Then, there exist some $0<\phi <1$  and step-sizes $\eta_k>0$, $k\in [K]$, such that with probability  $1-N^{-2} - p^{-2}$, the following holds for the EM iterates $ \wh\bomega^{(t)}$ in (\ref{iter_alpha}) and (\ref{iter_theta}), with  $t\ge 0$,
    \begin{equation}\label{rate_EM_final}
        d(\wh\bomega^{(t)}, \bomega^*) \le ~  \phi^t ~ \delta_\mu +  C''\sqrt{\oa K L \log N \over \ua^2 N}.
    \end{equation}
\end{theorem}

On the right-hand side of \eqref{rate_EM_final}, the first term is the algorithmic error, which vanishes exponentially as $t\to \i$, while the second term is the statistical error, which can be achieved within at most $\cO(\log N)$ iterations. \cref{thm_EM} further characterizes the explicit dependence of the statistical rate on $K$, $\ua$, $\oa$, and $L$.  In the case of balanced mixing probabilities, where $\oa \asymp \ua$, the rate simplifies to $\sqrt{KL\log(N)/(\ua N)}$, where $\ua N$ represents the smallest effective sample size across all mixture components. In light of this, condition \eqref{cond_N_rand} imposes a lower bound on this smallest sample size and is required for the convergence rate to vanish asymptotically. For fixed $K$ as considered in this paper, the rate further simplifies to $\sqrt{L\log(N)/N}$, which differs from the parametric rate for estimating an $L$-dimensional vector from $N$ i.i.d. samples by only a logarithmic factor.  Since the updates of $\wh\balpha^{(t)}$ in (\ref{iter_alpha}) also depend on $\wh\btheta_k^{(t)}$, the convergence rate of $\|\wh\balpha^{(t)} - \balpha^*\|_\infty := \max_k |\wh\alpha^{(t)}_k - \alpha_k^*|$ is primarily determined by the rate of $\|\wh\btheta_k^{(t)} - \btheta_k^*\|_2$. If one is interested in obtaining refined rates for estimating $\balpha^*$, a natural approach is to refit by maximizing the likelihood in (\ref{llh_samp}) over $\balpha$, with $\btheta_k$ replaced by $\wh\btheta_k^{(t)}$, and then appeal to the analysis in \cite{bing2022likelihood}.

    \begin{remark}[Separation]\label{rem-sep}
        Convergence of the EM iterates requires the separation condition in \eqref{cond_sep_rand} between the mixture components. Our analysis explicitly captures the dependence $\Delta$ on the number of mixture components $K$, the condition number $\os^2/\us^2$ of the information matrix, and the balance ratio $\ua/\oa$ of the mixing probabilities. When any of these quantities are large, the required separation increases only logarithmically. As illustrated in \cref{rem:tricky1} and \cref{examp_comp_2GMM} of \cref{app_sec_EM_fix}, deriving such a mild separation requirement under softmax mixtures is highly non-trivial and presents significantly greater challenges than in the case of Gaussian mixture models. Even for Gaussian location mixtures on $\RR^L$ with $K \ge 3$ components, the weakest known separation in terms of the squared Euclidean distances between mean vectors required for the EM algorithm to succeed is of the order of $L \wedge K$ \citep{yan2017convergence, Zhao2020}, whereas for Lloyd’s algorithm, it is of order $K / \ua$ \citep{lu2016statistical}.
    \end{remark}
     
     \begin{remark}[Initialization]\label{rem_init}
        It is well known that the EM algorithm is very sensitive to its starting value  $\wh \bomega^{(0)}$.  Our analysis specifies the initialization requirement under softmax mixtures, as given in \eqref{cond_init_rand}, and quantifies its  dependence on $L$ and $p$. The explicit dependence of $c_\mu$ on $\os_\mu^2, \us_\mu^2$ and $B$ is given in \eqref{def_c_mu} of \cref{app_proof_EM_random}.  
      One of our main contributions is to  construct a warm start $\wh\bomega^{(0)}$ that meets such requirement under suitable conditions. We meet this challenge in 
       \cref{MoM-estimation}.
     \end{remark}

    \begin{remark}[Effect of the step size]\label{rem_stepsize}
       Our theory also reveals that the step size $\eta_k$ cannot be chosen too large,  in order to ensure algorithmic convergence. On the other hand, choosing a smaller $\eta_k$ results in a slower convergence rate ($\phi$ gets closer to 1), but does not affect the final statistical accuracy of the EM updates, as shown in \cref{app_sec_EM_fix}.  We found that the choice of $\eta_k = 0.2$ yields overall satisfactory results in our numerical experiments.
    \end{remark}

\begin{remark} 
[Generic design points]\label{gen} \cref{app_sec_EM} collects theoretical results on the EM algorithm that are valid for any deterministic support points $\{x_1, \ldots, x_p\}$ of the softmax mixture  model (\ref{mix}). Specifically, Theorem \ref{thm_EM_fix}  is the counterpart of Theorem \ref{thm_EM} in this regime. It holds under similar assumptions, with Assumption \ref{ass_Htheta_mu}  replaced by its fixed design version. 
An interesting implication of a population-level counterpart  
is a local identifiability result, Corollary \ref{locid}, stated  in \cref{app_sec_EM_popu}. It shows that if there exist  two observationally equivalent parameters  $\bomega_1^*$ and $\bomega_2^*$ such  that   $\pi^* = \pi(\bomega_1^*) = \pi(\bomega^*_2)$, 
and if they are $\delta_0 $ apart, for $\delta_0 $ depending on the design,  with  explicit expression given in \eqref{cond_init_fix},  then the parameters  must coincide,  $\bomega_1^* = \bomega_2^*$. This result comes with the additional requirement that both  $\bomega_1^*$ and $\bomega_2^*$ satisfy the separation condition on $\Delta$ given by the analogue of \eqref{cond_sep}.  Nevertheless, the merit of this partial, local,  result pertaining to  softmax mixture models  with well separated atoms is that it holds for all $p \geq 1$ and any support point configuration. 
  We opted to include a comment on it here for contrast with a stronger, global identifiability result for softmax mixtures with random support, which holds for any atom separation $\Delta>0$, albeit for large values of $p$.  We give the latter  result, along with a discussion, in \cref{sec_mom_ident}.
  \end{remark}

\section{The Method of Moments in softmax mixtures}\label{sec_MoM}

\subsection{Parameter recovery from observable moments}\label{mom-system}
In Section \ref{structure} we outlined the principles underlying MoM in SMMs. We make them formal in this section.   We begin by introducing a  collection of latent moments of interest. Recall that the $K$-atomic, latent,  mixing measure is given by  $\rho^* := \sum_{k=1}^{K} \alpha_k^*\delta_{\btheta_k^*}$. Let $Z  \sim \rho^* $ be a discrete random vector in $\RR^L$. Its first coordinate $Z_1$ has the first $2K-1$ moments given by: 
\begin{equation}\label{mom} 
   m_r 
   :=  \EE_{\rho^*}[Z_1^r]  =   \sum_{k=1}^{K}\alpha_k^* (\theta_{1k}^{*})^r, \quad \quad \mbox{for \  $0 \leq r \leq 2K - 1$}. 
\end{equation}  
We also  consider the following mixed-moments: 
\begin{equation} \label{cross-mom} 
    m_{r1;i} := 
    \EE_{\rho^*}\left[ Z_1^rZ_i\right] =  \sum_{k=1}^{K}\alpha_k^* (\theta_{1k}^*)^r\theta_{ik}^*, \quad \quad \mbox{for \ $0 \leq r \leq K - 1$ and \ $2 \leq i \leq L$. }
\end{equation} 
The subscripts $r$ and $r1;i$ of $m$ are mnemonic of the fact that we consider either the $r$-th moment of $Z_1$, or moments of the product of $Z_1^rZ_i^1$. Let  $
    \bmvec  := (m_0, m_1, \dots, m_{2K-1})^\T$ and  $\bmvec_{1;i}:= (m_{01;i}, \ldots, m_{(K-1)1;i})^\T$ for  $2 \leq i \leq L.$
Equalities in \eqref{mom} and \eqref{cross-mom} form systems of equations that could be solved for the mixture parameters, given  $m_r$ and $m_{r1;i}$. We construct below observable functionals of $\pi^*$ that are  approximately equal to  $m_r$ and $m_{r1;i}$, $\mu$-almost surely. Such a construction requires knowledge of $\mu$ and additional smoothness of its density. We therefore strengthen the regularity conditions on $\mu$ imposed in \cref{random-points}.   
\begin{assumption}\label{ass:mu}
    Assume that $\mu$ is absolutely continuous with respect to the Lebesgue measure on $\RR^L$  
    and has a finite moment generating function everywhere. Its density $f_\mu$ is strictly positive,  and $ f_{\mu} \in C^{2K-1}(\RR^L)$, with mixed partial derivatives up to order $2K-1$ decaying super-exponentially at infinity.
\end{assumption}

Let $\be_1,\ldots,\be_L$ be the canonical vectors in $\RR^L$.
For any $r\in\NN$ and  $i \in \{2, \ldots, L\}$,  define
    \begin{align}\label{eq:hr_def}
    	h_r(x) & ~ :=~  h_r(x;\be_1)  ~ :=~  (-1)^r {1\over f_\mu(x)}  ~ \frac{\rd^r}{\rd t^r} f_\mu(x+t \be_1) \big|_{t=0},\\\label{eq:hr1_def}
    	h_{r1;i}(x)  & ~:=~  h_{r1;i}(x; \be_1,\be_i)  ~ :=~   (-1)^{r+1} {1\over f_\mu(x)}  ~ \frac{\rd^{r+1}}{\rd t^r \rd s} f_\mu(x+t\be_1 + s\be_i) \big|_{t,s = 0},
\end{align} 
for all $x\in \RR^L$.
The following proposition, proved in \cref{app_proof_crux}, shows that the above functions yield approximations to the latent moments in \eqref{mom} and \eqref{cross-mom}.

\begin{proposition}\label{crux}
    Let $X \sim \mu$ with $\mu$ satisfying \cref{ass:mu}. 
   For any given $\btheta \in \RR^L$,  any $r\in\NN$ and  $i \in \{2, \ldots, L\}$, we have:
    \begin{align}  
    \label{req_h_hi}
    {\EE_\mu[   h_r(X) \exp(X^\T \btheta)] \over \EE_\mu[ \exp(X^\T \btheta) ]}= \theta_1^r, \qquad   {\EE_\mu[ h_{r1;i}(X) \exp(X^\T \btheta)] \over \EE_\mu[ \exp(X^\T \btheta) ]}=  \theta_1^r \theta_i.
\end{align} 
Consequently, the observable moments (left hand side), converge to the corresponding latent moments (right hand side) $\mu$-almost surely as $p\to\infty$:
\begin{align}\label{cond-mom}
\bar{m}_r :=& ~ \EE_{\pi^*}[h_r(Y)] \to m_r,
\qquad
\bar{m}_{r1;i} := \EE_{\pi^*}[h_{r1;i}(Y)] \to m_{r1;i},
\end{align}

\end{proposition}  
To the best our knowledge,   the construction of $h_r$ and $h_{r1;i}$ is new for softmax mixtures.  We elaborate on its rationale, and give the  proof of Proposition \ref{crux} in \cref{app_proof_crux}.

\begin{Example}[Explicit choices of $h_r$]\label{examp_h}
When $\mu = \cN_L(a, \Sigma)$,  the functions given by (\ref{eq:hr_def}) or (\ref{eq:hr1_def}) take more familiar forms, and can be expressed in terms of the classical probabilist's Hermite polynomials $\{H_r\}_{r \geq 0}$, defined by
   \begin{equation}\label{def_Hermite_Poly}
       H_r(x) = r! \sum_{b=0}^{\lfloor r/2 \rfloor} \frac{(-1)^b}{b! (r - 2b)!2^b}x^{r - 2b},\qquad \forall ~ x\in\RR.
   \end{equation}
Specifically, one has 
    $
 		h_r(x) := h_r(x; a, \Sigma) =  \|\Sigma^{-1/2} \be_1\|_2^r H_r((x - a)^\T \Sigma^{-1}  \be_1/\|\Sigma^{-1/2}  \be_1\|_2). $
When $\mu = \sum_{j=1}^J \lambda_j \cN_L(a_j, \Sigma_j)$ is a finite Gaussian mixture   with $f_j$ being the density of $\cN_L(a_j, \Sigma_j)$, we have 
	$
		h_r(x) =   \sum_{j=1}^J \lambda_j f_j(x) ~ h_r(x; a_j,\Sigma_j) / (\sum_{j=1}^J\lambda_j f_j(x)).  
	$ 
\end{Example}

 With these ingredients, we can recover the model parameters by solving the system of equations given below. It is a variant of    \cref{Lindsay} in \cref{app_sec_MoM_latent}
 that replaces  the latent moments therein by their observable approximations. We first define the degree $K$ polynomial $\wt P(x)$, in one variable $x \in \RR$, by 
\begin{equation}\label{firstcoordinate}
		\wt P(x) :=  \mbox{\rm det}  \begin{pmatrix}
			1 &  \wt m_1 & \dots &  \wt m_{K} \\
			\wt m_1 &  \wt m_2 & \dots &  \wt m_{K+1} \\
			\vdots & \vdots & & \vdots \\
			\wt m_{K-1} &  \wt m_K & \dots &  \wt m_{2K-1} \\
			1 & x & \dots & x^K
		\end{pmatrix}.
	\end{equation}
The vector $\wt \bmvec := (\wt m_1, \ldots, \wt m_{2K - 1})$ is given by 
\begin{equation}\label{def_dmm} 
       \wt\bmvec 
       = \argmin_{ {\bm u}\in \cM} \| {\bm u}  - \bar \bmvec \|_2,
\end{equation}
where $\bar\bmvec := ( \bar m_1, \ldots, \bar m_{2K - 1})$ with $\bar m_r$ given in \eqref{cond-mom}, and  $      \cM := \{
             (M_1(\nu), \dots, M_{2K-1}(\nu)):   \supp(\nu) \in [-B, B]
        \} 
$ is the set of first $(2K-1)$ moments of a univariate probability measure $\nu$ supported within $[-B, B]$.  Since the entries of $\bar \bmvec$ may not correspond to moments of any bona fide measure, this projection step guarantees that the entries of $\wt \bmvec$ satisfy standard moment inequalities, which in turn guarantees,  that  $\wt P$ has $K$  roots, denoted by $\theta_{11}, \ldots, \theta_{1K}$. The moment projection step employs the idea originally proposed by \citet{WuYan20}. As observed therein, the optimization problem in~\eqref{def_dmm} can be formulated as a semidefinite program and hence solved in polynomial time~\citep{VanBoy96}.

 Then, for all $2\le i\le L$ and $k\in [K]$, we define the preliminary parameter $\bar\theta_{ik}'$ by
 \begin{equation}\label{rest-coordinates-empirical}
	\bar \theta_{ik}' = \begin{pmatrix}
		\bar m_{01; i} \\ \vdots \\  \bar m_{(K-1)1; i}
	\end{pmatrix}^\top   \begin{pmatrix}
		1 &  \wt m_1 & \dots &  \wt m_{K-1} \\
		\wt m_1 &  \wt m_2 & \dots &  \wt m_{K} \\
		\vdots & \vdots & & \vdots \\
		\wt m_{K-1} &  \wt m_K & \dots &  \wt m_{2K-2} 
	\end{pmatrix}^{\dagger}    \begin{pmatrix}
		1 \\  \bar \theta_{1k} \\ \vdots \\  (\bar \theta_{1k})^{K-1}
	\end{pmatrix},
\end{equation}
with $M^\dagger$ being the pseudo-inverse of any matrix $M$.
Then, 
we define $\bar \theta_{ik}$ to be the projection of $\bar \theta_{ik}'$ onto $[-B, B]$.
Finally,  the recovered mixture weights are given by 
\begin{equation}\label{weights-empirical}
	\bar {\bm \alpha}  = \Pi_{\Delta^{K-1}}\left\{\begin{pmatrix}
		1 & 1 & \ldots & 1 \\
		\bar \theta_{11}& \bar \theta_{12}&  \ldots & \bar \theta_{1K} \\
		\vdots & \vdots & & \vdots \\
		(\bar \theta_{11})^{K-1}& (\bar \theta_{12})^{K-1}&  \ldots & (\bar \theta_{1K})^{K-1} 
	\end{pmatrix}^{\dagger}\begin{pmatrix} 1 \\ \wt m_1 \\ \vdots \\ \wt m_{K -1}\end{pmatrix}\right\}\,,
\end{equation}
where $\Pi_{\Delta^{K-1}}$ is the projection operator to  $\Delta^{K-1}$. The entire procedure is summarized below.

\begin{algorithm}[H]
	\caption{Parameter Recovery via Observable Moments}\label{alg_MoM}
	\begin{algorithmic}[1]
		\Require  The moment vectors $\bar\bmvec \in \RR^{2K}$, $\bar\bmvec_{1;2},\ldots, \bar\bmvec_{1;L}\in \RR^{K}$ and a constant $B>0$.
		\Procedure{\textsc{MoM}}{$\bar\bmvec$, $\{\bar\bmvec_{1;i}\}_{2\le i\le L}$, $B$}
		\State Compute the projected moment vector $\wt \bmvec$ as in \eqref{def_dmm}.
		\State Solve the $K$ roots $\bar \theta_{11}, \ldots, \bar \theta_{1K}$ from $\wt P(x) = 0$.
		\State Solve $\bar \theta_{ik}$ for $i\in \{2,\ldots, L\}$ and $k\in [K]$ by projecting \eqref{rest-coordinates-empirical} to $[-B, B]$.
		\State 
		Solve for the weights $\bar \balpha$ from \eqref{weights-empirical}.
		\State \Return The mixing weights $\bar \balpha \in \Delta^{K-1}$ and the vectors $\bar \btheta_1, \ldots, \bar\btheta_K \in \RR^L$.
		\EndProcedure
	\end{algorithmic}
\end{algorithm}
  
We next connect the solution of \cref{alg_MoM} to the model parameters. We first recall Assumption \ref{separation}. It is immediate to show that if the vectors $\btheta_1^*, \ldots, \btheta_K^*$ in $\RR^L$ are distinct, than there exists a direction $v \in \RR^L$ such that their projections onto this direction are distinct. We provide an in-depth discussion on  the practical choice of $v$ in \cref{sec_mm_proj}. For ease of exposition we assume in this section that $v = \be_1$, the first canonical vector in $\RR^L$, and let $\Delta_1 > 0$ be a lower bound on the atom separation in their first coordinate 
\begin{equation}\label{cond_Delta1}
    \min_{k\ne k'} |\theta_{1k}^*-\theta_{1k'}^*| \geq \Delta_1.
\end{equation} 
Below we give a rate of convergence that strengthens the $\mu$-almost sure convergence in \cref{crux}. It requires finite third moments of $h_r(X)$ and $h_{r1;i}(X)$ for $X\sim\mu$.

\begin{assumption}\label{ass:mu_moment} 
    For the functions $h_r$ and $h_{r1;i}$ in \eqref{eq:hr_def} and \eqref{eq:hr1_def}, whenever they exist, assume that $\EE_\mu[|h_r(X)|^3]<\i$ for all $r\le 2K-1$ and that $\EE_\mu[|h_{r1;i}(X)|^3]<\i$ for all $r\le K-1$ and $2\le i\le L$.
\end{assumption}
Our analysis uses the law of the iterated logarithm, which requires having a finite second moment for $
 h_r(X)\exp(X^\T\btheta_k^*),$
and analogously for $h_{r1;i}$. By the sub-Gaussianity of $\mu$ and Hölder's inequality, the third-moment requirement in \cref{ass:mu_moment} can in fact be relaxed to the existence of any $q$th moment with $q>2$. A sufficient condition for \cref{ass:mu_moment} is
\begin{equation}\label{lip_h}
|h_r(x)|+|h_{r1;i}(x)| ~ \lesssim~  1+\|x\|_2^k
\end{equation}
for all $x\in \RR^L$ and 
for some $k\in\NN$, together with the sub-Gaussianity of $\mu$. In particular, when $\mu$ is Gaussian,   the above condition holds with $k=r+1$.

\begin{theorem}
    \label{prop:theta_gap}
Grant Assumptions  \ref{ass_alpha} - \ref{random-points} and \ref{ass:mu}, and let $\Delta_1$ be as in (\ref{cond_Delta1}). The following holds  up to some relabeling.\\
(i) There exists some constant $D > 0$, depending on $K,B,\Delta_1$ and $\ua$, such that
\begin{align*}
    &\|\bar\balpha -  \balpha^*\|_2    \leq  D \|\bar \bmvec - {\bm m}\|_2,\\  \quad 
    &\max_{k\in [K]}\|\bar\btheta_k - \btheta_k^*\|_2^2   \leq   D \Bigl(L\|\bar {\bm m} - {\bm m}\|^2_2 +  \sum_{i = 2}^L\|\bar\bmvec_{1;i}  - \bmvec_{1;i} \|^2_2\Bigr). 
\end{align*}
(ii) Moreover, if \cref{ass:mu_moment} also holds, 
the output $\bar \bomega : = (\bar \balpha, \bar \btheta_1, \ldots, \bar\btheta_K) $ given by \cref{alg_MoM}   satisfies   $d(\bar\bomega,\bomega^*) = \cO(\sqrt{L\log \log p / p})$ $\mu$-almost surely. 
\end{theorem}
The  proof of   Theorem  \ref{prop:theta_gap}  is given in \cref{app_proof_prop:theta_gap}. The proof of part (i)   is valid for {\it any} finite mixture estimated by the classical method MoM proposed by \cite{Lin89,Lindsay93}. Although partial results can be extracted from existing proofs, we are not aware of a comparable complete result for high-dimensional mixture models.
 
  \cref{prop:theta_gap}  shows that the quality of parameter recovery is  that of latent moment recovery, up to the constant $D$. Unfortunately,  this $D$   scales as $\Delta_1^{-c K}$ for some absolute constant $c$  and this scaling is tight (see  \cref{rem_rate} of \cref{app_sec_mom_rem} for more discussion). Such scaling in the atom separation distance $\Delta_1$ is  one of the reasons why  MoM is relatively unstable in finite samples, for mixtures with a large number of components $K$, even in low dimensions. We return to this point after \cref{thm_mom_est} of Section \ref{MoM-estimation}.

\subsection{MoM parameter estimation}\label{MoM-estimation}


MoM parameter estimators  in softmax mixtures can be constructed by an immediate application of \cref{alg_MoM}.  Recall that $Y_1, \ldots, Y_N$ are i.i.d. from $\pi^*$.  Given  functions $h_r$ and $h_{1r;i}$  defined by (\ref{eq:hr_def}) and (\ref{eq:hr1_def}), the empirical versions of $\bar m_r$ and $\bar m_{r1;i}$ in \eqref{cond-mom}   are 
\begin{equation}\label{m-hat} 
    \wh m_r := \frac{1}{N}\sum_{\ell=1}^{N}h_r(Y_\ell),  \qquad 
    \wh m_{r1;i} := \frac{1}{N}\sum_{\ell=1}^{N}h_{r1;i}(Y_\ell). 
\end{equation}
By forming the vectors $
    \wh{\bm m} = ( \wh m_1,\ldots, \wh m_{2K-1})^\T$ and $\wh \bmvec_{1;i} =   ( \wh m_{01;i},\ldots, \wh m_{(K-1)1;i})^\T$
for $i\in \{2,\ldots, L\}$, the MoM estimator $\wh\bomega = (\wh\balpha, \wh\btheta_1,\ldots,\wh\btheta_K)$ is given by \cref{alg_MoM}  with $\bar\bmvec$ and $\bar\bmvec_{1;i}$ replaced by $\wh\bmvec$ and $\wh\bmvec_{1;i}$, respectively.

The following theorem gives the rate of convergence of $d(\wh\bomega,\bomega^*)$, proved  in \cref{app_sec_proof_thm_mom_est}. 

 \begin{theorem}\label{thm_mom_est} Grant    \cref{ass_alpha,separation,ass_theta,ass:mu,ass:mu_moment,random-points} as well as condition (\ref{cond_Delta1}).
Then, $\mu$-almost surely, for all sufficiently large $p$, up to some label switching, $d(\wh\bomega, \bomega^*)  = \cO( \sqrt{L\log\log N / N} + \sqrt{L\log\log p/p})$ $\pi$-almost surely. 
 \end{theorem}  
The rate  of the MoM estimator is parametric in both $N$ and $p$, up to logarithmic factors. Comparing to the rate of the EM algorithm in \cref{thm_EM}, the rate is slower unless $p\ge N$. 

The proof of \cref{thm_mom_est}  reveals that the second term in the rate is inherited from \cref{prop:theta_gap}. Consequently, the finite-sample performance of the MoM estimator deteriorates exponentially in $K$, as discussed after \cref{prop:theta_gap}; see also \cref{rem_rate} of \cref{app_sec_mom_rem}. This fact is also  verified in our simulation studies in \cref{sec_sims}. 
\begin{remark}\label{rem-primary}  
 As it is  evident from  the sample counterpart of  \cref{alg_MoM},  the construction of the MoM  estimators begins by the estimation of the first coordinate of each mixture atom, a key step for the rest of the process.  In practice, the ``first coordinate'' of each atom   is  replaced by its projection  onto   an axis $v \in \RR^L$, which needs to be chosen so that \eqref{cond_Delta1} holds along this axis. Otherwise, one cannot expect parametric rates, as discussed in \cref{non-param} of \cref{app_sec_mom_rem}. To this end, we provide a detailed analysis, with theoretical guarantees, of possible random choices of $v$. Due to space limitations, these results are deferred to \cref{sec_mm_proj}. Our overall conclusion is that the practical choice of a primary axis is extremely delicate. Combined with the deteriorating performance of MoM as $K$ increases, this leads us to not  recommend the method of moments for parameter estimation in softmax mixtures when \(K\) is not small. Our empirical studies suggest that MoM-based parameter estimation can become problematic even for $K=10$.
\end{remark} 
 On the other hand, from a rate perspective, \cref{thm_mom_est} also shows that the MoM estimator   is a potential warm-start candidate for the EM algorithm: it satisfies the initialization requirement in \eqref{cond_init_rand} for any  $p\ge C(L\log p)^2$ and $N$ satisfying (\ref{cond_N_rand}), although $C=C(K)$ still grows exponentially in $K$. \cref{thm_EM_MoM} provides a formal statement of this result.



\subsection{MoM latent subspace estimation}\label{latent-space} 

In Section \ref{MoM-estimation} we argued that MoM for {\it parameter estimation} may be fragile, in general. Here we re-emphasize that the potential drawbacks of MoM stem from solving the system of equations in \cref{alg_MoM}, particularly for large feature dimension $L$ and when $K$ is not small (e.g., $K\ge 10$).  However, if the goal is instead the estimation of the {\it span} of $\btheta_1^*, \ldots, \btheta_K^*$, henceforth called {\it latent subspace estimation}, this task can be accomplished more directly, and does not require solving \cref{alg_MoM}.  This is particularly useful in practice when $L$  is much larger than  $K$. As important applications, the estimated subspace can be used in two ways: (1) to select a basis in which the primary axis condition in \eqref{cond_Delta1} holds (see \cref{sec_mm_proj}); and (2) to reduce the number of required random initializations for the EM algorithm, when such initializations are employed (see \cref{sec_EM_SS}). 

To estimate the subspace of  $\btheta_1^*, \ldots, \btheta_K^*$, we note that it suffices to estimate the $K$-dimensional subspace spanned by the columns of   
\begin{equation}\label{def_Gamma}
    \Gamma := \sum_{k=1}^K \alpha_k^*  \btheta_k^* \btheta_k^{*\T} \in \RR^{L\times L}.
\end{equation}
Recall the expression of $h_r$ in \eqref{eq:hr_def} with $r=2$. For
$
     h_2(x; \be_1)  
     :=  (f_\mu(x))^{-1} \be_1^\T \nabla^2 f_\mu(x) \be_1
$
with $\nabla^2 f_\mu(x)$ being the Hessian matrix of $f_\mu$ at $x$, 
we have 
\[
     {\EE_\mu[   h_2(X; \be_1) \exp(X^\T \btheta)]  \over  \EE_\mu[ \exp(X^\T \btheta) ]} = \be_1^\T \btheta \btheta^\T \be_1
\]
for any $\btheta\in\RR^L$. As this also holds with each of $\be_2,\ldots,\be_L$ in place of $\be_1$,  the moment matrix
\begin{equation}\label{def_Gamma_bar}
    \bar\Gamma := \EE_{\pi^*} \left[
       {\nabla^2 f_\mu(Y) \over  f_\mu(Y)}
    \right] 
\end{equation}
has its population (w.r.t $\mu$) version
$$
     \sum_{k=1}^K \alpha_k^*~ \EE_\mu\left[\exp(X^\T \btheta_k^*)   \nabla^2 f_\mu(X)\over f_\mu(X)\right] {1\over \EE_\mu[ \exp(X^\T \btheta_k^*) ]},
$$ 
 which is equal to $\Gamma$.
Therefore, when we observe  $Y_1, \ldots, Y_N$ i.i.d $\pi^*$, the matrix  
\begin{equation}\label{def_Gamma_hat}
    \wh \Gamma := {1\over N}\sum_{i=1}^N    
       {1\over f_\mu (Y_i)} \nabla^2 f_\mu(Y_i),
\end{equation}
 which is the sample analogue of $\bar{\Gamma}$,  should also estimate 
$\Gamma$ well, so that its first $K$ eigenvectors can be used to estimate the span of $\btheta_1^*, \ldots, \btheta_K^*$. The following proposition provides the justification and its proof is  stated in \cref{app_proof_prop_eigensp}. 
As in \cref{ass:mu_moment}, we require finite third moments of $h_r(x;\be_i)$ defined in \ref{eq:hr_def}, but only for $r=2$ and for all $i\in [L]$.
The existence of $h_2(x,\be_i)$ requires only second-order partial derivatives of $f_\mu$, rather than partial derivatives up to order $2K-1$ as in \cref{ass:mu}.
\cref{ass:mu_moment_2} substitutes  both \cref{ass:mu,ass:mu_moment} in this section.
 
\begin{assumption}\label{ass:mu_moment_2}
Assume that $\mu$ is absolutely continuous with respect to the Lebesgue measure on $\RR^L$ and has a finite moment generating function everywhere. Its density $f_\mu$ is strictly positive,  and $ f_{\mu} \in C^{2}(\RR^L)$, with 
partial derivatives up to order $2$ decaying super-exponentially at infinity.
Moreover, 
    $\EE_\mu[|h_2(X;\be_i)|^3] < \i$ for all $i\in [L]$.
\end{assumption}

 As in \eqref{lip_h}, a sufficient condition for the third moment condition in  \cref{ass:mu_moment_2} is
$
|h_2(x; \be_i)|  \lesssim   1+\|x\|_2^k
$
for all $x\in \RR^L$ and 
for some $k\in\NN$, together with the sub-Gaussianity of $\mu$ in \cref{random-points}.  

\begin{proposition}\label{prop_eigensp}
    Grant  \cref{random-points,ass:mu_moment_2,ass_theta}.
    Then $\mu$-almost surely for all $p$ sufficiently large,   
    $
        \|\wh \Gamma - \Gamma \|_\op  = \cO(L\sqrt{\log\log N/N} + L\sqrt{\log\log p/p}),
    $
    $\pi$-almost surely.
    \end{proposition}


\cref{prop_eigensp} requires neither strictly positive mixing weights nor separation between the mixture atoms, and is of independent interest. 

The estimator in \eqref{def_Gamma_hat} requires knowledge of the density $f_\mu$.
In the Gaussian case, $\mu = \cN_L(0,\Sigma)$, it takes the simple form
 \begin{equation}\label{def_Sigma_Y}
        \wh \Gamma  = {1\over N}\sum_{i = 1}^N  \Sigma^{-1} Y_i Y_i^\T\Sigma^{-1}  - \Sigma^{-1}.
    \end{equation}
When $\Sigma$ is unknown, it can be consistently estimated by the sample version $ p^{-1} \sum_{i=1}^p X_i X_i^\T$. 
Finally, in  \cref{sec:subspace-elliptical}, 
we estimate $\text{span}(\btheta_1,\ldots,\btheta_K)$ using only  the sample matrices $n^{-1}\sum_{i=1}^nY_iY_i^\top$ and $ p^{-1} \sum_{i=1}^p X_i X_i^\T$,  without making use of $\mu$ explicitly, but assuming that it has  a   centered elliptical symmetric density (the $\cN_L(0,\Sigma)$ density is a special case).

\section{Synthesis of  results}\label{sec_synthesis}

A combination of the results of  Sections \ref{EM} and \ref{sec_MoM} yield end-to-end procedures for parameter estimation in softmax mixtures. This is  the content of \cref{sec_EM_MoM,sec_EM_SS}.

\subsection{EM + MoM warm start}\label{sec_EM_MoM}

As discussed after Theorem \ref{thm_mom_est}, and recommended when $K$ is small, the method-of-moments (MoM) procedure yields parameter estimates that can be directly used as a warm start for the EM algorithm. The following theorem combines our analyses of the EM algorithm (\cref{thm_EM})  and the MoM parameter estimation (\cref{thm_mom_est}) to provide theoretical guarantees for this procedure.

\begin{theorem}\label{thm_EM_MoM}
    Under \cref{ass_alpha,separation,ass_theta,random-points,ass_Htheta_mu,ass:mu,ass:mu_moment}, assume condition \eqref{cond_sep_rand} for $\Delta$ and condition \eqref{cond_Delta1} for $\Delta_1$. Then $\mu$-almost surely for all $p$ sufficiently large such that $p\ge C_1L^2\log^2(p)$, the following holds $\pi$-almost surely for all sufficiently large $N$ satisfying   $N \ge C_1 L^2\log(p)\log (N)$: the EM iterates $\wh\bomega^{(t)}$, $t\ge 0$, defined by \eqref{iter_alpha} and \eqref{iter_theta}, with $\wh\bomega^{(0)}$ given by  the output of the MoM estimation procedure in \cref{alg_MoM} based on $\wh\bmvec$ and $\wh\bmvec_{1;i}$, $2\le i\le L$,   satisfy $d(\wh\bomega^{(t)},\bomega^*)\le C_2 \sqrt{\oa K L\log N/(\ua^2 N)}$ after $\cO(\log N)$ iterations, up to some label switching. The constant  $C_1$ depends on $\Delta_1$, $K$, $B$, $\ua$, $\os_\mu$ and $\us_\mu$ while $C_2$ is some absolute constant. 
\end{theorem}

The lower bounds on $p$ and $N$ are required for the MoM estimator to meet the initialization requirement of the EM algorithm in \eqref{cond_init_rand}. As discussed after \cref{prop:theta_gap}, the constant $C_1$ in \cref{thm_EM_MoM} scales as $\Delta_1^{-\cO(K)}$.  This dependence can be improved by using the initialization procedure introduced in the next section.


\subsection{EM + random subspace initializations }\label{sec_EM_SS}

 In addition to using the MoM parameter estimates mentioned in \cref{MoM-estimation},
  it is common practice to employ random initializations,  by simply drawing $\btheta_1^{(0)},\ldots, \btheta_K^{(0)}$ uniformly from a chosen sphere multiple times,  and selecting the corresponding EM estimate that yields the highest likelihood \citep{dasgupta2007probabilistic}.

The intuition is the following: Let $\btheta^* \in \bS^{L-1}$ be a  given target vector on the $\ell_2$-unit ball of $\RR^L$ around the origin, and let $\delta>0$ be the desired accuracy. Then for any $\varepsilon > 0$, with probability at least $1-\varepsilon$, there exists at least one vector $\bar v$ in independent draws $\{v_1,\ldots, v_m\}$ from $\bS^{L-1}$ such that $
     \|\bar v - \btheta^*\|_2 \le \delta
$
provided that
$
    m \ge C\sqrt{L}(1/\delta^2)^{(L-1)/2}\log(1/\varepsilon)
$
where $C>0$ is some absolute constant.
This result is a standard spherical-cap hitting bound; we provide a proof in \cref{app_sec_proof_lem_rand_init} to make its dependence on $L$, $\delta$ and $\varepsilon$ explicit. 
As a result, in the worst case one needs to use $\exp(\cO(L))$ random initializations and run the EM algorithm this many times, which is computationally expensive when $L$ is not small.

However, if $\btheta^*$ is known to lie within a subspace of dimension at most $K \ll L$, then one only needs $\exp(\cO(K))$ random draws from the unit sphere in this subspace to achieve the desired $\delta_\mu$ accuracy. We formalize this in the following lemma in our context. Recall that $\btheta_1^*,\ldots,\btheta_K^*$ lie in the column space of $\Gamma$ given in \eqref{def_Gamma}. Further recall that $\wh V \in \bO_{L\times K}$ contains the first $K$ leading eigenvectors of  $\wh \Gamma$, the estimator of $\Gamma$ given in \eqref{def_Gamma_hat}.

\begin{lemma}\label{lem_rand_init} 
     Fix $\btheta_1,\ldots,\btheta_K^*\in \bS^{L-1}$. Let $v_1,\ldots,v_m$ be defined by $v_i = {\wh V\wh V^\T u_i / \|\wh V\wh V^\T u_i\|_2}$ where $u_1,\ldots,u_m$ are i.i.d. samples from $\cN_L(0,\bI_L)$.  For 
      arbitrary $\varepsilon > 0$ and $\delta>0$, let
    \begin{equation}\label{lb_m}
        m \ge C\sqrt{K}\left(
                2/\delta^2 \right)^{(K-1)/2}
            \log(K/\varepsilon).
    \end{equation}
    Then on the event $\{\|\wh \Gamma - \Gamma\|_\op \le (\ua/8)\delta^2\}$, the following holds with probability at least $1-\varepsilon$: for each $k\in [K]$, there exists $\bar v_k \in  \{v_1,\ldots, v_m\}$ such that 
    $
         \|\bar v_k - \btheta_k^*\|_2 \le \delta.
    $
    
\end{lemma} 
The proof of \cref{lem_rand_init} can be found in \cref{app_sec_proof_lem_rand_init}. The order of $m$ is essentially the same as the cardinality of a discretization of the estimated subspace intersected with $\bS^{L-1}$. However, random sampling provides greater computational flexibility and is naturally amenable to parallel computation.

When  $\btheta_1^*,\ldots,\btheta_K^*\in \bS^{L-1}$, using  $\delta_\mu$ in \eqref{cond_init_rand} and \cref{prop_eigensp},  the event $\{\|\wh \Gamma - \Gamma\|_\op \le (\ua/8)\delta_\mu^2\}$ holds $\mu$-almost surely for all sufficiently large $p$ satisfying $L^3\log^2(p) = o(p)$ and $\pi$-almost surely for all sufficiently large $N$ such that $L^3\log(N)\log(p) = o(N)$. As a result, for $m$ random draws satisfying \eqref{lb_m} for any $\delta\le \delta_\mu$, we initialize $(\wh\btheta_1^{(0)},\ldots,\wh\btheta_K^{(0)})$ using each of the ${m\choose K}$ combinations of the $m$ draws. The following theorem ensures that at least one of the resulting EM outputs achieves a near-parametric rate within $\cO(\log N)$ iterations. Its proof follows by combining \cref{thm_EM}, \cref{prop_eigensp} and \cref{lem_rand_init}. For simplicity, we assume that $\balpha^*$ is known and focus on the more challenging problem of estimating the atoms.
 
\begin{theorem}\label{thm_EM_SS}
    Under \cref{ass_alpha,separation,ass_theta,random-points,ass_Htheta_mu,ass:mu_moment_2}, assume  $\btheta_1^*,\ldots,\btheta_K^*\in \bS^{L-1}$, $\balpha_k^*$ is known and condition \eqref{cond_sep_rand} holds for $\Delta$. Then for running the EM algorithm ${m}\choose{K}$ times initialized above with $m$ satisfying \eqref{lb_m} with any $\delta\le \delta_\mu$, at least one of the outputs $\wh\bomega^{(t)}$, $t\ge 0$, satisfy: $\mu$-almost surely for all sufficiently large $p$ such that $p \ge C_1L^3\log^2(p) $, and $\pi$-almost surely for all sufficiently large $N$ satisfying   $N \ge C_2 L^3\log(p)\log (N)$:  up to some label switching, $d(\wh\bomega^{(t)},\bomega^*)\le C_3 \sqrt{\oa K L\log N/(\ua^2 N)}$ after $\cO(\log N)$ iterations.  Here the positive constants  $C_1$ and $C_2$ depend  on $B$, $\os_\mu$ and $\us_\mu$ via $c_\mu$ in \eqref{cond_init_rand} while $C_3>0$ is some absolute constant. 
\end{theorem}
Compared with \cref{thm_EM_MoM}, the subspace estimation error based on MoM no longer depends on $\Delta_1^{-\cO(K)}$. Consequently, the lower-bound requirements on $p$ and $N$ needed to  meet the initialization requirement have a more favorable dependence on $\Delta_1$ and $K$, as reflected in the constants $C_1$ and $C_2$, which   differ from those in \cref{thm_EM_MoM}.

In \cref{thm_EM_SS}, we focused on the case where the atoms lie on the unit sphere,  which is a special case of \cref{ass_theta}. Under \cref{ass_theta} only, we additionally need to discretize the radius over $[0,B]$, which  requires only a simple one-dimensional discretization. A result similar to the one above can then be obtained. Similarly, for general $\balpha^*$, one can initialize $\wh\balpha^{(0)}$ from an $(\ua\delta_\mu)$-cover of the simplex $\Delta^{K-1}$, whose cardinality is of order $(\ua\delta_\mu)^{-K+1}$. Accordingly, the number of EM runs specified in \cref{thm_EM_SS} increases by a multiplicative factor of this order.

\subsection{On the limiting behavior of SMMs and their asymptotic identifiability}
\label{sec_mom_ident}

We begin by observing that, when $X_1, \ldots, X_p$ i.i.d. $\mu$,   the discrete softmax components may be viewed as random approximations to exponential tilts of ${\mu}$, as $p \rightarrow \infty$. 
This allows us to characterize the limiting behavior of SMMs, for large $p$, a result that may be of independent interest.
Let $\psi(\btheta)
:=
\log \EE_\mu [\exp(\btheta^\top X)]<\infty,
$
and define the exponentially tilted measure
%
%
\begin{equation} \label{tilt-measure} 
\mu(B;\btheta)
:= \int_B
\exp\!\left\{x^\top\btheta-\psi(\btheta)\right\} \, d{\mu}(x) \qquad \text{for every Borel set $B\in \RR^L$.}
\end{equation}

\begin{proposition}\label{prop:weak}
Assume  $\psi(\btheta_k)<\infty$ for all $k\in [K]$.
Then, for
$\mu^{\otimes\mathbb N}$-almost every sequence $\{X_i\}_{i\ge1}$, 
 $Y_p \sim 
\pi(\cdot;\bomega)= \pi_p(\cdot;\bomega)$ converges weakly to a limit $Y\sim \sum_{k=1}^K\alpha_k \mu(\cdot;\btheta_k)$,
as $p\to\infty$.
\end{proposition}
The proof is given in \cref{app-weak-limit-SMM}. 
For example, if $\mu$ is the $\cN_L(0,\bI_L)$, then
a simple calculation shows that 
$\mu(\cdot;\btheta)= \cN_L(\btheta,\bI_L)$ 
and thus
 the limiting model is a Gaussian mixture with the same mixing
weights $\alpha_k$ and $\cN_L(\btheta_k,\bI_L)$ components.

Finite Gaussian mixtures with nonzero weights and distinct component means are identifiable up to label permutation. More generally, in \cref{app-asym-iden}, as part of the proof of \cref{prop:iden}, we show that finite mixtures of distinct exponential tilts of $\mu$ are identifiable, provided that $\mu$ is absolutely continuous with respect to the Lebesgue measure and has a finite moment generating function. This, in turn, implies asymptotic identifiability in SMMs, as stated below.

\begin{theorem} \label{prop:iden}
Suppose that $\mu$  has continuous density $f_{\mu}$, and that
its moment generating function is finite everywhere.  
Let
$\bomega^*=(\alpha^*,\btheta_1^*,\ldots,\btheta_K^*)$ and 
$\bomega^\dagger = (\alpha^\dagger,\btheta_1^\dagger,\ldots,\btheta_K^\dagger)
$ be two fixed parameter configurations satisfying Assumptions \ref{ass_alpha} and \ref{separation}.

Then,
for $\mu^{\otimes\mathbb N}$-almost every sequence $(X_i)_{i\ge1}$,
$$
\bomega^*\sim\bomega^\dagger \quad\Longleftrightarrow\quad \pi(\cdot;\bomega^*)
= \pi(\cdot;\bomega^\dagger) \quad
\text{for all sufficiently large }p.
$$
Here $\sim$ denotes equality up to label permutation.\end{theorem}


\begin{remark} \label{rem-p0}
Theorem \ref{prop:iden} is general, but  its proof does not characterize the order of magnitude of a threshold $p_0$ above which its conclusion   holds. This can be obtained  at the price of additional assumptions. \cref{prop:iden2} in \cref{app-new-asym-iden} shows if  
 Assumptions  \ref{ass_alpha} - \ref{random-points} and \ref{ass:mu} - \ref{ass:mu_moment} hold,   then $p_0 = \cO(L/\Delta_{1}^{4K-2})$, for any $\Delta_1 > 0$, where $\Delta_1$ is the atom  separation defined in (\ref{cond_Delta1}). We believe that this order of magnitude is conservative but necessary for the MoM approach used in the proof of \cref{prop:iden2}, as noted in \cref{rem_rate,app_sec_mom_rem}; at a minimum, one needs $p>L$. In contrast, if we strengthen the requirement $\Delta_1 >0$ to $\Delta_1 \gtrsim \log K$, then  {\it local} identifiability holds, for all $p > 1$, and for any configuration of the design points satisfying the conditions of  \cref{locid} in \cref{app_sec_EM_popu}. Finding sufficient conditions under which {\it global} identifiability of SMMs holds, for {\it all} $p > 1$, remains a challenging open problem. 
\end{remark}
\begin{remark} 
It may be of independent interest  to re-emphasize that when $\mu$ has continuous density $f_{\mu}$,  
Proposition \ref{prop:weak}  shows that SMMs are  discretizations of continuous mixtures  of exponential tilts  of $f_{\mu}$ given by  
$f(x;\btheta)
:=
\exp\!\left\{x^\top\btheta-\psi(\btheta)\right\}f_{\mu}(x),$
and these mixtures  are identifiable by the proof of  Theorem \ref {prop:iden}, and under its conditions.
\end{remark}

\section{Simulations}\label{sec_sims}

In this section we conduct numerical experiments to corroborate the theoretical findings of  \cref{EM,sec_MoM,sec_synthesis}. We first examine how the performance of the EM and MoM estimators depends on $N$, $p$, $L$ and $K$. We then demonstrate the benefit of using the dimension reduction technique from \cref{sec_EM_SS} to initialize the EM algorithm. Further examination of the convergence rate of the EM algorithm is provided in \cref{app_sec_sim_rate}.

To generate the data, we generate $X_1, \ldots, X_p$ i.i.d. from $\mu = \cN_L(0, \bI_L)$. We set $\balpha^* = (1/K) 1_K$ as the mixing weights  
and set $\btheta_1^*, \ldots, \btheta_K^*$ as the $K$ left singular vectors of an $L\times K$ matrix with i.i.d. $\cN(0,1)$ entries. Finally,   $Y_1,\ldots, Y_N$ are resampled according to model \eqref{mix}.

We consider the following estimation methods: (1) MoM, the Method of Moments estimator in \cref{MoM-estimation} with $n = 200$ for choosing the projection direction as discussed in \cref{rem_axis}; (2) EM-MoM, the EM estimator in \cref{EM} that uses
the MoM estimator as initialization;
  (3) EM-dr-rand-10, the EM estimator achieving the highest likelihood among 10 random initializations restricted to the estimated subspace of $\btheta_1^*,\ldots,\btheta_K^*$;\footnote{Entries of $\wh\btheta_k^{(0)}$, $k\in [K]$, are i.i.d. from $\cN(0,1/\sqrt{L})$ while entries of $\wh\balpha^{(0)}$ are set to $1/K$.}
 (4) EM-oracle, the EM estimator that uses the true parameter values as the initialization. 
For all EM procedures, we choose the step size $\eta_k = 0.2$ and
use the stopping rule that the relative change of the log-likelihood is smaller than $10^{-6}$. 
To evaluate each method, for generic estimators $\wh\balpha$ and $\wh\btheta_1,\ldots,\wh\btheta_K$, we choose 
$
\text{Err}_\btheta =  (K^{-1}\sum_{k=1}^K \|\btheta_k^* - \wh \btheta_{\varrho(k)}\|_2^2)^{1/2}$ and $\text{Err}_\balpha =   \sum_{k=1}^K|\alpha_k^*-\wh \alpha_{\varrho(k)}|$
where $\varrho: [K]\to [K]$ is the best permutation that minimizes $\text{Err}_{\btheta}$. 

\begin{figure}[ht]
    \centering 
       \includegraphics[width=.85\linewidth]{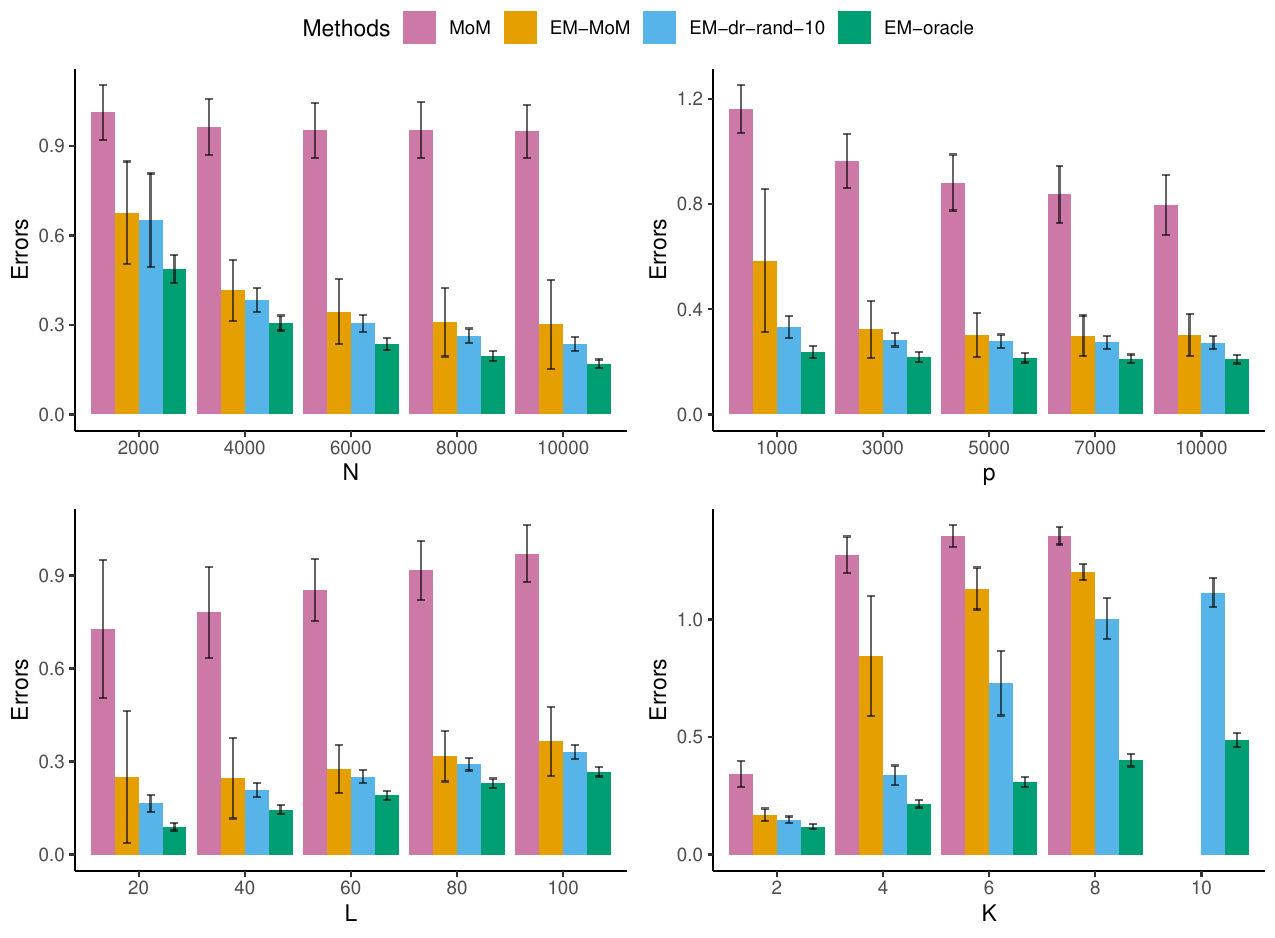}\vspace{-6mm}
    \caption{The averaged $\text{Err}_{\btheta}$ in different settings}\vspace{-2mm}
    \label{fig_errors}
\end{figure}

\paragraph{Dependence of estimation error on $N, p, L$ and $K$.}
 We  vary $N \in \{2, 4, 6, 8, 10\} \times 10^3$, $p\in \{1, 3, 5, 7, 10\}\times 10^3$, $L\in \{20, 40, 60, 80, 100\}$ and $K\in \{2,4,6,8,10\}$ one at a time to examine their effects on the estimation errors. The baseline setting uses $L = 50$ and $K = 3$ when these parameters are not varied. When $N$ is varied, we set $p = 5000$ and when $p$ is varied, we set $N = 7000$. When varying either $L$ or $K$, we chose $N = 10000$ and $p = 7000$. For each setting, we report the averaged errors  over $1000$ repetitions in \cref{fig_errors} for $\text{Err}_\btheta$. The corresponding plot of $\text{Err}_\balpha$ is provided in \cref{fig_weights} of \cref{app_sec_sim_plots}.

Regarding  $\text{Err}_\btheta$, all methods perform better as $N$ increases and $L$ or $K$ decreases. For EM-oracle, since it has no algorithmic error, our \cref{thm_EM} shows that its estimation error is purely the statistical error which is of order $\sqrt{L\log(N)/N}$. The MoM estimator is outperformed by the EM estimators in all settings.  Once $N \ge p$, further increasing $N$ does not improve the performance of MoM. When $p$ increases, the performance of MoM improves, whereas the EM estimators remain unchanged. We also note that the figures in which we  vary $N$ and $p$ suggest the rate for MoM is slower than the parametric rate, confirming the observation made in Remark \ref{non-param} of \cref{app_sec_mom_rem}. In \cref{app_sec_sim_parametric}, we conduct a separate simulation study below to verify that MoM can indeed enjoy a parametric rate.  

For $K = 3$, EM-MoM and EM-dr-rand-10 have overall comparable performance, with the former performing slightly better for large $N$. One drawback of EM-dr-rand-10 is its higher computational cost due to sampling multiple initializations and evaluating their likelihoods (the computational complexity scales linearly with the number of initializations multiplied by the ambient dimension $p$). As $K$ increases, the performance of all methods deteriorates, with MoM and EM-MoM degrading more rapidly than the others. For $K = 10$, MoM (so does EM-MoM) fails to recover all $K$ mixture components, as the root-finding step in \cref{alg_MoM} fails in this case. These findings are all aligned with our theory in \cref{EM,sec_MoM}.  


\begin{figure}[ht]
    \centering 
       \includegraphics[width=.9\linewidth]{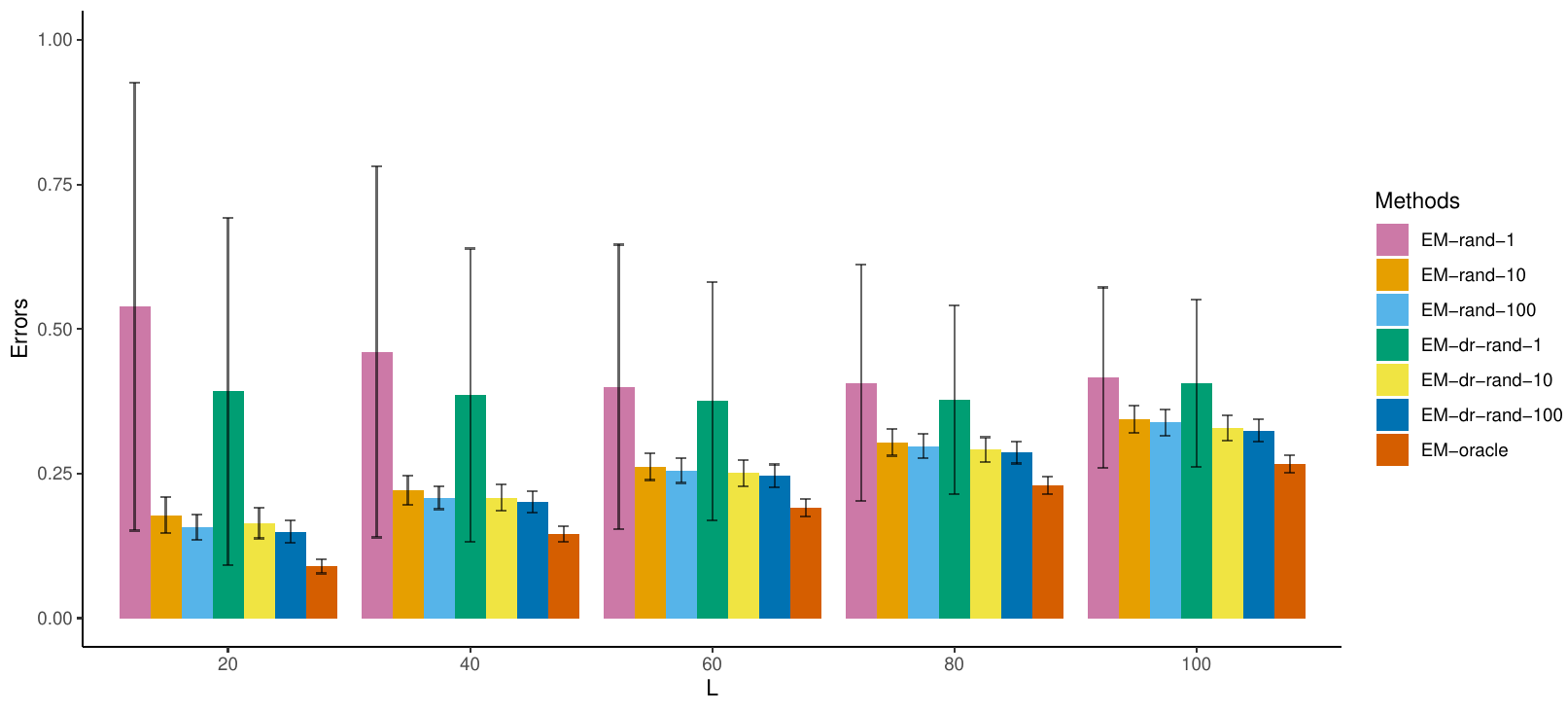}
    \caption{The averaged $\text{Err}_{\btheta}$ in different settings}
    \label{fig_rand_init}
\end{figure}

\paragraph{Multiple random initializations with dimension reduction.}
We proceed to verify the benefit of using dimension reduction as well as multiple random initializations in the EM algorithm. We set $N=10000$, $p=7000$ and $K=3$. In addition to EM-dr-rand-$m$ with $m\in \{1,10,100\}$, we also consider the variant, EM-rand-$m$, the EM estimator that uses $m$ random initializations without projected to the estimated subspace. 
\cref{fig_rand_init} shows that using multiple random initializations yields better performance than a single random draw. Moreover, the benefit of incorporating dimension reduction is evident for both single and multiple random initializations, and becomes increasingly important as the ratio $L/K$ grows. Finally, the gap between EM-dr-rand-$m$ and EM-oracle narrows as $m$ increases.

\bibliographystyle{abbrvnat}
\bibliography{ref}

 \newpage

\appendix

    \cref{app_sec_EM} contains additional theory of the EM algorithm for generic features and relevant proofs.  
    \cref{app_sec_MoM} contains supplementary results to the MoM estimation of \cref{sec_MoM} including practical selection of the primary axis and  subspace estimation under elliptically symmetric distributions.
    The proofs of \cref{sec_synthesis} and auxiliary lemmas are collected in \cref{app_proof_sec_syn} and \cref{app_auxiliary}, respectively. 
    Additional simulation results are stated in \cref{app_sec_sim}. A real data analysis is given in \cref{sec_real_data}.

\section{Results  for  the EM algorithm: Section \ref{EM} }\label{app_sec_EM}

We give theoretical guarantees of the EM algorithm when  the feature matrix $\bX = (x_1^\T,\ldots, x_p^\T)^\T \in \RR^{p\times L}$ is treated as deterministic. When the features are random under \cref{random-points}, the results derived here can be understood as conditional on $\bX$. The results in \cref{EM-theory} can then be readily obtained by unconditioning on $\bX$. We first establish a convergence result for the population-level EM algorithm and discuss its implications for local identifiability. We then establish the convergence of the sample-level EM algorithm.

We need an analogue of \cref{ass_Htheta_mu} for the deterministic design. For any $\btheta \in \RR^L$ with $A(\btheta)\in \Delta^{p-1}$, recall that $H_{\btheta} = \bX^\T (\diag(A(\btheta)) - A(\btheta)A(\btheta)^\T) \bX\in \RR^{L\times L}$ denotes the Fisher information matrix under a single softmax parametrization $A(\btheta)$.
    
    \begin{assumption}\label{ass_X}
       There exist some constants $0< \us^2 \le \os^2 <\i$  and $\vs^2<\i$, such that  
        all $a,b\in [K]$ and $u\in [0,1]$ with $\btheta = u\btheta_a^*+(1-u)\btheta_b^*$, 
        $
             \us^2\le \lambda_L(H_\btheta)\le \lambda_1(H_\btheta) \le \os^2$
       and $\lambda_1(H_{\btheta}^{-1/2}\bX^\T \diag(A(\btheta)) \bX H_{\btheta}^{-1/2})\le \vs^2.$
    \end{assumption}
    The first condition   ensures that $H_{\btheta}$ remains well-conditioned along the line segment connecting any pair of atoms $\btheta_1^*,\ldots,\btheta_K^*$. The second condition  is technical, which follows from the first condition and $\lambda_1( \bX^\T \diag(A(\btheta)) \bX)\le  \vs^2/ \us^2$, that is, the $L\times L$ matrix $\bX^\T \diag(A(\btheta)) \bX $ is well-behaved.  We show in the proof of \cref{thm_EM} that Assumption \ref{ass_Htheta_mu} implies  \cref{ass_X}, $\mu$-almost surely, with  $\us^2 = \us_\mu^2/2$ and $\os^2= 2\os_\mu^2$  , and $\vs^2\le C(\os_\mu^2,B)$. For future reference, we note that  $\vs \ge 1$ and $\|\bX\|_{\i,2} = \max_{j\in[p]} \|x_j\|_2\ge \os$.

    The analogue of the separation requirement in \eqref{cond_sep} under \cref{ass_X} becomes 
     \begin{equation}\label{cond_sep}
        \us^2   \Delta  \ge   C  \left\{
                    \log K + \log {\os^2\over \us^2} + \log {\oa\over \ua}
                \right\}
     \end{equation}
     for some absolute constant $C>0$. Similarly,  the corresponding sample complexity condition in \eqref{cond_N_explict} is  
     \begin{equation}\label{cond_N_explict}
         {\ua N\over \log N} \ge C'~  {\oa   \over \ua}{\os^2\over \us^2}  {\|\bX\|_{\i,2}^2\over \us^2} ~ KL
     \end{equation}
     for some absolute constant $C'>1$.

 Finally, with a slight abuse of notation, we let $d(\bomega,\bomega’)$ in \eqref{def_dist} denote the corresponding distance with $\os_\mu$ replaced by $\os$.

\subsection{Convergence of the population-level EM algorithm and local identifiability: generic design}\label{app_sec_EM_popu}

   We start by defining the {\em population-level} EM algorithm, which is obtained by replacing the empirical frequency $\wh \pi$ in the iterates $\wh\bomega^{(t+1)}$ in \eqref{iter_alpha} and $\wh \btheta_k^{(t+1)}$ in \eqref{iter_theta} with the true probability vector $\pi(\bomega^*)$, and uses the population-level function 
   \begin{align}\label{def_Q_popu}
			 Q(\bomega \mid \bomega') 
			& =  \sum_{j=1}^p \pi(x_j;\bomega^*)  \sum_{k=1}^K {\alpha_k' A(x_j;\btheta_k') \over \pi(x_j;\bomega')}    \left[
			\log ( \alpha_k) + x_j^\T \btheta_k  - \log\left(
			\sum_{\ell=1}^p \exp\left({x_{\ell}^\T \btheta_k}\right)
			\right)
			\right] 
	\end{align}
        Specifically, for some initialization $\bomega^{(0)} = (\balpha^{(0)}, \btheta_1^{(0)}, \ldots,\btheta_K^{(0)})$, the population-level EM iterates as: for $t\ge 1$,
	\begin{align}\label{iter_alpha_popu}
    	 \alpha_k^{(t+1)}  
      &=\sum_{j=1}^p \pi(x_j;\bomega^*) 
      {\alpha_k^{(t)}   A(x_j;  \btheta_k^{(t)}) \over 
      \pi (x_j;  \bomega^{(t)})}
      :=  M_k( \bomega^{(t)}),
    	\end{align}
    and  
    	\begin{align}\label{iter_theta_popu} 
    		\btheta_k^{(t+1)} &=  \btheta_k^{(t)} +  \eta_k ~  \nabla_{\btheta_k}   Q(\bomega \mid      \bomega^{(t)})   \mid_{\bomega =  \bomega^{(t)}}\\\nonumber
            &=    \btheta_k^{(t)} +  \eta_k ~  \sum_{j=1}^p  \pi(x_j;\bomega^*) {  \alpha_k^{(t)} A (x_j; \wh \btheta_k^{(t)})\over \pi(x_j;  \bomega^{(t)})} \left( x_j - \bX ^\T A( \btheta_k^{(t)})\right) 
    	\end{align} 
    for $k\in [K]$. The following corollary gives its convergence guarantee.

    \begin{corollary}[Convergence of the population-level EM]\label{cor_EM_popu}
       Given any support points $\{x_1, \ldots, x_p\}$ of the softmax mixture,  
         let $\bomega^*$ satisfy  Assumptions \ref{ass_alpha}, \ref{separation} and \ref{ass_X}, with $\Delta$ satisfying  \eqref{cond_sep}. 
        Assume  that the initialization  $\bomega^{(0)} $ satisfies 
               \begin{align}\label{cond_init_fix}
                  d( \bomega^{(0)} ,  \bomega^*)\le \delta_0  \qquad\text{ with }\quad  \delta_0  \le  {c_0  \over  \vs^2}{\os\over  \|\bX\|_{\i,2}}
            \end{align}
            for some sufficiently small constant $c_0\in [0,1/2)$.  Then, there exist some $0<\phi <1$  and step-sizes $\eta_k>0$, $k\in [K]$, such that the population-level EM iterates $ \bomega^{(t)}$ satisfy:  
    $$
      d(\bomega^{(t)}, \bomega^*) ~ \le ~    \delta_0 ~  \phi^t,\qquad \forall ~ t\ge 0.
    $$ 
	\end{corollary} 
    \begin{proof} 
      We only prove \cref{thm_EM_fix} below since the proof of \cref{cor_EM_popu}  follows by replacing  the quantities  $\wh M_k$,
        $\wh Q$ and $\wh\bomega^{(t)} $ by
        $ M_k$, $Q$ and $\bomega^{(t)}$, respectively, and setting  $N=\infty$.
    \end{proof}

An important implication of \cref{cor_EM_popu} is the following local identifiability result for the softmax mixture model.
 
\begin{corollary}[Local identifiability]\label{locid}
    Grant Assumptions \ref{ass_alpha}, \ref{separation} and 
    \cref{ass_X} with $\Delta$ satisfying  \eqref{cond_sep}. Suppose there exist two parameter points $\bomega_1^*$ and $\bomega_2^*$ such that $\pi^* = \pi(\bomega_1^*) = \pi(\bomega^*_2)$, and both satisfy \eqref{cond_sep}   for their corresponding $\btheta_k^*$'s. If $d(\bomega_1^*,\bomega_2^*) \le  \delta_0 /2$, for $\delta_0 $ given by \eqref{cond_init_fix}, then $\bomega_1^* = \bomega_2^*$. 
\end{corollary}
 \begin{proof}
		Fix any $\bomega^{(0)}$ that satisfies $d(\bomega^{(0)}, \bomega_1^*) \le \delta_0/2$. By triangle inequality, we also have 
		$d(\bomega^{(0)}, \bomega_2^*) \le \delta_0$. By  \cref{cor_EM_popu}, $\lim_{t \rightarrow \infty} d(\bomega^{(t)},  \bomega_1^*) = 0 =  \lim_{t \rightarrow \infty}d(\bomega^{(t)},  \bomega_2^*)$, and thus $\bomega_1^* = \bomega_2^*$, by the uniqueness of the limit in metric spaces. 
\end{proof}

     Corollary \ref{locid}, via    \cref{cor_EM_popu},  offers sufficient conditions for local identifiability of the  softmax mixture model. The proof is constructive, and shows that any
    $\bomega'$ that is observationally equivalent to $\bomega^*$ in the stated $(\delta_0 /2)$ neighborhood must coincide with $\bomega^*$. 
     
     It is classically known \citep{Rothenberg1971} that under weak regularity conditions local identifiability is equivalent to non-singularity of the information matrix for general parametric families.  In the context of this paper, these conditions would therefore be relative to the mixture model. In contrast,  our results in \cref{cor_EM_popu,locid} on local identifiability under softmax mixtures rely on \cref{ass_X}, a more transparent condition that depends only on  the information matrix of  a {\em single} softmax component, rather than that of the {\em entire} mixture. Moreover, the bound on $\delta_0 $ in \eqref{cond_init_fix} provides an explicit quantification of the neighborhood within which local identifiability holds.


    \subsection{Convergence of the sample-level EM algorithm and rates of convergence:  generic design }\label{app_sec_EM_fix}
    The following theorem is the analogue of \cref{thm_EM} for the deterministic design. 

 \begin{theorem}[Convergence of the EM algorithm under the deterministic design]\label{thm_EM_fix}
For any  support points $\{x_1, \ldots, x_p\}$ of the softmax mixture,  
 let $\bomega^*$ satisfy  Assumptions \ref{ass_alpha}, \ref{separation} and \ref{ass_X}, with $\Delta$ satisfying \eqref{cond_sep}.     
         Assume \eqref{cond_N_explict} and the initialization  $\wh \bomega^{(0)} $ satisfies \eqref{cond_init_fix} in lieu of $\bomega^{(0)}$.   
     Then, there exist some $0<\phi <1$, some absolute constant $C''>0$  and step-sizes $\eta_k>0$, $k\in [K]$, such that with probability at least $1- \cO(N^{-L})$, the following holds for the whole sequence 
    $\wh \bomega^{(t)}$ in (\ref{iter_alpha}) -- (\ref{iter_theta}), with $t\ge 1$,  
    \begin{equation}\label{rate_EM_fix}
        d(\wh\bomega^{(t)}, \bomega^*) \le ~  \phi^{t-1} ~ \delta_0  +  C''\sqrt{\oa K L \log N \over \ua^2 N}.
    \end{equation}
	\end{theorem}  
\begin{proof} 
The  problem at hand is non-standard in that we are dealing with a  hybrid between the standard EM for $\balpha$ in step (\ref{iter_alpha}) and a first-order EM for $\btheta_k$, $k\in [K]$, in step (\ref{iter_theta}). 
We use induction to prove that with the desired probability, 
\begin{align}\label{update-EM}
		d( \wh \bomega^{(t)}, \bomega^*)  &~ \le~   	 \phi^{t} \delta_0 + {1 - \phi^{t} \over 1-\phi} \delta_N,\qquad \forall ~ t\ge 0,
\end{align}
for $\delta_N = \cO(\epsilon_N/\ua)$ with $\epsilon_N$ given in  \cref{lem_dev_EM} and for some $\phi \in (0,1)$ with $\delta_N \le (1-\phi)\delta_0$. 

It is easy to see \eqref{update-EM} holds for $t=0$ as  $d(\wh\bomega^{(0)} ,\bomega^*)\le \delta_0$. Suppose that \eqref{update-EM} holds for some arbitrary $t\in\NN$. We first note  that
$d(\wh \bomega^{(t)}, \bomega^*) \le \phi^t \delta_0 + (1-\phi^t)\delta_0 = \delta_0$ so that $\wh \bomega^{(t)} \in \cB(\bomega^*,\delta_0)$, the size-$\delta_0$ ball around $\bomega^*$ with respect to $d$ in \eqref{def_dist}. To establish \eqref{update-EM} for $t+1$,  we first  study the updates $\wh \alpha_k^{(t+1)} - \alpha_k^* = \wh M_{k}(\wh \bomega^{(t)}) - M_{k}(\bomega^*)$, where we recall $M_k(\cdot)$ and $\wh M_k(\cdot)$ from \eqref{iter_alpha_popu} and \eqref{iter_alpha}, respectively.
Since	\begin{align*} 
    |\wh M_{k}(\wh \bomega^{(t)}) - M_{k}(\bomega^*)|  \le \sup_{\bomega\in\cB(\bomega^*,\delta_0)}   |\wh M_{k}( \bomega ) - M_{k}(\bomega)|+  |M_{k}( \wh\bomega ^{(t)} ) - M_{k}(\bomega^*)|
	\end{align*}
    as $d(\wh\bomega^{(t)} ,\bomega^*)\le \delta_0$,
 \cref{lem_dev_EM,lem_GS_theta} imply that, for $\kappa$  given in \cref{lem_GS_theta},
 \[ 
    \|\wh {\bm \alpha}^{(t+1)} - {\bm \alpha}^*\|_\i \le C \epsilon_N + \kappa ~ d(\wh\bomega^{(t)}, \bomega^*)
\]
holds   with probability $1-\cO(N^{-L})$.
We analyze the first-order EM-updates $\wh \btheta_k^{(t+1)}$
in \eqref{iter_theta} as follows:	
\begin{align*} 
		\|\wh \btheta_k^{(t+1)} - \btheta_k^*\|_2 & = \|\wh \btheta^{(t)}_k + \eta_k \nabla_{\btheta_k} \wh Q(\wh \bomega^{(t)} \mid \wh \bomega^{(t)}) - \btheta_k^*\|_2\\\nonumber
		&\le \|\wh \btheta^{(t)}_k - \btheta_k^* + \eta_k \nabla_{\btheta_k} Q(\wh \bomega^{(t)} \mid \bomega^*)  \|_2  +  \eta_k \| \nabla_{\btheta_k} Q(\wh \bomega^{(t)} \mid \wh \bomega^{(t)}) -  \nabla_{\btheta_k} Q(\wh \bomega^{(t)} \mid  \bomega^{*}) \|_2\\\nonumber
		&\qquad + \eta_k \| \nabla_{\btheta_k} \wh Q(\wh \bomega^{(t)} \mid \wh \bomega^{(t)}) -  \nabla_{\btheta_k} Q(\wh \bomega^{(t)} \mid \wh \bomega^{(t)}) \|_2
	\end{align*}
Invoking \cref{lem_GS_theta,lem_dev_EM} gives that, with probability $1-\cO(N^{-L})$, 
\begin{align*} 
		\|\wh \btheta_k^{(t+1)} - \btheta_k^*\|_2 
		&\le \|\wh \btheta^{(t)}_k  - \btheta_k^* + \eta_k~  q(\wh \bomega^{(t)}  ) \|_2 +  \eta_k   \left( \os \kappa~  d(\wh \bomega^{(t)}, \bomega^*)    +  C  \os\epsilon_N\right).
	\end{align*}
Here, we write $q_k(\bomega):= \nabla_{\btheta_k} Q(\bomega \mid \bomega^*)$ with $q_k(  \bomega^*)=0$,  and its smoothness and strong-concavity  properties are stated in \cref{lem_oracle_Q_sandwich}. Let $\mu_k$ and $\gamma_k$ as in \eqref{def_mu_gamma} and set $\eta_k = {2/( \mu_k + \gamma_k )}$.
 After we square the first term on the right and work out the squares, we find
	\begin{align*}
		&\|\wh \btheta^{(t)}_k  - \btheta_k^* + \eta_k q_k ( \wh\bomega^{(t)} ) \|_2 ^2\\ 
		&= \|\wh \btheta^{(t)}_k - \btheta_k^*\|_2^2 + \eta_k^2 \| q_k(\wh \bomega^{(t)} )  \|_2^2 + 2 \eta_k (\wh\btheta_k^{(t)} - \btheta_k^*)^\T \left(q_k(\wh \bomega^{(t)}  ) - q_k( \bomega^* ) \right) \\ 
		&\le \left(
		1 - {2\eta_k\mu_k\gamma_k \over \mu_k + \gamma_k}
		\right)\|\wh \btheta^{(t)}_k - \btheta_k^*\|_2^2 + \eta_k\left( \eta_k - {2 \over \mu_k + \gamma_k}\right) \|q_k(\wh \bomega^{(t)}  )  \|_2^2 &&\text{by \cref{lem_oracle_Q_sandwich}}\\
        &\le \left({\mu_k-\gamma_k \over \mu_k+\gamma_k}\right)^2  \|\wh \btheta^{(t)}_k - \btheta_k^*\|_2^2
        &&\text{by } \eta_k = {2\over \mu_k + \gamma_k}.
	\end{align*}   
Summarizing, we find with probability $1-\cO(N^{-L})$ that
\begin{align*}
		d(\wh \bomega^{(t+1)}, \bomega^*) &\le      
		 C \max\left\{ {1\over \ua},~ \os^2 \max_k{ \eta_k}\right\}
     \epsilon_N   +  \phi ~  d(\wh \bomega^{(t)}, \bomega^*) 
	\end{align*}
	where
    \begin{align*}
      \phi &= \max_k { \mu_k-\gamma_k\over \mu_k+\gamma_k} +\kappa \max\left( {1\over \ua}, ~ \max_k {2\os^2\over \mu_k+\gamma_k} \right)\\
      &\le {(1+c_0) \os^2-(1-c_0)\us^2\over (1+c_0)\os^2 +(1-c_0)\us^2} + {2\os^2\over (1+c_0)\os^2 +(1-c_0)\us^2} {\kappa\over\ua}\\
      &<1- {2(1-2c_0)\us^2\over  (1+c_0)\os^2 +(1-c_0)\us^2} &&\text{since $\kappa<c_0 \ua\, {\us^2\over \os^2}$ by \eqref{cond_sep}}\\
      &<1 &&\text{since $c_0<1/2$}
 \end{align*}
Now by setting $\delta_N =  2C\epsilon_N  / \ua \le   (1-2c_0)  \delta_0 {\us^2} / \os^2
     \le (1-\phi) \delta_0$ by \eqref{cond_N_explict}, we obtain  
\[
    d(\wh \bomega^{(t+1)}, \bomega^*)  \le \delta_N +  \phi ~  d(\wh \bomega^{(t)}, \bomega^*) \le   \phi^{t+1} \delta_0 + {1 - \phi^{t+1} \over 1-\phi} \delta_N
\]
so that 
\eqref{update-EM} holds for $t+1$. This proves the induction step and
      the proof is complete.
 \end{proof}

The choice of $\eta_k = 2/(\mu_k+\gamma_k)$ is chosen for simplicity. Inspecting the proof reveals that for any $0<\eta_k\le 2/(\mu_k+\gamma_k)$, the same conclusion holds with the corresponding contraction factor $\phi$ changed to
        \[
            \phi =   \max\left\{{\kappa\over \ua}, \max_k \left(1 - {2 \mu_k\gamma_k\over \mu_k + \gamma_k}\eta_k + \kappa \os^2 \eta_k \right) \right\}\le \max\left\{{\kappa\over \ua},
            ~ 1- \min_{k\in [K]}{\mu_k\gamma_k \over \mu_k + \gamma_k}\eta_k   \right\}< 1,
        \] 
        ensured by the separation condition in \eqref{cond_sep}.

\begin{remark}\label{rem:tricky1}
We follow the road-map developed in 
    \cite{EM2017} for analyzing the EM algorithm for general mixture models.
Specifically, we   establish (a) the smoothness  and strong concavity of $\bomega \mapsto Q(\bomega \mid \bomega^*)$ and  (b) the Lipschitz continuity of $\bomega' \mapsto \nabla_{\btheta_k}Q(\bomega \mid \bomega')$ for all $\bomega$ in a local neighborhood of $\bomega^*$
and (c) the rate of convergence   of 
$\max_{k\in[K]}\|\nabla_{\btheta_k}\wh Q(\bomega \mid \bomega)-\nabla_{\btheta_k}Q(\bomega \mid \bomega)\|_2$ uniformly over $\bomega$ within a size $\delta_0$-neighborhood of $\bomega^*$. Although these are high-level quantities, as the authors noted in \cite{EM2017}, the real challenge in analyzing EM-type algorithms lies in establishing properties (a),  (b) and (c) under specific models. Their work demonstrates this framework for the standard 2-GMM  and two of its variants. To the best of our knowledge, a theoretical analysis of the EM algorithm under softmax mixture models has not yet been developed. 
The establishment of   properties (a), (b) and (c) 
 proves to be significantly more challenging under softmax mixture models than in the GMM setting, see Example \ref{examp_comp_2GMM} below.
Indeed, for property (a), the fact that  $Q(\bomega \mid \bomega^*)$ are quadratic in $\btheta_k$ under the GMM  implies that their gradient $\nabla_{\btheta_k} Q(\bomega \mid \bomega^*)$ is {\em linear} in $\btheta_k$. As a result, the strong concavity and smoothness of  $Q(\cdot \mid \bomega^*)$ with respect to $\btheta_k$ follows immediately. In stark contrast, $\nabla_{\btheta_k} Q(\bomega \mid \bomega^*)$ under the softmax mixture model is {\em non-linear} in $\btheta_k$, and its expression in \eqref{grad_QN}  still involves $A(x_j;\btheta_k)$ for $j\in [p]$. The strong concavity and smoothness of $Q(\cdot \mid \bomega^*)$ in this setting are established in \cref{lem_oracle_Q_sandwich} of \cref{app_sec_EM}, and require a careful perturbation analysis of several softmax-related quantities stated in \cref{lem_perturb,lem_hess_unif} of \cref{app_sec_lemmas_EM}.
The difficulty is further elevated when establishing property (b), which concerns the Lipschitz continuity of 
$\|\nabla_{\btheta_k}Q(\bomega \mid \bomega) - \nabla_{\btheta_k}Q(\bomega \mid \bomega^*)\|_2$, for all $\bomega$ within a $\delta_0$-neighborhood of $\bomega^*$. This step involves the most technically demanding derivations, even in the simple case of the symmetric and isotropic 2-GMM \citep{EM2017}, and extending the analysis to isotropic $K$-GMMs already requires substantial refinements \citep{yan2017convergence}. In \cref{examp_comp_2GMM} below, we illustrate that verifying property (b) for softmax mixtures -- even in the case $K = 2$ -- is significantly more challenging than for the 2-GMM. For general $K \ge 2$, property (b) is established in \cref{lem_GS_theta} of \cref{app_sec_EM}, building on several technical results presented in \cref{lem_perturb,lem_hess_unif,lem_perturb_hess,lem_coh} of \cref{app_sec_lemmas_EM}. Finally, since our EM algorithm employs a hybrid M-step to estimate both $\balpha^*$ and $\btheta_1^*, \ldots, \btheta_K^*$, an analogous version of property (b) must also be verified for the closed-form update of $\balpha^{(t)}$. This result is also stated in \cref{lem_GS_theta}. 

 Existing analyses of property (c)
 under GMMs typically rely on empirical process techniques such as symmetrization and Ledoux-Talagrand type contraction results \citep{EM2017, yan2017convergence, CHIME2019}. However, in the case of softmax mixture models, the Lipschitz conditions required for applying the Ledoux-Talagrand contraction are challenging to verify. 
 Instead, we develop a carefully tailored discretization argument to establish the necessary uniform convergence guarantees in \cref{lem_dev_EM} of  \cref{app_sec_EM}.  
\end{remark}

   In the following example, we illustrate the difficulty of verifying the Lipschitz continuity of the map $\bomega' \mapsto \nabla_{\btheta_k} Q(\bomega \mid \bomega')$ under softmax mixture models (\cref{lem_GS_theta}) by comparing it to the Gaussian mixture model case in an even simplified setting.
    \begin{Example}\label{examp_comp_2GMM}
    We focus the discussion on two mixture components with equal weights. Start  with a  2-GMM where the observable feature $Y \in \RR^p$ comes from 
    $\cN_p( (\eta/2) \btheta^*, \bI_p)$, conditioning on $\eta$,  with $\PP(\eta =1) = \PP(\eta=-1) = 1/2$. The only parameter is $\btheta^*\in \RR^p$,  with the separation between $Y \mid \eta = 1$ and $Y\mid \eta = -1$ being $\|\btheta^*\|_2^2$. The M-step of the population-level EM algorithm uses the operator $M$  given by 
        \[
           M( \btheta)  =  {\EE[\gamma(Y; \btheta) Y] \over \EE[\gamma(Y;\btheta)]}
        \]
    while evaluating 
        $
                \gamma(Y; \btheta) = 1  / (1 + \exp(-Y^\T \btheta))
        $
    is the E-step. Establishing its contraction requires deriving the Lipschitz continuity of $M$, which in turn hinges on bounding the difference $|\EE[\gamma(Y;\btheta) - \gamma(Y;\btheta^*)]| \le \kappa \|\btheta -\btheta^*\|_2$ for some small $\kappa$. Derivation of $\kappa$ is intuitively simple as 
        \[
            {d \gamma(Y;\btheta) \over d\btheta} = {\exp(-Y^\T\btheta) \over [1+\exp(-Y^\T\btheta)]^2} Y =  {\exp(-(\eta/2) \btheta^\T \btheta^* - W^\T \btheta) \over [1+\exp(-(\eta/2) \btheta^\T \btheta^* - W^\T \btheta)]^2}Y 
        \]
        for some $W\sim \cN_p(0,\bI_p)$.
        When $\btheta$ is close to $\btheta^*$, by $W^\T \btheta \sim \cN(0,\|\btheta\|_2^2)$,  the fraction in front of $Y$ can be bounded (in expectation) by $\exp(-c\|\btheta^*\|_2^2)$, which leads to $\kappa \le \exp(-c\|\btheta^*\|_2^2)$.

        Now consider the softmax mixtures with $K=2$, $\alpha_1 = \alpha_2 = 1/2$ and $\btheta_1^* = -\btheta_2^* =: \btheta^*$.  From \eqref{grad_QN}, bounding $\|\nabla_{\btheta_k}Q(\bomega \mid \bomega') - \nabla_{\btheta_k} Q(\bomega \mid \bomega^*) \|_2$ requires to bound the $\ell_2$-norm of 
        \[
             \sum_{j=1}^p \pi(x_j;\bomega^*) \left(   { A (x_j;\btheta') \over A (x_j;\btheta') + A (x_j;-\btheta') } - { A (x_j;\btheta^*) \over A (x_j;\btheta^*) + A (x_j;-\btheta^*) }\right) (x_j - \bX^\T A(\btheta_k))
        \]
        where, explicitly,
        \[
             { A (x_j;\btheta') \over A (x_j;\btheta') + A (x_j;-\btheta') }  = {1\over 1 + \exp(-2x_j^\T \btheta') {\sum_\ell \exp(x_\ell^\T\btheta') \over \sum_\ell \exp(-x_\ell^\T\btheta')}}.
        \]
        The derivative of the above term with respect to $\btheta'$ is notably complex, and even when ignoring the ratio involving the summations over $\ell$ in the denominator, deriving a Lipschitz constant in terms of $\exp(-c\|\btheta^*\|_2^2)$ remains highly non-trivial. This difficulty is further exacerbated when the mixing weights are unknown and the number of mixture components exceeds two, a case we address in \cref{lem_GS_theta} of \cref{app_sec_EM}.
    \end{Example}

    \subsection{ Proof of \cref{thm_EM}. Convergence of the sample-level EM algorithm and rates of convergence: random  design.} \label{app_proof_EM_random}
    \begin{proof}
        In view of \cref{thm_EM_fix}, it suffices to prove that conditions in \cref{thm_EM_fix} (fixed design) hold  under conditions in \cref{thm_EM} (random design) with probability at least $1-p^{-2}$. 
        
        To this end, for random features $X$ satisfying \cref{random-points},  the event
      \begin{equation}\label{def_event_E2}
    			\cE_2 = \left\{
    			\max_{j \in [p]} \|X_j\|_2 \le \bar\sigma \left( \sqrt{L} + \sqrt{2(s+1)\log(p)} \right)
    			\right\}
    		\end{equation}
          holds with $\mu$-probability at least $1-p^{-s}$ for all $s\ge 2$.
          See, for instance, \cref{lem_quad}. This also means that $\cE_2$ holds $\mu$-almost surely by the Borel-Cantelli lemma. This fact will be used throughout the proofs.

        We proceed to verify \cref{ass_X} under Assumptions \ref{ass_theta}, \ref{random-points} and \ref{ass_Htheta_mu}. Regarding $\os^2$ and $\us^2$, by invoking \cref{lem_hess_subG}, one has that with $\mu$-probability at least $1-4p^{-3}$, 
        \[
             \us_\mu^2/2 \le  \us^2 \le \os^2 \le C\os_\mu^2
        \]
        for $p \ge p_0(B,\os_\mu)$ with the expression of $p_0$ given in \eqref{lb_p}. To derive a bound for $\vs^2$, we first bound from above 
        \[
            \vs_\mu^2 := \lambda_1\left(
        [ H_\btheta^{(\mu)} ]^{-1/2} {\EE_\mu[XX^\T \exp(X^\T\btheta)] \over \EE_\mu[\exp(X^\T\btheta)]}  [ H_\btheta^{(\mu)} ]^{-1/2} \right).
        \]
        \cref{random-points} and   Jensen's inequality yields $\EE_\mu[\exp(X^\T\btheta)] \ge 1$. By the Cauchy-Schwarz inequality together with   \cref{ass_theta,ass_Htheta_mu}, one has  
\begin{align*}
      \vs_\mu^2 ~ &\le {1\over \us_\mu^2} \lambda_1\left( \EE_\mu[XX^\T \exp(X^\T\btheta)]   \right)\\  &\le {1\over \us_\mu^2} \sup_{\|v\|_2=1}\sqrt{\EE_\mu [(X^\T v)^4]}\sqrt{\EE_\mu[\exp(2X^\T\btheta)]}\\&\le C{\os_\mu^2\over \us_\mu^2}\exp(2\sigma_\mu^2 B^2).
\end{align*}
    By invoking \cref{lem_I_op,lem_II} together with \cref{lem_expectations_subG}, one can deduce that  $\vs^2 \le C' \vs^2_\mu$ with $\mu$-probability at least $1-4p^{-3}$ for $p\ge p_0.$ This verifies \cref{ass_X} under conditions in \cref{thm_EM}.

    Finally, with $\mu$-probability $1-p^{-2}$, the initialization condition  \eqref{cond_init_rand} with 
    \begin{equation}\label{def_c_mu}
        c_\mu = {c_0\over C}{\os_\mu^2\over \us_\mu^2}\exp(-2\os_\mu^2B^2)
    \end{equation}
    implies \eqref{cond_init_fix}. 
    Moreover, conditions \eqref{cond_sep_rand} and \eqref{cond_N_rand} imply the ones of \eqref{cond_sep} and \eqref{cond_N_explict} under the fixed design, respectively. Invoking \cref{thm_EM_fix} therefore completes the proof. 
    \end{proof}

 \subsection{Key lemmas for the proof of Theorem \ref{thm_EM_fix}} 

    \noindent {\bf Notation.}
    For any $\bomega^* =(\balpha^*,\btheta_1^*,\ldots,\btheta_K^*)$, recall that 
    $
        \Delta = \min_{k\ne k'} \|\btheta_k^*-\btheta_{k'}^*\|_2.
    $
     For any $\bomega = (\balpha, \btheta_1, \ldots, \btheta_K)$, we write for each $j\in[p]$,
    \[
        A_{\btheta_k}(x_j) = A(x_j; \btheta_k),
    \qquad  \pi_{\bomega}(x_j) =  \pi(x_j;\bomega) = \sum_{k=1}^K \alpha_k A_{\btheta_k}(x_j).
    \]
     For any $\btheta \in \RR^L$ with $A_\btheta=(A_\btheta(x_1),\ldots,A_{\btheta}(x_p))^\T \in \Delta^{p-1}$, write
	$
	\Sigma_{A_\btheta}  :=  \diag(A_\btheta)  - A_\btheta A_\btheta^\T.
	$    
    Further let 
	\begin{equation}\label{def_N_I_II}
		\begin{split}
			&N_\btheta = \sum_{j=1}^p e^{x_j^\T \btheta} \in \RR,\quad
			\rI_\btheta = \sum_{j=1}^p e^{x_j^\T \btheta} x_j x_j^\T\in \RR^{L\times L},\quad
			\rII_\btheta = \sum_{j=1}^p e^{x_j^\T \btheta} x_j   \in \RR^L
		\end{split}
	\end{equation}
	and note that
	\begin{equation}\label{def_H}
		H_\btheta := \bX^\T \Sigma_{A_{\btheta}} \bX =  {\rI_\btheta \over N_\btheta} - {\rII_\btheta \rII_\btheta^\T \over N_\btheta^2}.
	\end{equation}
 We let $\cB(v;r)$ denote the $\ell_2$-ball in $\RR^L$ centered at $v$ with radius $r$.
The following are non-trivial results that establish 
strong concavity and local smoothness of the function $q_k(\bomega)= \nabla_{\btheta_k}Q(\bomega \mid \bomega^*)
$, 
  smoothness of $\bomega'\mapsto \nabla_{\btheta_k}Q(\bomega \mid \bomega')$,  Lipschitz continuity of $M_{k}(\bomega)$ and
  maximal inequalities 
  for 
 $ | \wh M_k(\bomega) -  M_k(\bomega)| $ and 
 $\| \nabla_{\btheta_k} \wh Q(\bomega  \mid \bomega)  - \nabla_{\btheta_k} Q(\bomega  \mid \bomega) \|_2$, uniformly over $\cB_d(\bomega^*, \delta_0)$.
The proofs are rather involved and can be found in separate sections below.

	
	\begin{lemma}\label{lem_oracle_Q_sandwich}
    Under \cref{ass_X} and \eqref{cond_init_fix}, for all $ k \in [K]$, we set 
    \begin{equation}\label{def_mu_gamma}
        \gamma_k = {(1-c_0)}\alpha_k^* \us^2,\qquad 
        \mu_k = {(1+c_0)}\alpha_k^* \os^2
    \end{equation}
    with $c_0$ specified in (\ref{cond_init_fix}). For any  $\bomega,\bomega'$ and $k\in [K]$, if $\btheta_k,\btheta_k'\in \cB(\btheta_k^*;\delta_0/\os)$, then
		\begin{align}\label{Q_concavity}
			(\btheta_k -\btheta_k')^\T \bigl(q_{k} (\bomega  ) - q_{k} (\bomega' )\bigr)    &\le -
   \gamma_k 
   \|\btheta_k - \btheta_k'\|_2^2\\\label{Q_smoothness}
			\|q_k (\bomega ) - q_k (\bomega' )\|_2    &\le 
   \mu_k
   \|\btheta_k - \btheta_k'\|_2
		\end{align}
	and 
		\begin{align}\label{Q_concavity_stronger} 
			& (\btheta_k -\btheta_k')^\T \left(q_k(\bomega ) - q_k(\bomega' )\right)  \le -
   { \mu_k\gamma_k \over \mu_k + \gamma_k} \|\btheta_k - \btheta_k'\|_2^2 - { 1\over \mu_k + \gamma_k}  \|q_k(\bomega) - q_k(\bomega' )\|_2^2.
		\end{align}
	\end{lemma}
    \begin{proof}
        See \cref{app_sec_proof_lem_oracle_Q_sandwich}.
    \end{proof}

  
    \begin{lemma}\label{lem_GS_theta}
 Grant \cref{ass_alpha,separation,ass_X} and \eqref{cond_init_fix}.  
    For any  $ \bomega  \in \cB_d(\bomega^*, \delta_0)$, \begin{align}\label{cond_GS_M}
			\max_{k\in [K]} | M_{k}(\bomega) -  M_{k}(\bomega^{*})|   &\le \kappa ~ d(\bomega, \bomega^*)\\\label{cond_GS_theta}
            \max_{k\in [K]}\| \nabla_{\btheta_k} Q(\bomega \mid \bomega ) - \nabla_{\btheta_k} Q(\bomega \mid \bomega^{*}) \|_2 &\le \os~  \kappa   ~ d(\bomega, \bomega^*)
		\end{align}
    where   for some large absolute constant $C>0$,  
	\begin{equation}\label{def_kappa}
		\kappa = C K\oa (1+\os^2 \Delta^2) \exp(-\us^2 \Delta^2 /8).
	\end{equation} 
	\end{lemma} 
    \begin{proof}
        See \cref{app_sec_proof_lem_GS_theta}.  
    \end{proof}


    \begin{lemma}\label{lem_dev_EM}
	Grant \cref{ass_alpha,ass_X} and conditions \eqref{cond_init_fix} \& \eqref{cond_N_explict}.  Set 
    $$
    \epsilon_N  = \sqrt{\oa KL\log(N) \over  N}.
    $$ 
	There exists some absolute constant $C>0$ such that with probability at least  $1- \cO(N^{-L})$, 
    \begin{align*}
		  \sup_{\bomega\in \cB_d(\bomega^*,\delta_0)} \max_{k\in [K]} \left| \wh M_k(\bomega) -  M_{k}(\bomega)\right|    &  \le    C ~ \epsilon_N,\\
	 \sup_{\bomega\in \cB_d(\bomega^*,\delta_0)}	\max_{k\in [K]} \left\| \nabla_{\btheta_k} \wh Q(\bomega  \mid \bomega)  - \nabla_{\btheta_k} Q(\bomega  \mid \bomega) \right\|_2   & \le C ~ \os ~ \epsilon_N.
		\end{align*}
	\end{lemma}
    \begin{proof}
        See \cref{app_sec_proof_lem_dev_EM}.
    \end{proof}

    \subsection{Proofs of \cref{lem_oracle_Q_sandwich,lem_GS_theta,lem_dev_EM}}\label{app_sec_lemmas_EM}

    To prove \cref{lem_oracle_Q_sandwich,lem_GS_theta,lem_dev_EM}, we first state and prove a few technical lemmas.

    \subsubsection{Technical lemmas used to prove \cref{lem_oracle_Q_sandwich,lem_GS_theta,lem_dev_EM}}
	
	The following lemma proves certain Lipschitz continuity of the function $A_\btheta$ and $\pi_\bomega$ relative to changes in   $\bomega=(\balpha,\btheta_1,\ldots,\btheta_K)$.

	\begin{lemma}\label{lem_perturb}
		For any $\bomega,\bomega'\in \Omega$ such that  
		\begin{equation} \label{cond_epsilon_theta} 
			\max_{k\in [K]} \|\btheta_k - \btheta_k'\|_2 \le  {1\over 2\|\bX\|_{\i, 2}},
		\end{equation} 
		we have, for any $j\in [p]$ and $k\in [K]$,  
		\begin{align}\label{bd_perturb_A}
			&{ |A_{\btheta_k}(x_j) - A_{\btheta_k'}(x_j)|  \over A_{\btheta_k}(x_j) \wedge  A_{\btheta_k'}(x_j)} \le   2 \|\bX\|_{\i, 2}  \max_{k\in [K]} \|\btheta_k - \btheta_k'\|_2.
		\end{align}
        Furthermore, if 
        \begin{equation}\label{cond_epsilon_alpha}
			\|\balpha-\balpha'\|_\i \le  {1\over 3} \min_{k\in [K]}\alpha_k,
        \end{equation}
        holds additionally, then  
        \begin{equation}\label{bd_perturb_pi}
			 {\left| \pi_{\bomega}(x_j) - \pi_{\bomega'}(x_j)
				\right| \over \pi_{\bomega}(x_j)} \le  
			\max_{k\in [K]}  {\|\balpha-\balpha'\|_\i \over  \alpha_k}   + 3  \|\bX\|_{\i,2}  \max_{k\in [K]} \|\btheta_k - \btheta_k'\|_2.
        \end{equation}
	\end{lemma}
	\begin{proof}
		We first prove \eqref{bd_perturb_A}. Pick any $k\in [K]$ and $j\in [p]$. By definition, we have 
		\begin{equation}\label{eq_A_j_theta_diff}
			\begin{split}
				A_{\btheta_k}(x_j) -A_{\btheta_k'}(x_j) &= 
				{1\over \sum_{\ell=1}^p e^{(x_\ell - x_j)^\T \btheta_k}} - {1\over \sum_{\ell=1}^p e^{(x_\ell - x_j)^\T \btheta_k'}} \\
				& =  {1\over \sum_{\ell=1}^p e^{(x_\ell - x_j)^\T \btheta_k}} {\sum_{\ell=1}^p e^{(x_\ell - x_j)^\T \btheta_k'} \left[1 - e^{(x_\ell - x_j)^\T (\btheta_k - \btheta_k')}\right] \over \sum_{\ell=1}^p e^{(x_\ell - x_j)^\T \btheta_k'}}\\
				&\le  A_{\btheta_k}(x_j) ~   \max_{\ell \in [p]} \left|
				1 - e^{(x_\ell - x_j)^\T (\btheta_k - \btheta_k')}
				\right| 
			\end{split}
		\end{equation}
		This bound, the inequality 
		\begin{equation*}
			(x_\ell - x_j)^\T (\btheta_k - \btheta_k') \le 2\|\bX\|_{\i,2}  \max_{k\in [K]} \|\btheta_k - \btheta_k'\|_2 \overset{\eqref{cond_epsilon_theta}}{\le}  1,
		\end{equation*}
		and the basic inequality $|1-e^t| \le 2 |t|$ for any $|t|\le 1$ combined give that
		\begin{equation*}
			A_{\btheta_k}(x_j) -A_{\btheta_k'}(x_j) \le   
			2 A_{\btheta_k}(x_j) ~   \|\bX\|_{\i,2} \max_{k\in [K]} \|\btheta_k - \btheta_k'\|_2.
		\end{equation*}
		The same arguments hold after we swap  $\btheta_k$ and $\btheta_k'$, and \eqref{bd_perturb_A} follows.

		To prove \eqref{bd_perturb_pi}, we have that ${|\pi_{\bomega}(x_j) - \pi_{\bomega'}(x_j)|/ \pi_{\bomega}(x_j)}$ is bounded from above by
		\begin{align*}
			&    {1\over \pi_{\bomega}(x_j)}\sum_{k=1}^K \left(|\alpha_k - \alpha_{k'}| A_{\btheta_k}(x_j) + \alpha_{k}' \left|A_{\btheta_k}(x_j) -A_{\btheta_k'}(x_j)\right| \right)\\
			&\le  \sum_{k=1}^K \left({	\|\balpha-\balpha'\|_\i \over \alpha_k} {\alpha_k A_{\btheta_k}(x_j) \over \pi_{\bomega}(x_j)}  + {\alpha_{k}' \over \alpha_k} {\alpha_k A_{\btheta_k}(x_j) \over \pi_{\bomega}(x_j)} 2  \|\bX\|_{\i,2}  \max_{k\in [K]} \|\btheta_k - \btheta_k'\|_2 \right) &&\text{by \eqref{bd_perturb_A}}\\
			&\le  \max_{k\in [K]} \left({	\|\balpha-\balpha'\|_\i \over  \alpha_k}   + {3 \alpha_k' \over \alpha_k}    \|\bX\|_{\i,2} \max_{k\in [K]} \|\btheta_k - \btheta_k'\|_2 \right).
		\end{align*}
        The bound \eqref{bd_perturb_pi} follows   by invoking   \eqref{cond_epsilon_alpha} in the above display. 
	\end{proof}

    The following lemma controls the eigenvalues of $H_{\btheta_k}$ defined in \eqref{def_H} for all $\btheta_k \in  
	\cB(\btheta_k^*; \delta_0/\os)$ with $\delta_0$ satisfying  \eqref{cond_init_fix} and $k\in [K]$. 
   
	\begin{lemma}\label{lem_hess_unif}
        Fix any $\btheta^*\in \RR^L$ and any $\delta_0$ satisfying \eqref{cond_init_fix}. Under  \cref{ass_X}, 
       there exists some constant $c = c(c_0)\in (0,1)$ such that  for all $k\in [K]$,
    \begin{equation}\label{def_event_H_local}
    		(1-c)\us^2 \le	\lambda_L(H_{\btheta_k}) \le \lambda_1(H_{\btheta_k}) \le (1+c) \os^2,\qquad \forall ~  \btheta_k \in \cB(\btheta_k^*; \delta_0/\os)
    	\end{equation} 
	\end{lemma}
	\begin{proof} 
        Fix $\btheta^* \in \{\btheta_1^*,\ldots,\btheta_k^*\}$.
        Write $H_{\btheta^*}^{1/2}$ as the matrix square root of $H_{\btheta^*}=\bX^\T \Sigma_{A_{\btheta^*}} \bX$. 
		Let  $\btheta \in \cB(\btheta^*, \delta_0/\os)$ be arbitrary. We first bound  from above
		\begin{align*}
			\|H_{\btheta^*}^{-1/2}( H_\btheta - H_{\btheta^*})H_{\btheta^*}^{-1/2}\|_\op &= \sup_{v \in \bS^{L-1}} v^\T H_{\btheta^*}^{-1/2} (H_\btheta - H_{\btheta^*})H_{\btheta^*}^{-1/2}  v\\
                &\le  \sup_{v \in \bS^{L-1}} \left ( R_1(v) + R_2(v) \right)
		\end{align*}
		with 
        \begin{align*}
             R_1(v) &:=  \sum_{j=1}^p (x_j^\T H_{\btheta^*}^{-1/2} v)^2 \left(
		A_{\btheta^*}(x_j) - A_{\btheta}(x_j)
		\right),\\
        R_2(v) &:=  \left| \sum_{j=1}^p x_j^\T H_{\btheta^*}^{-1/2} v  \left(A_{\btheta^*}(x_j) + A_{\btheta}(x_j)\right) \right| ~  \left|
		\sum_{j=1}^p x_j^\T H_{\btheta^*}^{-1/2} v  \left(A_{\btheta^*}(x_j) - A_{\btheta}(x_j) \right) 
		\right|.
        \end{align*} 
		We observe that   $\btheta \in \cB(\btheta^*; \delta_0/\os)$  implies $ \|\btheta^*- \btheta\|_2 \le \delta_0 / \os$. After we invoke \eqref{bd_perturb_A} in \cref{lem_perturb} with $K=1$ and $\btheta^*$ and $\btheta$ in lieu of $\btheta_k$ and $\btheta_k'$, respectively, we find 
		\begin{align}\label{bd_R1}  
			\sup_{v \in \bS^{L-1}} R_1(v) & \le   \sup_{v \in \bS^{L-1}} \sum_{j=1}^p (x_j^\T H_{\btheta^*}^{-1/2} v)^2   A_{\btheta^*}(x_j) ~  3 (\delta_0/\os) \|\bX\|_{\i,2}   \\ 
			&\le  {3c_0 \over \vs^2}  \| H_{\btheta^*}^{-1/2} \bX^\T \diag(A_{\btheta^*}) \bX H_{\btheta^*}^{-1/2} \|_\op &&\text{by \eqref{cond_init_fix}}\nonumber\\  
			&\le  3c_0  &&\text{by \cref{ass_X}}. \nonumber
		\end{align}  
		Next, we observe that 
		\begin{align*}
			R_2 (v)  &\le ~  2 \left| \sum_{j=1}^p (x_j^\T H_{\btheta^*}^{-1/2} v) A_{\btheta^*}(x_j) \right| \left|
			\sum_{j=1}^p (x_j^\T H_{\btheta^*}^{-1/2} v) (A_{\btheta^*}(x_j) - A_{\btheta}(x_j))
			\right| \\
   &\qquad +  \left|
			\sum_{j=1}^p (x_j^\T H_{\btheta^*}^{-1/2} v) (A_{\btheta^*}(x_j) - A_{\btheta}(x_j))
			\right|^2.
		\end{align*}
		By repeating the arguments in \eqref{bd_R1}, we find that
		\begin{align} \label{bd_E_A_diff_ell_2}\nonumber
			 \sum_{j=1}^p (x_j^\T H_{\btheta^*}^{-1/2}  v) (A_{\btheta^*}(x_j) - A_{\btheta}(x_j))  & \le 3 (\delta_0/\os)\|\bX\|_{\i, 2}  \sum_{j=1}^p \left|x_j^\T H_{\btheta^*}^{-1/2}  v\right| A_{\btheta^*}(x_j)\\\nonumber
			&\le  {3   c_0 \over \vs^2} \Bigl(\sum_{j=1}^p (x_j^\T H_{\btheta^*}^{-1/2}  v)^2 A_{\btheta^*}(x_j)\Bigr)^{1/2}  \Bigl(\sum_{j=1}^p A_{\btheta^*}(x_j)\Bigr)^{1/2}\\\nonumber
			&\le {3   c_0 \over \vs}   \sqrt{\|H_{\btheta^*}^{-1/2}  \bX^\T \diag(A_{\btheta^*}) \bX H_{\btheta^*}^{-1/2} \|_\op}\\
            &\le  3 c_0
		\end{align}
		so that
		\begin{align*}
		\sup_{v \in \bS^{L-1}}	R_2(v)  &\le {6 c_0  \over \vs}\sup_{v \in \bS^{L-1}} \sum_{j=1}^p \left|x_j^\T H_{\btheta^*}^{-1/2}  v\right| A_{\btheta^*}(x_j) +  9 c_0^2 
			\le 3(2c_0  + 3c_0 ^2) . 
		\end{align*} 
		Combination of the bounds for $R_1(v)$ and $R_2(v)$, uniformly over $v\in \bS^{L-1}$ yields
		\[
		\|H_{\btheta^*}^{-1/2}( H_\btheta - H_{\btheta^*})H_{\btheta^*}^{-1/2}\|_\op  \le 9c_0(1+c_0).
		\]
        Together with   Weyl's inequality, the eigenvalues of $H_{\btheta^*}^{-1/2} H_\btheta H_{\btheta^*}^{-1/2}$ satisfy 
        \[
            \left| 1 - \lambda_\ell( H_{\btheta^*}^{-1/2} H_\btheta H_{\btheta^*}^{-1/2}) \right| \le 9c_0(1+c_0),\qquad \forall~ 1\le \ell \le L.
        \]
        In particular, 
        \[
        \left(1-9c_0-9c_0^2\right) \lambda_L(H_{\btheta^*})  \le   \lambda_L(H_\btheta) \le \lambda_1(H_\btheta) \le \lambda_1(H_{\btheta^*}) \left(1+9c_0+9c_0^2\right)
        \]
        which completes the proof. 
    \end{proof}

    The following two lemmas are crucial to  the proof of \cref{lem_GS_theta}.  For any  $a,k\in [K]$, let $\bar\btheta_{ak}^*$ be the midpoint of $\btheta_a^*$ and $\btheta_k^*$
    \begin{equation}\label{def_mid_theta}
        \bar \btheta_{ak}^* := {1\over 2}(\btheta_a^* + \btheta_k^*).
    \end{equation}

    \begin{lemma}\label{lem_perturb_hess}
    Grant  \cref{ass_X} and pick any  $a,k\in [K]$. For any $\bomega$ with $\btheta_k\in \cB(\btheta_k^*, \delta_0/\os)$ with $\delta_0$ satisfying \eqref{cond_init_fix},   we have 	
    \[
            \left\|  \bX^\T \bigl(\diag(A_{\bar\btheta_{ak}^*}) - A_{\btheta_k}A_{\btheta_k}^\T \bigr)\bX  \right\|_\op \lesssim ~ \os^2 +  \os^4 \| \btheta_a^*  -    \btheta_k^*\|_2^2.
		\]
	\end{lemma}
    \begin{proof}
		For simplicity, let us write $\bar\btheta^* = \bar\btheta_{ak}^*$. Using the notation in  \eqref{def_N_I_II}, it suffices to analyze
		\begin{align*}
			&{1\over \sum_{\ell =1}^p e^{x_{\ell}^\T \bar \btheta^*}} \sup_{v\in \bS^{L-1}}   \sum_{j=1}^p     e^{x_j^\T \bar\btheta^*} 
			v^\T \bX^\T (\be_j - A_{\btheta_k})  
			(\be_j - A_{\btheta_k})^\T \bX v \\ 
			&= {1\over N_{\bar \btheta^*}}\sup_{v\in \bS^{L-1}}  \left\{
			\sum_{j=1}^p e^{x_j^\T \bar\btheta^*} (x_j^\T v)^2 + 	\sum_{j=1}^p e^{x_j^\T \bar \btheta^*}(v^\T \bX^\T A_{\btheta_k})^2 - 2 \sum_{j=1}^p e^{x_j^\T \bar \btheta^*} x_j^\T v (v^\T \bX^\T A_{\btheta_k})
			\right\}\\ 
            &= \sup_{v\in \bS^{L-1}}   
			 \left\{ v^\T H_{\bar \btheta^*} v  + 	 v^\T \bX^\T (A_{\btheta_k} - A_{\bar\btheta^*})(A_{\btheta_k} - A_{\bar\btheta^*})^\T \bX v  
			\right\}\\
            &\le \lambda_1(H_{\bar \btheta^*}) + 2\|\bX^\T (A_{\btheta_k^*} - A_{\bar\btheta^*})\|_2^2+2\|\bX^\T (A_{\btheta_k} - A_{\btheta_k^*})\|_2^2
		\end{align*}
		By repeating the arguments in the proof of \eqref{bd_square_grad}, we obtain
        \begin{align*}
        \|\bX^\T (A_{\btheta_k^*} - A_{\bar\btheta^*})\|_2 &~ \le ~   \|\btheta_k^* - \bar\btheta^*\|_2 \sup_{u\in[0,1]}\lambda_1(H_{\bar\btheta_{u}^*}) \\
       \|\bX^\T (A_{\btheta_k} - A_{\btheta_k^*})\|_2 &~ \le ~ \|\btheta_k - \btheta_k^*\|_2 \sup_{u\in[0,1]}\lambda_1(H_{\btheta_{k,u}})
        \end{align*}
        where we write $\bar\btheta_{u}^* = u \btheta_a^* + (1-u)\btheta_k^*$ and  $\btheta_{k,u} = u \btheta_k^* + (1-u)\btheta_k \in \cB(\btheta_k^*,\delta_0/\os)$. The proof is completed by invoking  \cref{ass_X} and  \eqref{def_event_H_local}.
	\end{proof}

    Recall that $\btheta_{k,u} = u\btheta_k + (1-u)\btheta_{k}^*$ for any $u\in [0,1]$ and $k\in [K]$.

    \begin{lemma}\label{lem_coh}
        Let  $\bomega \in \cB_d(\bomega^*, \delta_0)$ with $\delta_0$ satisfying \eqref{cond_init_fix}. Under \cref{ass_X}, the following holds for any $a,k\in [K]$ with $a\ne k$,
        \[    
        \max_{j\in [p]} \sup_{u\in [0,1]}	{\alpha_a^*\alpha_k^*\over A_{\bar \btheta_{ak}^*}(x_j)}  {A_{\btheta_a^*}(x_j) A_{\btheta_{k,u}}(x_j)  \over \pi_{\bomega}(x_j)}
     \lesssim (\alpha_a^* \vee \alpha_k^*) \exp\left(
    		-{\us^2\over 8}\|\btheta_a^*-\btheta_k^*\|_2^2\right).
		\] 
        For any $\bomega$ with $\btheta_k\in \cB(\btheta_k^*,\delta_0/\os)$ for all $k\in [K]$,  the bound above gets replaced by
        \[    
            {\max_k\alpha_k^* \over \min_k \alpha_k} (\alpha_a^* \vee \alpha_k^*) \exp\left(
        		-{\us^2\over 8}\|\btheta_a^*-\btheta_k^*\|_2^2\right).
		\] 
	\end{lemma}

        \begin{proof}
		We first prove for $\bomega \in \cB_d(\bomega^*,\delta_0)$ which ensures  $\|\btheta_{k,u} - \btheta_k^*\|_2 \le \delta_0/\os $ and $\|\balpha - \balpha^*\|_\i \le \delta_0\ua$. Under condition \eqref{cond_init_fix}, after invoking \eqref{bd_perturb_A} and \eqref{bd_perturb_pi} of \cref{lem_perturb}, we obtain    
		\begin{align}\label{bd_ratio_Api}
			{A_{\btheta_a^*}(x_j) A_{\btheta_{k,u}}(x_j)  \over \pi_{\bomega}(x_j)} & \le {A_{\btheta_a^*}(x_j)A_{\btheta_k^*}(x_j)  (1 + 3\delta_0 \|\bX\|_{\i, 2}/\os) \over \pi_{\bomega^*}(x_j) (1 - \delta_0  - 4\delta_0 \|\bX\|_{\i,2}/\os)} \overset{\eqref{cond_init_fix}}{\lesssim}   {A_{\btheta_a^*}(x_j) A_{\btheta_k^*}(x_j)    \over \pi_{\bomega^*}(x_j)}.
		\end{align}
		Since 
		$A_{\btheta_k^*}(x_j)    = e^{x_j^\T \btheta_k^*} / N_{\btheta_k^*}$, we further obtain
		\begin{align*}
			\alpha_a^*\alpha_k^*  {A_{\btheta_a^*}(x_j) A_{\btheta_k^*}(x_j)    \over \pi_{\bomega^*}(x_j)} 
			&= {	\alpha_a^*\alpha_k^*  e^{x_j^\T (\btheta_a^*+\btheta_k^*) }/ (N_{\btheta_k^*}  N_{\btheta_a^*}) \over 
				\alpha_a^* e^{x_j^\T \btheta_a^*} / N_{\btheta_a^*}  + \alpha_k^* e^{x_j^\T \btheta_k^*} / N_{\btheta_k^*}  + \sum_{b\ne a,k} \alpha_b^* e^{x_j^\T \btheta_b^*}/ N_{\btheta_b^*} }\\
			&\le {	\alpha_a^*\alpha_k^*  e^{x_j^\T (\btheta_a^*+\btheta_k^*) }/ (N_{\btheta_k^*}  N_{\btheta_a^*}) \over 
				\alpha_a^* e^{x_j^\T \btheta_a^*} / N_{\btheta_a^*}  + \alpha_k^* e^{x_j^\T \btheta_k^*} / N_{\btheta_k^*}}\\
			& \le  {(\alpha_a^* \vee \alpha_k^* )~	\exp\left(
				x_j^\T (\btheta_a^*+\btheta_k^*) / 2 \right)    \over 
				N_{\btheta_k^*} \exp\left( x_j^\T (\btheta_a^*-\btheta_k^*)/2\right)   +   N_{\btheta_a^*}\exp\left(- x_j^\T (\btheta_a^*-\btheta_k^*)/2\right)}\\
            &\le (\alpha_a^* \vee \alpha_k^* )  {	\exp(
				x_j^\T \bar\btheta_{ak}^* )    \over 
				2  \sqrt{N_{\btheta_k^*} N_{\btheta_a^*}}}
		\end{align*} 
        so that 
        \begin{equation}\label{bd_cumu_gen}
            {\alpha_a^*\alpha_k^*\over A_{\bar \btheta_{ak}^*}(x_j)}  {A_{\btheta_a^*}(x_j) A_{\btheta_{k,u}}(x_j)  \over \pi_{\bomega}(x_j)} \le {\alpha_a^* \vee \alpha_k^* \over 2} {	 N_{\bar \btheta_{ak}^*}   \over \sqrt{N_{\btheta_k^*}N_{\btheta_a^*}}}.
        \end{equation}
        It remains to bound from above 
        \[
            \log N_{\bar \btheta_{ak}^*} - {1\over 2}\left(\log  N_{\btheta_k^*} + \log N_{\btheta_a^*}\right).
        \]
        By letting $g(\btheta) = \log N_{\btheta} = \log (\sum_{j=1}^p e^{x_j^\T \btheta})$ for any $\btheta \in \RR^L$, if there exists some $\nu >0$ such that $g(\btheta)$ is strongly $\nu$-convex over $u\btheta_a^* + (1-u)\btheta_k^*$ for all $u\in [0,1]$, then 
        \[
            \log   N_{\bar \btheta_{ak}^*}  - {1\over 2}\left(\log  N_{\btheta_k^*} + \log N_{\btheta_a^*}\right) = g(\bar \btheta_{ak}^*) - {1\over 2}\left[ g(\btheta_a^*) + g(\btheta_k^*)\right] \le -{\nu \over 8}\|\btheta_a^*-\btheta_k^*\|_2^2
        \]
        which yields the desired result. To verify the strongly $\nu$-convexity of $g(\btheta)$, taking the derivative with respect to $\btheta$ twice and interchanging the expectation with derivatives give  
        \begin{align*}
            \nabla^2 g(\btheta) = {\sum_{j=1}^p x_j x_j^\T e^{x_j^\T \btheta} \over \sum_{\ell=1}^p e^{x_\ell^\T \btheta}} - {(\sum_{j=1}^p x_je^{x_j^\T \btheta})(\sum_{j=1}^p x_je^{x_j^\T \btheta})^\T \over (\sum_{\ell=1}^p e^{x_\ell^\T \btheta})^2} =  H_{\btheta}.
        \end{align*}
        By \cref{ass_X}, we know that $\lambda_L(H_{\btheta}) \ge \us^2$ for all $\btheta =u\btheta_a^* + (1-u)\btheta_k^*$. This implies the strong $\us^2$-convexity hence completes the proof for any $\bomega \in \cB_d(\bomega^*,\delta_0)$. 

        When we only have $\btheta_k\in \cB(\btheta_k^*,\delta_0/\os)$ for all $k\in [K]$, the only change is in  \eqref{bd_ratio_Api} where invoke \eqref{bd_perturb_A} and \eqref{bd_perturb_pi} of \cref{lem_perturb} to obtain 
        \begin{align*}
            {A_{\btheta_a^*}(x_j) A_{\btheta_{k,u}}(x_j)  \over \pi_{\bomega}(x_j)}  & ~ \le~  {2\max_k\alpha_k^*\over \min_k \alpha_k}{A_{\btheta_a^*}(x_j)A_{\btheta_k^*}(x_j)  (1 + 3\delta_0 \|\bX\|_{\i, 2}/\os) \over \pi_{\bomega^*}(x_j) (1 - 2\delta_0 \|\bX\|_{\i,2}/\os)}\\
            &~ \overset{\eqref{cond_init_fix}}{\lesssim} ~   {\max_k\alpha_k^*\over \min_k \alpha_k}{A_{\btheta_a^*}(x_j) A_{\btheta_k^*}(x_j)    \over \pi_{\bomega^*}(x_j)}.
        \end{align*}
        This completes the proof. 
	\end{proof}

	\subsubsection{Proof of \cref{lem_oracle_Q_sandwich}: strong concavity and smoothness of the gradient function $q(\cdot) = \nabla_{\theta_k} Q(\cdot \mid \omega^*)$}\label{app_sec_proof_lem_oracle_Q_sandwich}


	\begin{proof} 
	First, the fact that the statement in \eqref{Q_concavity_stronger} follows from \eqref{Q_concavity} and \eqref{Q_smoothness} is a classical result on strongly-convex, Lipschitz functions, see, for instance,  \citet[Theorem 2.1.12]{nesterov2013introductory}. We prove \eqref{Q_concavity} and \eqref{Q_smoothness} below.
 
    Let $\bomega, \bomega'\in \cB_d(\bomega^*,\delta_0)$   with  $\delta_0$ satisfying \eqref{cond_init_fix}.
     Pick any $k\in [K]$.    By noting that 
     \begin{align} \nonumber
    		\btheta_k^{(t+1)} &=  \btheta_k^{(t)} +  \eta_k ~  \nabla_{\btheta_k}   Q(\bomega \mid      \bomega^{(t)})   \mid_{\bomega =  \bomega^{(t)}}\\\label{grad_QN}
            &= \btheta_k^{(t)} +  \eta_k ~  \sum_{j=1}^p  \pi(x_j;\bomega^*) { \alpha_k^{(t)} A (x_j; \btheta_k^{(t)})\over \pi(x_j; \bomega^{(t)})} \left( x_j - \bX ^\T A( \btheta_k^{(t)})\right),
    	\end{align} 
     we find
   \begin{align*}
		q_k(\bomega) = \nabla_{\btheta_k} Q(\bomega \mid \bomega^*)  &=  \sum_{j=1}^p   {\alpha^*_k A_{\btheta_k^*}(x_j)   }  
   ( x_j - \bX^\T A_{\btheta_k}) 
  = \alpha_k^* ~ \bX^\T (A_{\btheta_k^*} - A_{\btheta_k})
	\end{align*}
       so that we obtain
		\begin{align}\label{bd_cross_grad}\nonumber
			&(\btheta_k - \btheta_k')^\T \left(q_k(\bomega) - q_k(\bomega') \right)\\ \nonumber
			& = -\alpha_k^*    \sum_{j=1}^p (\btheta_k - \btheta_k')^\T x_j \left( A_{\btheta_k}(x_j)-A_{\btheta_k'}(x_j)  \right)\\\nonumber
			&= - \alpha_k^*    \sum_{j=1}^p(\btheta_k  - \btheta_k')^\T x_j  \int_{0}^{1} A_{\btheta_{k,u}}(x_j)  (\be_j - A_{\btheta_{k,u}})^\T \bX (\btheta_k  - \btheta_k') \rd u \\\nonumber
			&=  - \alpha_k^* \int_{0}^{1}  (\btheta_k- \btheta_k')^\T 
			\left[ 	\sum_{j=1}^p  A_{\btheta_{k,u}}(x_j)  x_j x_j^\T - \bX^\T A_{\btheta_{k,u}} A_{\btheta_{k,u}}^\T \bX \right] (\btheta_k  - \btheta_k') \rd u\\\nonumber
			&\le -\alpha_k^*  \|\btheta_k  - \btheta_k'\|_2^2  \inf_{u\in [0,1]} \lambda_L\left(\bX^\T \Sigma_{A_{\btheta_{k,u}}}\bX\right)\\
			&= -\alpha_k^*  \|\btheta_k  - \btheta_k'\|_2^2  \inf_{u\in [0,1]} \lambda_L(H_{\btheta_{k,u}})
		\end{align}
		The second equality uses an Taylor expansion of $A_{\btheta_k} (x_j)$ around $\btheta_k'$ and we use the notation $\btheta_{k,u} = u\btheta_k  + (1-u)\btheta_k'$ for any $u\in [0,1]$. The last step uses \eqref{def_H}. 
  
		Similarly, we have 
		\begin{align}\label{bd_square_grad}\nonumber
			&\| \nabla_{\btheta_k} Q(\bomega \mid \bomega^*) - \nabla_{\btheta_k} Q(\bomega' \mid \bomega^*)\|_2\\ \nonumber
        &=\alpha_k^* 
			\left\|
			\sum_{j=1}^p \int_{0}^{1} A_{\btheta_{k,u}}(x_j)  x_j (\be_j - A_{\btheta_{k,u}})^\T \bX (\btheta_k - \btheta_k') \rd u \
			\right\|_2\\
			&\le   \alpha_k^* \|\btheta_k - \btheta_k'\|_2   \sup_{u\in [0,1]} \lambda_1(H_{\btheta_{k,u}}).
		\end{align}  
		In view of \eqref{bd_cross_grad} and \eqref{bd_square_grad} and the fact that
		$\bomega, \bomega' \in \cB_d(\bomega^*, \delta_0)$ implies 
		$\btheta_k, \btheta_k' \in \cB(\btheta_k^*, \delta_0/\os)$, the Euclidean ball around $\btheta_k^*$ with radius $\delta_0/\os$, so that  $\btheta_{k,u} \in \cB(\btheta_k^*, \delta_0/\os)$ for any $u\in [0,1]$,   \eqref{Q_concavity} and \eqref{Q_smoothness} follow by invoking \cref{lem_hess_unif} with $\btheta^* =\btheta_k^*$ and $\btheta_k = \btheta_{k,u}$ for all $k\in[K]$. The proof is complete. 
	\end{proof}

	\subsubsection{Proof of \cref{lem_GS_theta}: the gradient smoothness of the surrogate function $Q$ and the Lipschitz continuity of $M_\alpha$}\label{app_sec_proof_lem_GS_theta}

	\begin{proof} 
	Let $\bomega  \in \cB_d(\bomega^*, \delta_0)$ be arbitrary. 
    We first prove \eqref{cond_GS_theta}. From \eqref{grad_QN}, we argue 
		\begin{align*}
		&	\nabla_{\btheta_k} Q(\bomega \mid \bomega ) - \nabla_{\btheta_k} Q(\bomega \mid \bomega^{*})\\ &= \sum_{j=1}^p \pi_{\bomega^*}(x_j)\left( 
			{\alpha_k A_{\btheta_k}(x_j) \over  \pi_{\bomega}(x_j)} -
			{\alpha_k^* A_{\btheta_k^*}(x_j) \over \pi_{\bomega^*}(x_j)}\right) \bX^\T (\be_j - A_{\btheta_k})\\
			&= \sum_{j=1}^p  {1\over  \pi_{\bomega}(x_j) }
			\left[\alpha_k A_{\btheta_k}(x_j) \sum_{a\ne k} \alpha_a^* A_{\btheta_a^*}(x_j) - \alpha_k^* A_{\btheta_k^*}(x_j)\sum_{a\ne k} \alpha_a A_{\btheta_a}(x_j) \right]  \bX^\T (\be_j - A_{\btheta_k}).
		\end{align*}
		By adding and subtracting terms, it now suffices to bound from above 
		\begin{align*}
			&T_1 :=   \sum_{a\ne k}  \left\| \sum_{j=1}^p   
			\left(\alpha_k A_{\btheta_k}(x_j) - \alpha_k^* A_{\btheta_k^*}(x_j)\right) {\alpha_a^* A_{\btheta_a^*}(x_j)  \over  \pi_{\bomega}(x_j)   } \bX^\T (\be_j - A_{\btheta_k})\right\|_2\\
			&T_2 :=   \sum_{a\ne k} \left\| \sum_{j=1}^p  (\alpha_a A_{\btheta_a}(x_j) - \alpha_a^* A_{\btheta_a^*}(x_j)) { \alpha_k^* A_{\btheta_k^*}(x_j)\over  \pi_{\bomega}(x_j)   } \bX^\T (\be_j - A_{\btheta_k})\right\|_2
		\end{align*}

		\paragraph{Bounding $T_1$.}  
		We start with  the inequality $T_1 \le T_{11} + T_{12}$ where
		\begin{align*}
			T_{11} & =  |\alpha_k - \alpha_k^*| \sum_{a\ne k}  \left\| \sum_{j=1}^p   
			{\alpha_a^* A_{\btheta_a^*}(x_j)  \over  \pi_{\bomega}(x_j)  } A_{\btheta_k}(x_j) \bX^\T (\be_j - A_{\btheta_k})\right\|_2 \\
			T_{12} &= \alpha_k^* \sum_{a\ne k}  \left\| \sum_{j=1}^p  
			\left(A_{\btheta_k}(x_j)-A_{\btheta_k^*}(x_j)\right) {\alpha_a^* A_{\btheta_a^*}(x_j)  \over  \pi_{\bomega}(x_j)  } \bX^\T (\be_j - A_{\btheta_k})\right\|_2
		\end{align*}
		For the term $T_{12}$, after a Taylor expansion of  $A_{\btheta_k}(x_j)$ around $\btheta_k^*$, we find  that 
		\begin{align*}
			T_{12} &=  \alpha_k^* \sum_{a\ne k}  \left\| \sum_{j=1}^p    {\alpha_a^* A_{\btheta_a^*}(x_j)  \over  \pi_{\bomega}(x_j)  } 
			\int_{0}^1  A_{\btheta_{k,u}}(x_j) (\btheta_k - \btheta_k^*)^\T \bX^\T (\be_j - A_{\btheta_{k,u}})
			\bX^\T (\be_j - A_{\btheta_k})\rd u\right\|_2\\
			&\le  \sup_{u\in [0,1]}  \sum_{a\ne k}  \left\| \sum_{j=1}^p    {\alpha_a^* A_{\btheta_a^*}(x_j) \alpha_k^* A_{\btheta_{k,u}}(x_j)  \over  \pi_{\bomega}(x_j)  } 
			\bX^\T (\be _j - A_{\btheta_{k,u}})
			(\be_j - A_{\btheta_k})^\T \bX \right\|_\op \|\btheta_k - \btheta_k^*\|_2.
		\end{align*}
		Here, we recall $\btheta_{k,u} = u\btheta_k + (1-u)\btheta_k^*$. 
  Further we denote 
		\[
		\rho_j :=  {\alpha_a^* A_{\btheta_a^*}(x_j) \alpha_k^* A_{\btheta_{k,u}}(x_j)  \over  \pi_{\bomega}(x_j)  },\qquad \forall~ j\in [p].
		\]
		We proceed to bound from above 
		\[
		T_{121}:=  \sup_{u\in [0,1]}  \sum_{a\ne k}  \left\| \sum_{j=1}^p    \rho_j 
		\bX^\T (\be_j - A_{\btheta_k})
		(\be_j - A_{\btheta_k})^\T \bX \right\|_\op \|\btheta_k - \btheta_k^*\|_2
		\]
		and 
		\[
		T_{122} :=  \sup_{u\in [0,1]}  \sum_{a\ne k}  \left\| \sum_{j=1}^p    \rho_j 
		\bX^\T (\be_j - A_{\btheta_{k,u}})
		(\be_j - A_{\btheta_{k,u}})^\T \bX \right\|_\op \|\btheta_k - \btheta_k^*\|_2.
		\]
  This is indeed sufficient as  $T_{12}  \le  \sqrt{T_{121}T_{122}}$ follows from the Cauchy-Schwarz inequality. 
		For any $a\ne k$, define the midpoint between $\btheta_a^*$ and $\btheta_k^*$ as 
		\[
			\bar \btheta_{ak}^* := {1\over 2}(\btheta_a^* + \btheta_k^*).
		\]
		We have 
		\begin{align*}
			T_{121} & =  \sup_{u\in [0,1]}  \sum_{a\ne k}  \left\| \sum_{j=1}^p   {\rho_j \over A_{\bar\btheta_{ak}^*}(x_j)}A_{\bar\btheta_{ak}^*}(x_j)
			\bX^\T (\be_j - A_{\btheta_k})
			(\be_j - A_{\btheta_k})^\T \bX \right\|_\op \|\btheta_k - \btheta_k^*\|_2\\
			&\le  \sup_{u\in [0,1]}  \sum_{a\ne k} \max_{j\in [p]}  {\rho_j \over A_{\bar\btheta_{ak}^*}(x_j)} \left\| \sum_{j=1}^p   A_{\bar\btheta_{ak}^*}(x_j)
			\bX^\T (\be_j - A_{\btheta_k})
			(\be_j - A_{\btheta_k})^\T \bX \right\|_\op \|\btheta_k - \btheta_k^*\|_2.
		\end{align*}
		After we invoke \cref{lem_perturb_hess,lem_coh}, we find   that 
		\begin{equation} \label{bd_T_12}
			T_{121} \lesssim \sum_{a\ne k} \left(
			\os^2 +  {\os^4\over 4} \| \btheta_a^*  -    \btheta_k^*\|_2^2
			\right) (\alpha_a^* \vee \alpha_k^*)\exp\left(
			-{\us^2\over 8}\|\btheta_a^*-\btheta_k^*\|_2^2\right) \|\btheta_k-\btheta_k^*\|_2. 
		\end{equation}
		We can repeat the same arguments to prove that the bound in \eqref{bd_T_12} also holds for $T_{122}$, and hence for $T_{12}$. 
		
		Now, regarding the term $T_{11}$, by using the midpoint \eqref{def_mid_theta}, we have 
		\begin{align*}
			T_{11} &= {|\alpha_k - \alpha_k^*| \over \alpha_k^*} \sum_{a\ne k}  \left\| \sum_{j=1}^p   
			{\alpha_a^*\alpha_k^* A_{\btheta_a^*}(x_j) A_{\btheta_k}(x_j)  \over  \pi_{\bomega}(x_j) A_{\bar\btheta_{ak}^*}(x_j) }  A_{\bar\btheta_{ak}^*}(x_j) \bX^\T (\be_j - A_{\btheta_k})\right\|_2.
		\end{align*} 
		Since 
		\begin{align*}
			& \left\| \sum_{j=1}^p   
			{\alpha_a^* \alpha_k^*A_{\btheta_a^*}(x_j) A_{\btheta_k}(x_j)  \over  \pi_{\bomega}(x_j) A_{\bar\btheta_{ak}^*}(x_j) }  A_{\bar\btheta_{ak}^*}(x_j) \bX^\T (\be_j - A_{\btheta_k})\right\|_2 \\
			&=\sup_{v \in \bS^{L-1}} \sum_{j=1}^p   
			{\alpha_a^* \alpha_k^* A_{\btheta_a^*}(x_j) A_{\btheta_k}(x_j)  \over  \pi_{\bomega}(x_j) A_{\bar\btheta_{ak}^*}(x_j) }  A_{\bar\btheta_{ak}^*}(x_j) v^\T \bX^\T (\be_j - A_{\btheta_k})\\
			&\le \sup_{v \in \bS^{L-1}} \left(
			\sum_{j=1}^p A_{\bar\btheta_{ak}^*}(x_j) [v^\T \bX^\T (\be_j - A_{\btheta_k})]^2
			\right)^{1/2} \left(
			\sum_{j=1}^p \left({\alpha_a^*\alpha_k^* A_{\btheta_a^*}(x_j) A_{\btheta_k}(x_j)  \over  \pi_{\bomega}(x_j) A_{\bar\btheta_{ak}^*}(x_j) } \right)^2 A_{\bar\btheta_{ak}^*}(x_j)
			\right)^{1/2}\\
			&\le \left\| \sum_{j=1}^p   A_{\bar\btheta_{ak}^*}(x_j)
			\bX^\T (\be_j - A_{\btheta_k})
			(\be_j - A_{\btheta_k})^\T \bX \right\|_\op^{1/2}  \max_{j\in [p]} {\alpha_a^*\alpha_k^* A_{\btheta_a^*}(x_j) A_{\btheta_k}(x_j)  \over  \pi_{\bomega}(x_j) A_{\bar\btheta_{ak}^*}(x_j) },
		\end{align*}
		invoking \cref{lem_perturb_hess} and \cref{lem_coh}  gives that 
		\begin{equation}\label{bd_T_11}
			T_{11} \lesssim  {|\alpha_k - \alpha_k^*| \over \alpha_k^*} \sum_{a\ne k}\left(
			\os + {\os^2 \over 2}\|\btheta_a^* - \btheta_k^*\|_2
			\right)  (\alpha_a^* \vee \alpha_k^*)\exp\left(
			-{\us^2\over 8}\|\btheta_a^*-\btheta_k^*\|_2^2\right).
		\end{equation}
		
		Finally, combing \eqref{bd_T_12} and \eqref{bd_T_11}  yields that 
		\begin{equation}\label{bd_T_1}
			\begin{split}
				T_1 &\lesssim \sum_{a\ne k} \left(
				\os^2 +  {\os^4\over 4} \| \btheta_a^*  -    \btheta_k^*\|_2^2
				\right) (\alpha_a^* \vee \alpha_k^*)\exp\left(
				-{\us^2\over 8}\|\btheta_a^*-\btheta_k^*\|_2^2\right) \|\btheta_k-\btheta_k^*\|_2 \\
				& \quad  +\sum_{a\ne k}\left(
				\os + {\os^2 \over 2}\|\btheta_a^* - \btheta_k^*\|_2
				\right)  (\alpha_a^* \vee \alpha_k^*)\exp\left(
				-{\us^2\over 8}\|\btheta_a^*-\btheta_k^*\|_2^2\right)  {|\alpha_k - \alpha_k^*| \over \alpha_k^*} \\
				&\lesssim  \left(\os +  \os^2  \Delta\right)\exp\left(-{\us^2\over 8} \Delta^2 \right) K\oa   \left( {|\alpha_k - \alpha_k^*| \over \alpha_k^*} + \left(\os +  \os^2  \Delta\right) \|\btheta_k-\btheta_k^*\|_2\right).
			\end{split}
		\end{equation}
		
		\paragraph{Bounding $T_2$.}  Bounding $T_2$ essentially follows the same arguments as that of $T_1$.  Start with 
		$
		T_2 \le T_{21} + T_{22}
		$
		where 
		\begin{align*}
			&T_{21}:=   \sum_{a\ne k} {|\alpha_a-\alpha_a^*| \over \alpha_a^*}\left\| \sum_{j=1}^p     { \alpha_k^* A_{\btheta_k^*}(x_j)   \alpha_a^* A_{\btheta_a}(x_j) \over  \pi_{\bomega}(x_j)   } \bX^\T (\be_j - A_{\btheta_k})\right\|_2,\\
			&T_{22}:=  \sum_{a\ne k} \left\| \sum_{j=1}^p  ( A_{\btheta_a}(x_j) - A_{\btheta_a^*}(x_j)) {  \alpha_a^* \alpha_k^* A_{\btheta_k^*}(x_j)  \over  \pi_{\bomega}(x_j)   } \bX^\T (\be_j - A_{\btheta_k})\right\|_2.
		\end{align*}
		Note that 
		\[
		T_{21} \le  {\|\balpha-\balpha^*\|_\i \over \ua} \sum_{a\ne k}   \left\| \sum_{j=1}^p     { \alpha_k^* A_{\btheta_k^*}(x_j)   \alpha_a^* A_{\btheta_a}(x_j) \over  \pi_{\bomega}(x_j)   } \bX^\T (\be_j - A_{\btheta_k})\right\|_2.
		\]
		The same arguments of bounding $T_{11}$ above gives 
		\begin{equation}\label{bd_T_21}
			T_{21}\lesssim  \left(\os  + {\os^2\over 2} \Delta\right)\exp\left(-{\us^2\over 8} \Delta^2 \right) K\oa   {\|\balpha-\balpha^*\|_\i \over \ua}.
		\end{equation}
		On the other hand, repeating the arguments of bounding $T_{12}$ gives that
		\begin{equation}\label{bd_T_22}
			T_{22} \lesssim   \left(
			\os^2 +  {\os^4\over 4} \Delta^2
			\right)  \exp\left(
			-{\us^2\over 8}\Delta^2\right) K\oa \max_{a \in [K]}\|\btheta_a-\btheta_a^*\|_2. 
		\end{equation}
		Collecting \eqref{bd_T_21}, \eqref{bd_T_22} as well as \eqref{bd_T_1} yields that 
		\begin{align*}
			&\| \nabla_{\btheta_k} Q(\bomega \mid \bomega ) - \nabla_{\btheta_k} Q(\bomega \mid \bomega^{*}) \|_2\\
			&\quad  \lesssim\left(\os +  \os^2  \Delta\right)\exp\left(-{\us^2\over 8} \Delta^2 \right) K\oa   \left( {\|\balpha   - \balpha ^*\|_\i \over \ua} + \left(\os +  \os^2  \Delta\right) \max_{a \in [K]} \|\btheta_a-\btheta_a^*\|_2\right).
		\end{align*} 
		Finally, we complete the proof of \eqref{cond_GS_theta} by observing that both \cref{lem_perturb_hess} and \cref{lem_coh} as well as the arguments above are valid uniformly over $\bomega \in \cB_d(\bomega^*, \delta_0)$.

        Next we prove \eqref{cond_GS_M}.
		By definition in \eqref{iter_alpha_popu}, we can  split, for any $k\in [K]$,
		\begin{align*}
			M_k({\bomega}) -  M_k({\bomega^*})  &=  \sum_{j=1}^p \pi_{\bomega^*}(x_j) \left(
			{\alpha_k A_{\btheta_k}(x_j) \over \pi_{\bomega}(x_j)}
			-{\alpha_k^* A_{\btheta_k^*}(x_j)   \over \pi_{\bomega^*}(x_j)}
			\right) = S_1 + S_2
		\end{align*}
		with 
		\begin{align*}
			S_1 &= \sum_{a\ne k} \sum_{j=1}^p 
			{\alpha_k A_{\btheta_k}(x_j) -\alpha_k^* A_{\btheta_k^*}(x_j)   \over \pi_{\bomega}(x_j)}  \alpha_a^* A_{\btheta_a^*}(x_j)\\
			S_2 &=\sum_{a\ne k}   \sum_{j=1}^p   
			{\alpha_a A_{\btheta_a}(x_j)- \alpha_a^* A_{\btheta_a^*}(x_j) \over \pi_{\bomega}(x_j)}
			\alpha_k^* A_{\btheta_k^*}(x_j)  
		\end{align*}
        
		\paragraph{Bounding of $S_1$.}
		We start with the decomposition of $S_1$ 
		\begin{align*}
			S_1 &=   {\alpha_k -\alpha_k^*\over \alpha_k^*} \sum_{a\ne k} \sum_{j=1}^p 
			{\alpha_k^*  \alpha_a^* A_{\btheta_a^*}(x_j) A_{\btheta_k}(x_j)  \over \pi_{\bomega}(x_j)}  + 
			\sum_{a\ne k} \sum_{j=1}^p 
			\alpha_k^* \alpha_a^* A_{\btheta_a^*}(x_j) {A_{\btheta_k}(x_j) - A_{\btheta_k^*}(x_j)   \over \pi_{\bomega}(x_j)}\\
			&:= S_{11} + S_{12}.
		\end{align*}
		Using the midpoint notation in \eqref{def_mid_theta}, we find that 
		\begin{align}\label{bd_S_11}\nonumber
			|S_{11}| & = {|\alpha_k -\alpha_k^*|\over \alpha_k^*} \sum_{a\ne k} \sum_{j=1}^p 
			{\alpha_k^*  \alpha_a^* A_{\btheta_a^*}(x_j) A_{\btheta_k}(x_j)  \over \pi_{\bomega}(x_j) A_j(\bar \btheta_{ak}^*)}A_{\bar \btheta_{ak}^*}(x_j)\\\nonumber
			&\le {|\alpha_k -\alpha_k^*|\over \alpha_k^*} \sum_{a\ne k} \max_{j\in [p]} 	{\alpha_k^*  \alpha_a^* A_{\btheta_a^*}(x_j) A_{\btheta_k}(x_j)  \over \pi_{\bomega}(x_j) A_{\bar \btheta_{ak}^*}(x_j)}\\
			&\lesssim {|\alpha_k -\alpha_k^*|\over \alpha_k^*} \sum_{a\ne k}  (\alpha_a^* \vee \alpha_k^*)\exp\left(
			-{\us^2\over 8}\|\btheta_a^*-\btheta_k^*\|_2^2\right)&&\text{by \cref{lem_coh}}.
		\end{align}
		Regarding $S_{12}$, we have that 
		\begin{align*}
			|S_{12}| &\le \sum_{a\ne k} \left| \sum_{j=1}^p 
			{\alpha_k^* \alpha_a^* A_{\btheta_a^*}(x_j) \over \pi_{\bomega}(x_j)} (A_{\btheta_k}(x_j) - A_{\btheta_k^*}(x_j)  )\right|\\
			&= \sum_{a\ne k} \left| \sum_{j=1}^p 
			{\alpha_k^* \alpha_a^* A_{\btheta_a^*}(x_j) \over \pi_{\bomega}(x_j)} \int_{0}^1 A_{\btheta_{k,u}}(x_j) (\btheta_k-\btheta_k^*)^\T \bX^\T (\be_j - A_{\btheta_{k,u}}) \rd u\right|\\
			&\le \sup_{u\in [0,1]} \sum_{a\ne k} \left\| \sum_{j=1}^p 
			{\alpha_k^* \alpha_a^* A_{\btheta_a^*}(x_j)A_{\btheta_{k,u}}(x_j) \over \pi_{\bomega}(x_j)}  \bX^\T (\be_j - A_{\btheta_{k,u}}) \right\|_2 \|\btheta_k-\btheta_k^*\|_2. 
		\end{align*}
		By repeating the same arguments of bounding  $T_{21}$ above, we further find
		\begin{equation}\label{bd_S_12}
			|S_{12}|\lesssim   \sum_{a\ne k}\left(
			\os + {\os^2 \over 2}\|\btheta_a^* - \btheta_k^*\|_2
			\right)  (\alpha_a^* \vee \alpha_k^*)\exp\left(
			-{\us^2\over 8}\|\btheta_a^*-\btheta_k^*\|_2^2\right) \|\btheta_a^* - \btheta_k^*\|_2.
		\end{equation}

		\paragraph{Bounding $S_2$.} 
		Using the inequality
		\begin{align*}
			|S_2| &\le \sum_{a\ne k}  {|\alpha_a - \alpha_a^*| \over \alpha_a^*} \sum_{j=1}^p   
			{\alpha_a^* 	\alpha_k^* A_{\btheta_k^*}(x_j)  A_{\btheta_a}(x_j)  \over \pi_{\bomega}(x_j)}
			+ \sum_{a\ne k}  \left| \sum_{j=1}^p   
			{\alpha_a^* 	\alpha_k^* A_{\btheta_k^*}(x_j)    \over \pi_{\bomega}(x_j)} (A_{\btheta_a}(x_j) - A_{\btheta_a^*}(x_j))\right|
		\end{align*}
and	after repeating  the above arguments, we find that 
		\begin{align}\label{bd_S_2}\nonumber
			|S_2| &\lesssim {\|\balpha - \balpha^*\|_\i \over \ua}   \sum_{a\ne k}  (\alpha_a^* \vee \alpha_k^*)\exp\left(
			-{\us^2\over 8}\|\btheta_a^*-\btheta_k^*\|_2^2\right)
			\\
			&\quad + \sum_{a\ne k}  \|\btheta_a-\btheta_a^*\|_2 \left(
			\os + {\os^2 \over 2}\|\btheta_a^* - \btheta_k^*\|_2
			\right)  (\alpha_a^* \vee \alpha_k^*)\exp\left(
			-{\us^2\over 8}\|\btheta_a^*-\btheta_k^*\|_2^2\right).
		\end{align} 
		Combining \eqref{bd_S_11}, \eqref{bd_S_12} and \eqref{bd_S_2} yields the following bound 
        \begin{align*}
			\max_{k\in [K]} | M_k({\bomega}) -  M_k({\bomega^*})|
			&\lesssim   \exp\left(
			-{\us^2\over 8}\Delta^2\right)  K\oa\left( 
			{\|\balpha - \balpha^*\|_\i \over \ua}   +   \left(
			1 + \os\Delta
			\right)\os \max_{a\in [K]}  \|\btheta_a-\btheta_a^*\|_2
			\right)
		\end{align*}for any fixed $\bomega \in \cB_d(\bomega^*, \delta_0)$.
        Since the arguments hold uniformly over  $\cB_d(\bomega^*, \delta_0)$, the proof  is complete. 
	\end{proof}

	\subsubsection{Proof of \cref{lem_dev_EM}: concentration inequality of the EM-updates within the specified neighborhood}\label{app_sec_proof_lem_dev_EM}

	\begin{proof} 
		    Our proof is based on the following discretization of  
	$$
	\cB_d(\bomega^*, \delta_0) = \left\{
	\bomega:   \|\balpha - \balpha^*\|_\i  \le \delta_0 \ua, ~\|\btheta_k - \btheta_k^*\|_2 \le \delta_0/\os,~ \forall~  k\in [K]
	\right\}.
	$$  
	For any given 
	$ \epsilon_1 \in (0, \ua/4]$ and $ \epsilon_2 \in (0, \delta_0/\os]$, let $\cN_{\epsilon_1} (\Delta^{K-1})$  be an $\epsilon_1$-covering set  (in $\ell_\i$-norm) of    $\Delta^{K-1}$
	and $\cN_{\epsilon_2}$ be the $\epsilon_2$-net (in $\ell_2$-norm) of $\{	\btheta\in \RR^L:   \|\btheta\|_2 \le \delta_0/\os \}$. Then,
	for any $k\in [K]$,
	$
	\cN_{\epsilon_2}(\btheta_k^*) := \left\{
	\btheta + \btheta_k^*: \btheta \in \cN_{\epsilon_2}
	\right\}
	$
	is the $\epsilon_2$-net of $\{\btheta\in \RR^L: \|\btheta - \btheta_k^*\|_2 \le  \delta_0/\os\}$. 
	Consider the set 
	$$
	\cN_{\epsilon_1,\epsilon_2} = \cN_{\epsilon_1} (\Delta^{K-1}) \otimes \cN_{\epsilon_2} (\btheta_1^*)\otimes \cdots \otimes \cN_{\epsilon_2}(\btheta_K^*). 
	$$
	We have that for any $\bomega \in \cB_d(\bomega^*, \delta_0)$, there exists some $\bomega' \in \cN_{\epsilon_1,\epsilon_2}$ such that 
	\begin{equation}\label{net_property}
		\|\balpha - \balpha'\|_\i \le  \epsilon_1, \quad \max_{k\in [K]} \|\btheta_k - \btheta_k'\|_2 \le \epsilon_2 
	\end{equation}
	as well as 
	\begin{equation}\label{net_property_2}
		\|\balpha^* - \balpha'\|_\i \le \epsilon_1 + \delta_0\ua , \quad \max_{k\in [K]} \|\btheta_k^*  - \btheta_k'\|_2 \le \delta_0/\os.
	\end{equation}
	Moreover, from \citet[Lemma A.4]{ghosal2001entropies} and the classical result on the covering number of an Euclidean ball, the cardinality of $\cN_{\epsilon_1,\epsilon_2}$ satisfies
	\begin{equation}\label{card_N12}
		|\cN_{\epsilon_1,\epsilon_2}| \le |\cN_{\epsilon_1} (\Delta^{K-1})| |\cN_{\epsilon_2}|^K \le   \left(
		5  \over   \epsilon_1
		\right)^{K-1} \left(
		3 \delta_0 \over \os\epsilon_2
		\right)^{KL}.
	\end{equation}

		Since for any $\bomega \in \cB_d(\bomega^*, \delta_0)$, there exists $\bomega' \in \cN_{\epsilon_1,\epsilon_2}$ satisfying \eqref{net_property} -- \eqref{net_property_2} such that for all $k\in [K]$,
		\begin{equation}\label{eq_start_discre}  
            | \wh M_{k}(\bomega) -  M_{k}(\bomega)|      \le  | \wh M_{k}(\bomega') -  M_{k}(\bomega')| + |  \wh M_{k}(\bomega) -    M_{k}(\bomega) - \wh M_{k}(\bomega') + M_{k}(\bomega')|,
		\end{equation}
		we first bound the second term:
		 	\begin{equation} \label{decomp_M_diff}
			\begin{split}
            & \  \left | 	 \wh M_{k}(\bomega) -    M_{k}(\bomega) - \wh M_{k}(\bomega') + M_{k}(\bomega')\right| \\
            &=     \left | 
            \sum_{j=1}^p (\wh \pi_j - \pi_{\bomega^*}(x_j)) \left(  {\alpha_k A_{\btheta_k}(x_j)\over \pi_{\bomega}(x_j)}
            -    {\alpha_k' A_{\btheta_k '}(x_j) \over \pi_{\bomega'}(x_j)}\right) 
            \right|\\ 
				&\le \|\wh \pi - \pi_{\bomega^*}\|_1  \max_{j\in [p]}\left|  {\alpha_k A_{\btheta_k}(x_j)\over \pi_{\bomega}(x_j)}
				-    {\alpha_k' A_{\btheta_k '}(x_j) \over \pi_{\bomega'}(x_j)}\right|  \\
				&\le    2 \max_{j\in [p]} \left(
				|\alpha_k - \alpha_k'| {A_{\btheta_k}(x_j)\over \pi_{\bomega}(x_j)} + {\alpha'_k |A_{\btheta_k}(x_j)-A_{\btheta_k '}(x_j) | \over \pi_{\bomega}(x_j)} + {\alpha'_k A_{\btheta_k '}(x_j) \over \pi_{\bomega'}(x_j) } {|\pi_{\bomega}(x_j) - \pi_{\bomega'}(x_j)|\over \pi_{\bomega}(x_j)}
				\right)\\
				& \le  2 \max_{j\in [p]} \left(
				{\epsilon_1 \over \alpha_k}  {\alpha_k A_{\btheta_k}(x_j)\over \pi_{\bomega}(x_j)} + {\alpha'_k |A_{\btheta_k}(x_j)-A_{\btheta_k '}(x_j) | \over \pi_{\bomega}(x_j)} + {\alpha'_k A_{\btheta_k '}(x_j) \over \pi_{\bomega'}(x_j) } {|\pi_{\bomega}(x_j) - \pi_{\bomega'}(x_j)|\over \pi_{\bomega}(x_j)}
				\right)\\
				&\le  2 \max_{j\in [p]} \left(
				{\epsilon_1 \over \alpha_k}   + {\alpha'_k |A_{\btheta_k}(x_j)-A_{\btheta_k '}(x_j) | \over \pi_{\bomega}(x_j)} +  {|\pi_{\bomega}(x_j) - \pi_{\bomega'}(x_j)|\over \pi_{\bomega}(x_j)}
				\right).
			\end{split}
		\end{equation}  
		In order to invoke our perturbation bounds in \cref{lem_perturb}, we need to verify that its conditions are satisfied. This follows by noting that $\epsilon_2 \le \delta_0/\os \le c_0/\|\bX\|_{\i,2}$ under \eqref{cond_init_fix}, $\delta_0\le c_0<1/2$ and 
		\begin{equation}\label{lb_alpha}
			\alpha_k \ge  \alpha_k^*   - \delta_0 \ua  \ge 
   \alpha_k^* -{1\over2} \alpha_k^* \ge
      {1\over 2 }     \ua   \ge  2 \epsilon_1.
		\end{equation} 
		Hence,  invoking  \cref{lem_perturb} gives
		\begin{align}\label{bd_lipschitz}\nonumber
			&\max_{k\in [K]} \left| \wh M_{k}(\bomega)  -  M_{k}(\bomega) -  \wh M_{k}(\bomega') + M_{k}(\bomega') \right|\\\nonumber
			&\quad \le {4\epsilon_1 \over \ua} +   {2\alpha_k' \over \alpha_k}   {\alpha_k A_{\btheta_k}(x_j) \over \pi_{\bomega}(x_j)}  ~ 3 \epsilon_2 \|\bX\|_{\i, 2}  +   {4\epsilon_1 \over  \ua}   + 8\epsilon_2 \|\bX\|_{\i,2} \\
			&\quad  \lesssim   \epsilon_2\|\bX\|_{\i,2} +  {\epsilon_1 \over  \ua}.
		\end{align}	
		The last step uses \eqref{lb_alpha} and 
		\begin{equation}\label{ub_alpha}
			\alpha_{k'} \le \alpha_k + \epsilon_1 \le \alpha_k^* + \delta_0\ua + 
			\ua / 4 \le
   2\alpha_k^*.
		\end{equation} 
		In conjunction with \eqref{eq_start_discre}, we further obtain
		\begin{equation}\label{bd_M_penult}
			\sup_{\bomega \in \cB_d(\bomega^*, \delta_0)} \max_{k\in [K]} \left| \wh M_{k}(\bomega) -  M_{k}(\bomega)\right|  \lesssim  \max_{\bomega \in \cN_{\epsilon_1,\epsilon_2}}\max_{k\in [K]} \left| \wh M_{k}(\bomega) -  M_{k}(\bomega)\right| + \epsilon_2\|\bX\|_{\i,2} +  {\epsilon_1 \over  \ua}.
		\end{equation}
		We proceed to bound from above the first term on the right.
		To this end, fix any $\bomega \in \cN_{\epsilon_1,\epsilon_2}$ satisfying \eqref{net_property_2}. We find that 
		\begin{align*}
			\left| \wh M_{k}(\bomega) -  M_{k}(\bomega)\right| & = \left|
			\sum_{j=1}^p (\wh \pi_j - \pi_{\bomega^*}(x_j)) { \alpha_k A_{\btheta_k}(x_j)\over \pi_{\bomega}(x_j)}
			\right| = {1\over N} \left| \sum_{i=1}^N \left(
			E_i - \pi_{\bomega^*}
			\right)^\T h_k
			\right|
		\end{align*}
		with $[h_k]_j :=  \alpha_k A_{\btheta_k}(x_j)/\pi_{\bomega}(x_j)$ for all $j\in [p]$ and $E_1, \ldots, E_N$ are i.i.d. samples from Multinomial$(1; \pi_{\bomega^*})$. 
  Note that  
		\begin{align*}
			\var(E_i^\T h_k)  &\le \sum_{j=1}^p  \pi_{\bomega^*}(x_j)  {\alpha_k^2 A_{\btheta_k}(x_j)^2 \over \pi_{\bomega}(x_j)^2}\\ 
            &\le  \left( 1 +{|\pi_{\bomega^*}(x_j) - \pi_{\bomega}(x_j)| \over \pi_{\bomega}(x_j)}\right) \sum_{j=1}^p    {\alpha_k^2 A_{\btheta_k}(x_j)^2 \over \pi_{\bomega}(x_j) }\\
			&\le  \left(1 +  {\epsilon_1 \over  \ua} + \delta_0  + 4\epsilon_2 \|\bX\|_{\i,2}\right) \alpha_k &&\text{by \eqref{bd_perturb_pi} in \cref{lem_perturb} and \eqref{net_property_2}}\\
			& \le 4\alpha_k.
		\end{align*}  
        The last step uses \eqref{cond_init_fix}, $\delta_0\le c_0<1/2$, 
		$\epsilon_2 \le \delta_0/\os$ and $\epsilon_1 \le  \ua/4$. 
		Further note that
		\[
		|E_i^\T h_k| \le \max_{j\in [p]}  {\alpha_k  A_{\btheta_k}(x_j)  \over \pi_{\bomega}(x_j) } \le 1.
		\]
		An application of the Bernstein inequality together with the union bounds argument yields that, for any $t>0$,
		\begin{align*}
			&\PP\left\{\max_{\bomega \in \cN_{\epsilon_1,\epsilon_2}}\max_{k\in [K} \left| \wh M_{k}(\bomega) -  M_{k}(\bomega)\right|   \gtrsim 
			\sqrt{\alpha_k t \over N} + {t \over N}\right\} \\
			&\qquad \le  2 \exp\left(-t + \log |\cN_{\epsilon_1,\epsilon_2}| \right)\\
			& \qquad \le   2\exp\left\{-t +  KL\log\left({3 \delta_0\over  \os \epsilon_2}\right)  +  (K-1)\log\left({5\over   \epsilon_1}\right)  \right\} &&\text{by \eqref{card_N12}}.
		\end{align*} 
		In view of \eqref{bd_M_penult}, by invoking the event $\cE_2$, 
		the proof of the first result is completed by choosing 
		\[
		\epsilon_1 =  \ua \left({1\over 4} \wedge { KL \over N}\right), \qquad \epsilon_2 = {1\over \os}\left(\delta_0 \wedge {\os\over \|\bX\|_{\i,2}} { KL \over N}\right),\qquad t = CKL\log(N)
		\]
        and using \eqref{cond_N_explict} to collect terms.

		We use similar argument to bound from above
		\begin{align*}
			& \sup_{\bomega \in \cB_d(\bomega^*, \delta_0)}  \| \nabla_{\btheta_k} \wh Q(\bomega  \mid \bomega)  - \nabla_{\btheta_k} Q(\bomega  \mid \bomega) \|_2
			 \le  \max_{\bomega' \in \cN_{\epsilon_1,\epsilon_2}}  \| \nabla_{\btheta_k} \wh Q(\bomega'  \mid \bomega')  - \nabla_{\btheta_k} Q(\bomega'  \mid \bomega') \|_2\\
			&\quad  + \sup_{\substack{\bomega \in \cB_d(\bomega^*, \delta_0), ~ \bomega'\in \cN_{\epsilon_1,\epsilon_2}\\ \bomega,~ \bomega' \text{ satisfy }\eqref{net_property}}} \| \nabla_{\btheta_k} \wh Q(\bomega  \mid \bomega)  - \nabla_{\btheta_k} Q(\bomega  \mid \bomega)  - \nabla_{\btheta_k} \wh Q(\bomega'  \mid \bomega') + \nabla_{\btheta_k} Q(\bomega'  \mid \bomega') \|_2.
		\end{align*}
		Pick any $\bomega \in  \cB_d(\bomega^*, \delta_0) $ and $\bomega'\in \cN_{\epsilon_1,\epsilon_2}$ satisfying \eqref{net_property}. We bound from above 
		\begin{align*}
			&\| \nabla_{\btheta_k} \wh Q(\bomega  \mid \bomega)  - \nabla_{\btheta_k} Q(\bomega  \mid \bomega)  - \nabla_{\btheta_k} \wh Q(\bomega'  \mid \bomega') + \nabla_{\btheta_k} Q(\bomega'  \mid \bomega') \|_2\\
			& = \left\|
			\sum_{j=1}^p (\wh \pi_j - \pi_{\bomega^*}(x_j)) \left( {\alpha_k A_{\btheta_k}(x_j) \over \pi_{\bomega}(x_j)}\bX^\T (\be_j - A_{\btheta_k}) -  {\alpha_k' A_{\btheta_k '}(x_j)  \over \pi_{\bomega'}(x_j)}\bX^\T (\be_j - A_{\btheta_k'}) \right)
			\right\|_2\\
			&\le  \left\|
			\sum_{j=1}^p (\wh \pi_j - \pi_{\bomega^*}(x_j)) \left(  {\alpha_k A_{\btheta_k}(x_j) \over \pi_{\bomega}(x_j)} -  {\alpha_k' A_{\btheta_k '}(x_j)  \over \pi_{\bomega'}(x_j)}\right) \bX^\T (\be_j - A_{\btheta_k})  
			\right\|_2 \\
        &\qquad +    \left\|
			\sum_{j=1}^p (\wh \pi_j - \pi_{\bomega^*}(x_j))  {\alpha_k' A_{\btheta_k '}(x_j)  \over \pi_{\bomega'}(x_j)}\bX^\T (A_{\btheta_k} - A_{\btheta_k'}) 
			\right\|_2\\
			&\le \|\wh \pi - \pi_{\bomega^*}\|_1 \max_{j\in [p]} \left|
			{\alpha_k A_{\btheta_k}(x_j) \over \pi_{\bomega}(x_j)} -  {\alpha_k' A_{\btheta_k '}(x_j)  \over \pi_{\bomega'}(x_j)}
			\right|  \|\bX^\T (\be_j - A_{\btheta_k})\|_2 \\
            &\qquad +   \|\wh \pi - \pi_{\bomega^*}\|_1\max_{j\in [p]}  {\alpha_k' A_{\btheta_k '}(x_j)  \over \pi_{\bomega'}(x_j)}\|\bX^\T (A_{\btheta_k}-A_{\btheta_k'})\|_2\\
			&\le  4  \max_{j\in [p]} \left|
			{\alpha_k A_{\btheta_k}(x_j) \over \pi_{\bomega}(x_j)} -  {\alpha_k' A_{\btheta_k '}(x_j)  \over \pi_{\bomega'}(x_j)}
			\right|   \|\bX\|_{\i,2}  + 2\|\bX\|_{\i, 2}\max_{j\in [p]} |A_{\btheta_k}(x_j)-A_{\btheta_k '}(x_j) |.
		\end{align*}
		By the argument  in \eqref{decomp_M_diff}, \eqref{bd_perturb_A} and \eqref{bd_perturb_pi},  the  above is bounded from above by (in order)
		\begin{equation}\label{bd_lips_Q}
			\left( \epsilon_2\|\bX\|_{\i,2}+  {\epsilon_1 \over  \ua}\right) \|\bX\|_{\i,2} .
		\end{equation}

		It remains to bound from above 
		\begin{align*}
			&\max_{\bomega  \in \cN_{\epsilon_1,\epsilon_2}}  \| \nabla_{\btheta_k} \wh Q(\bomega   \mid \bomega )  - \nabla_{\btheta_k} Q(\bomega   \mid \bomega ) \|_2\\
			&\quad \le 2 \max_{\bomega  \in \cN_{\epsilon_1,\epsilon_2}}\max_{v \in \cN_L(1/2)}   v^\T \left(\nabla_{\btheta_k} \wh Q(\bomega   \mid \bomega )  - \nabla_{\btheta_k} Q(\bomega   \mid \bomega ) \right)\\
			&\quad = 2 \max_{\bomega  \in \cN_{\epsilon_1,\epsilon_2}}\max_{v \in \cN_L(1/2)}  \sum_{j=1}^p (\wh \pi_j - \pi_{\bomega^*}(x_j))  {\alpha_k A_{\btheta_k}(x_j) \over \pi_{\bomega}(x_j)} v^\T \bX^\T (\be_j - A_{\btheta_k})
		\end{align*}
        where $\cN_L(1/2)$ is the $(1/2)$-net of $\bS^{L-1}$ and satisfies $|\cN_L(1/2)| \le 5^L$ (see, for instance, \cite{vershynin2018high}). 
		Fix any $\bomega  \in \cN_{\epsilon_1,\epsilon_2}$ and $v\in \cN_L(1/2)$. Observe that  
		\[
		\sum_{j=1}^p (\wh \pi_j - \pi_{\bomega^*}(x_j))  {\alpha_k A_{\btheta_k}(x_j) \over \pi_{\bomega}(x_j)} v^\T \bX^\T (\be_j - A_{\btheta_k}) := {1\over N}\sum_{i=1}^N (E_i - \pi_{\bomega^*})^\T h_v
		\]
		with 
		$$[h_v]_j =  {\alpha_k A_{\btheta_k}(x_j) \over \pi_{\bomega}(x_j)} v^\T \bX^\T (\be_j - A_{\btheta_k}),\qquad \forall j\in [p].$$
		Also note that 
		\begin{align*}
			 \var(E_i^\T h_v)  &\le \sum_{j=1}^p \pi_{\bomega^*}(x_j) {\alpha_k^2 A_{\btheta_k}(x_j)^2 \over \pi_{\bomega}(x_j)^2} v^\T  \bX^\T (\be_j - A_{\btheta_k}) (\be_j - A_{\btheta_k})^\T \bX v\\
			&\le \max_{j\in [p]} {\pi_{\bomega^*}(x_j) \over \pi_{\bomega}(x_j)}  {\alpha_k^2 A_{\btheta_k}(x_j) \over \pi_{\bomega}(x_j)}  \sum_{j=1}^p A_{\btheta_k}(x_j)v^\T  \bX^\T (\be_j - A_{\btheta_k})(\be_j - A_{\btheta_k})^\T \bX v\\
			&\le \alpha_k\max_{j\in [p]} \left(1 + {|\pi_{\bomega}(x_j) - \pi_{\bomega^*}(x_j)| \over \pi_{\bomega}(x_j)}\right)   \lambda_1\left(
			\bX^\T \Sigma_{A_{\btheta_k}} \bX
			\right)\\
			& \overset{(i)}{\lesssim} \alpha_k \left(
			1 + \epsilon_2\|\bX\|_{\i,2}+  {\epsilon_1 \over  \ua} + \delta_0
			\right)\lambda_1\left(
			H_{\btheta_k}
			\right)\\
			&\overset{(ii)}{\lesssim} \alpha_k ~ \os^2
		\end{align*} 
        where the step $(i)$ uses \eqref{bd_perturb_pi} and \eqref{net_property_2} while the step $(ii)$ is due to  \eqref{def_event_H_local}, \eqref{cond_init_fix}, 
			$\epsilon_2 \le {\delta_0/ \os}$ and $\epsilon_1 \le\ua/4$. By further noticing 
		\[
		E_i^\T h_v \le \max_{j\in [p]} {\alpha_k A_{\btheta_k}(x_j) \over \pi_{\bomega}(x_j)} |v^\T \bX^\T (\be_j - A_{\btheta_k})| \le  2\|\bX\|_{\i,2},
		\]
		applying  Bernstein's inequality and the union bound over $\bomega  \in \cN_{\epsilon_1,\epsilon_2}$ and $v\in \cN_L(1/2)$ yields 
		\[
		\max_{\bomega  \in \cN_{\epsilon_1,\epsilon_2}}\max_{v \in \cN_L(1/2)}  \sum_{j=1}^p (\wh \pi_j - \pi_{\bomega^*}(x_j))  {\alpha_k A_{\btheta_k}(x_j) \over \pi_{\bomega}(x_j)} v^\T \bX^\T (\be_j - A_{\btheta_k}) \lesssim \os \sqrt{\alpha_k t \over N} + {t \|\bX\|_{\i, 2}\over N}
		\]
		with probability at least 
		\begin{align*}
			&1 - 2 \exp\left(-t + \log |\cN_{\epsilon_1,\epsilon_2}| + \log |\cN_L(1/2)|\right)\\
			& \qquad \le  1 - 2 \exp\left\{-t + KL\log\left({3\delta_0\over  \os \epsilon_2}\right)  +(K-1)\log\left({5\over  \epsilon_1}\right) + L\log(5) \right\} &&\text{by \eqref{card_N12}}.
		\end{align*}
		We complete the proof by choosing 
		\[
		\epsilon_1 =   \ua\left( {1\over 4}  \wedge {  KL \over N}\right), \qquad \epsilon_2 = {\delta_0\over \os}\wedge  {   KL \over  \|\bX\|_{\i,2}  N},\qquad t = C K L\log(N),
		\]
		taking the union bounds over $k\in [K]$ and using \eqref{cond_N_explict} to collect terms.
	\end{proof}

    \subsection{Concentration inequalities for quantities related with the information matrix}\label{app_sec_embedding}

    The following subsections contain deviation inequalities between $N_\btheta$, $\rI_\btheta$, $\rII_\btheta$, $H_\btheta$ in \cref{def_N_I_II,def_H} and their corresponding expectations, derived under \cref{random-points}.

    \subsubsection{Concentration inequalities related with $\rI_\theta$, $\rII_\theta$ and $N_\theta$} 

	
	\begin{lemma}\label{lem_I_op}
Grant \cref{random-points}. 
Let  $\btheta \in \RR^L$  with $\|\btheta\|_2\le B$ and $s\ge2$ be arbitrary
and assume $p$ is large enough such that 
$$ c_{\mathsf{Bern}} \cdot p>  L\log(7) + s\log(p)$$ where $c_{\mathsf{Bern}}$ is the universal constant appearing in Bernstein's exponential inequality for sums of independent sub-exponential random variables.
 We have, with probability $1-4p^{1-s}$,
\begin{align*}
    \left\|\rI_\btheta  -   \EE[ \rI_\btheta]\right\|_\op & \lesssim \os^2 \exp(2\os^2 B) p^{1-s/2 } +   \os^2 p^{{1\over2}+ \os B \sqrt{2s/ \log p}} \sqrt{   L\log(7) + s\log(p) }
\end{align*}
In particular, 
 for arbitrary $\delta>0$ and  for $p\ge p_0$ with $p_0$ being the smallest integer such that 
\begin{equation}\label{lb_p} \delta^2 \log p_0 \ge 2s \os^2 B^2,\end{equation}
we have
\begin{align*}
  \PP\left\{   \left\|\rI_\btheta  -   \EE[ \rI_\btheta]\right\|_\op  \lesssim   \os^2 p^{{1\over2} + \delta}   \sqrt{ L\log(7) + s \log(p) } \right\} \ge 1-4 p^{1-s}.
\end{align*}
\end{lemma}
\begin{proof} By definition and a standard discretization argument (see, for instance, \cite{vershynin2018high})
\begin{align*}
			\left \|\rI_\btheta - \EE \left[ \rI_\btheta \right] \right\|_\op &= \sup_{v \in \bS^{L-1}} v^\T \left(
			\rI_\btheta - \EE \left[ \rI_\btheta \right] 
			\right)v \\
			&=   \sup_{v \in \bS^{L-1}} 
			\sum_{j=1}^p \left\{ (X_j^\T v)^2 e^{X_j^\T \btheta}  
			- \EE \left[ (X_j^\T v)^2 e^{X_j^\T \btheta}   \right]  \right\}\\
   &\le 3\max_{v\in \cN_L(1/3)} \sum_{j=1}^p \left\{ (X_j^\T v)^2 e^{X_j^\T \btheta}  
			- \EE \left[ (X_j^\T v)^2 e^{X_j^\T \btheta}   \right]  \right\}
		\end{align*}
Here, $\cN_L(1/3)$ is a $(1/3)$-net of $\bS^{L-1}$ and satisfies $|\cN_L(1/3)| \le 7^L$ (see, for instance, \cite{vershynin2018high}).
Next, we use a truncation device. For fixed $\btheta\in \RR^L$ with $\|\btheta\|_2\le B$, the random variables   $X_j^\T \btheta$
are zero mean sub-Gaussian random variables with sub-Gaussian constant no greater than  $\|\btheta\|_2^2\, \os^2\le B^2 \os^2$. Consequently,
the events
\begin{equation}\label{def_event}
			\cX_j(s,\btheta) =  
   \left\{
			  |X_j^\T \btheta| \le \os B \sqrt{2 s\log(p)}) 
			\right\}
		\end{equation}
have probabilities 
		\begin{equation}\label{bd_event_tail}
			\PP\left(
			\cX_j(s,\btheta)
			\right)  \ge 1 - 2 p^{-s}.
		\end{equation} 
Clearly,
\begin{align*}
			&\left| \sum_{j=1}^p \left\{ (X_j^\T v)^2 e^{X_j^\T \btheta}  
			- \EE \left[ (X_j^\T v)^2 e^{X_j^\T \btheta}   \right]  \right\} \right|\\
   &\le \left| \sum_{j=1}^p \left\{ (X_j^\T v)^2 e^{X_j^\T \btheta} 1_{\cX_j(s,\btheta) }
			- \EE \left[ (X_j^\T v)^2 e^{X_j^\T \btheta}  1_{\cX_j(s,\btheta)}  \right]  \right\} \right|\\
   &+   \left| \sum_{j=1}^p \left\{ (X_j^\T v)^2 e^{X_j^\T \btheta}  1_{\cX_j^c(s,\btheta)} 
			- \EE \left[ (X_j^\T v)^2 e^{X_j^\T \btheta}  1_{\cX_j^c(s,\btheta)}  \right]  \right\} \right|
		\end{align*}
On the event $\cap_{j\in [p]} \cX_j(s,\btheta)$, which holds with probability at least $1-2p^{1-s}$, we have
\[  \sum_{j=1}^p  (X_j^\T v)^2 e^{X_j^\T \btheta}  1_{\cX_j^c(s,\btheta)}  =0\]
by definition,
while
\begin{align*} 
			\sum_{j=1}^p\EE \left[ (X_j^\T v)^2 e^{X_j^\T \btheta}  1_{\cX_j^c(s,\btheta)} \right] & \le 
			\sum_{j=1}^p \sqrt{
				\EE \left[ (X_j^\T v)^4 e^{2X_j^\T \btheta}\right]
			}  \sqrt{\PP(\cX_j^c(s,\btheta))}\\
			&\lesssim  p \os^2 e^{2\os ^2B^2} 2 p^{-s/2}  &&\text{by \cref{lem_moment_bds} and \eqref{bd_event_tail}} 
		\end{align*}
Since $X_j^\T v$ is $\os$ sub-Gaussian, the distribution of $(X_j^\T v)^2$ is sub-exponential (with parameter $\le \os^2$). This implies that the distribution of 
\[ W_j(s,\btheta,v) := (X_j^\T v)^2 e^{X_j^\T \btheta}  1 _{\cX_{j(s,\btheta)}} \]
is sub-exponential, with parameter $$ \| W_j(s,\btheta,v)\|_{\psi_1}\le \os^2 p^{\os B \sqrt{2s/ \log p}} :=\kappa$$
Bernstein's inequality for sums of independent sub-exponential random variables, see \cite[Section 2.8.1]{vershynin2018high},
states that, for some numerical constant $c>0$ and any $t\ge 0$,
		\[
		\PP\left\{
		 \sum_{j=1}^p \left(W_j(s,\btheta,v)
		- \EE \left[ W_j(s,\btheta,v)  \right] \right) \ge  p  \kappa t
		\right\} \le 2 \exp
		\left\{
		-c p \min\left(
		t, t^2
		\right) 
		\right\}
		\]
		We choose $t^2 = c^{-1}( L\log(7) + s\log(p) )/ p <1$, and we conclude, using  the union bound over $v \in \cN_L(1/3)$,
		\begin{equation}\label{bd_W_bern}
			\max_{v\in \cN_L(1/3)} \sum_{j=1}^p \left( W_j(s,\btheta,v) 
			- \EE \left[W_j(s,\btheta,v)  \right] \right) \lesssim  \os^2 p^{\os B \sqrt{2s/ \log p}} \sqrt{  p L\log(7) + sp\log(p) }
		\end{equation}
		with probability at least
		\begin{align*}
			&1 - 2 \cdot  7^L  \exp
			\left\{
			-c p  \min\left(
			t, t^2
			\right) 
			\right\}\ge  1 - 2 p^{-s}.
		\end{align*}
	The proof is complete. 
 	\end{proof}

     \bigskip 
     
	
	\begin{lemma}\label{lem_II}
	Grant \cref{random-points}.
    Let  $\btheta \in \RR^L$  with $\|\btheta\|_2\le B$ and $\delta>0$ and $s\ge2$. 
  For $p\ge p_0=p_0(B,\delta,s,\os)$, we have
		\begin{align*}
			\PP\left\{\left\|\rII_\btheta  -   \EE[ \rII_\btheta]\right\|_2  \lesssim \os \sqrt{ L+ \log(p) } ~ p^{\frac12+\delta}  
			\right\} \ge 1-4 p^{1-s}.
		\end{align*}
	\end{lemma}
	\begin{proof}
		We use the same arguments to prove \cref{lem_I_op}. 
  Again, the standard discretization argument ensures that 
		\[
		\left\|\rII_\btheta  -   \EE[ \rII_\btheta]\right\|_2 \le 2\max_{v\in \cN_L(1/3)} v^\T  \left(
		\rII_\btheta  -   \EE[ \rII_\btheta]
		\right) .
		\]
		Fix any $v \in \cN_L(1/3)$ and observe that 
		\begin{align*}
			v^\T \left(
			\rII_\btheta  -   \EE[ \rII_\btheta]
			\right) &\le \sum_{j=1}^p \left\{e^{X_j^\T \btheta} X_j^\T v 1_{\cX_j} - \EE\left[
			e^{X_j^\T \btheta} X_j^\T v 1_{\cX_j} 
			\right]\right\} + 
			\sum_{j=1}^p\left|
			\EE\left[
			e^{X_j^\T \btheta} X_j^\T v 1_{\cX_j^c} 
			\right]
			\right|
		\end{align*}
  on the event $\cap_{j\in [p]} \cX_j$ with the events $\cX_j:= \cX_j(s,\btheta)$ defined in \eqref{def_event}.
  The second term is no greater than
		\[
		p^{-s/2}  \sum_{j=1}^p \sqrt{
			\EE\left[
			e^{2X_j^\T \btheta} (X_j^\T v)^2\right]
		}  \le p^{1-s/2 } \os\left(
		1 + \os \|\btheta\|_2
		\right) e^{\os^2\|\btheta\|_2^2}
		\]
  by the Cauchy-Schwarz inequality, \cref{lem_moments} and \eqref{bd_event_tail}.
  
For the first term, we notice that $W_j (s,\btheta,v)= e^{X_j^\T \btheta} X_j^\T v 1_{\cX_j(s,\btheta,v)} $  is sub-Gaussian with sub-Gaussian parameter 
		\[
		\| W_j (s,\btheta,v) \|_{\psi_2} \le \kappa' = \os \exp\left( \os B \sqrt{2s \log p} \right) \le p^{\delta}.
		\]
  Moreover, 
  $\sum_{j=1}^p (W_j(s,\btheta,v) - \EE[W_{j}(s,\btheta,v)])$ is sub-Gaussian with sub-Gaussian parameter $2\kappa'\sqrt{p}$, whence,
  for all $t\ge 0$,
		\begin{equation}
			\PP\left\{
			\sum_{j=1}^p \left\{e^{X_j^\T \btheta} X_j^\T v 1_{\cX_j(s,\btheta,v)} - \EE\left[
			e^{X_j^\T \btheta} X_j^\T v 1_{\cX_j(s,\btheta,v)} 
			\right]\right\}  \ge  2  \kappa'\sqrt{p} ~ t
			\right\} \le 2\exp\left(
			-{t^2 \over 2}
			\right).
		\end{equation}
	We take	 $t^2 = C (L\log(7) + s\log (p) )$ and take the union bound over $v\in \cN_L(1/3)$ to complete the proof. 
	\end{proof}
	
	 \bigskip
	

 \begin{lemma}\label{lem_N}
Grant \cref{random-points}. 
 Let  $\btheta \in \RR^L$  with $\|\btheta\|_2\le B$ and $\delta>0$ and $s\ge2$. 
  For $p\ge p_0=p_0(B,\delta,s,\os)$, we have
		\begin{align*}
			\PP\left\{\left| N_\btheta -   \EE[N_\btheta]\right|  \lesssim   \sqrt{p^{1+\delta} \log(p)}
			\right\} \ge 1- 4 p^{1-s}.
		\end{align*}
	\end{lemma}
	\begin{proof}
 Again, 
		we follow the same arguments that we used to prove \cref{lem_I_op,lem_II}. Recall the events $\{\cX_j(s,\btheta)\}_{j\in [p]}$ from \eqref{def_event}. On the intersection of the events, we  have 
		\begin{align*}
			\left| N_\btheta -   \EE[N_\btheta]\right|  & \le  \left|
			\sum_{j=1}^p \left(
			e^{X_j^\T \btheta} 1_{\cX_j(s,\btheta)} - \EE\left[
			e^{X_j^\T \btheta} 1_{\cX_j(s,\btheta)} 
			\right]
			\right)
			\right| +  \sum_{j=1}^p
			\EE\left[ e^{X_j^\T \btheta} 1_{\cX_j^c(s,\btheta)} \right]\\
   &\le  \left|
			\sum_{j=1}^p \left(
			e^{X_j^\T \btheta} 1_{\cX_j(s,\btheta)} - \EE\left[
			e^{X_j^\T \btheta} 1_{\cX_j(s,\btheta)} 
			\right]
			\right)
			\right| + p^{1-s/2} e^{2\os^2 \|\btheta\|_2^2}
		\end{align*}
		using  Cauchy-Schwarz and \eqref{bd_event_tail}.
The first term on the right is a sum of independent bounded random variables, with
\[ \left| e^{X_j^\T \btheta} 1_{\cX_j(s,\btheta)}\right| \le \exp\left( \os B \sqrt{2s\log p}\right) \lesssim p^\delta
\]
almost surely and the result follows easily from  Hoeffding's inequality.
%
	\end{proof}

    \subsubsection{Concentration inequalities related with $H_\theta$ under Gaussianity}  
\label{sec:concentration-H-Gausz}
    For any $\btheta \in \RR^L$, the following lemma contains results on the moments related with $\rI_\btheta$, $\rII_\btheta$ and $N_\btheta$ under the condition
     \begin{assumption}\label{ass_E_gauss}
         $X_1,\ldots, X_p$ are i.i.d. from $\cN_L(0,\Sigma)$ with all the eigenvalues of $\Sigma$ belonging to the fixed interval $[\us^2,\os^2]\subset(0,\i)$.
    \end{assumption}

    \begin{lemma}\label{lem_expectations} 
		Grant \cref{ass_E_gauss}. Let $\btheta \in \RR^L$. Let $N_\btheta$, $\rI_\btheta$ and $\rII_\btheta$ be defined in \eqref{def_N_I_II}.  For any $v\in \RR^{L-1}$, we have 
		\begin{align*}
			&v^\T \EE[\rII_\btheta] =  p   (v^\T \Sigma \btheta) e^{\btheta^\T \Sigma \btheta/2};\\
			& v^\T \EE[\rI_\btheta] v= p   \left(
			v^\T \Sigma v +   (v^\T \Sigma\btheta)^2
			\right)  e^{\btheta^\T \Sigma \btheta / 2}.
		\end{align*}
		Moreover, 
		\begin{align*}
			&\EE[N_\btheta] = pe^{\btheta^\T \Sigma \btheta/2},\\
			&\|\EE[\rII_\btheta]\|_2 =  p  \|\Sigma \btheta\|_2 e^{\btheta^\T \Sigma \btheta / 2},\\
			&\lambda_1\left(\EE[\rI_\btheta]\right) = p  \left(
			\lambda_1(\Sigma) +   \|\Sigma \btheta\|_2^2 
			\right)  e^{\btheta^\T \Sigma \btheta / 2},\\
			&\lambda_L\left(\EE[\rI_\btheta]\right)  = p \lambda_L(\Sigma) e^{\btheta^\T \Sigma \btheta / 2}.
		\end{align*}
	\end{lemma}
	\begin{proof}
		Fix any $v\in \RR^L$.  By \cref{lem_moments} with $\sigma^2 = 1$, $u = \Sigma^{1/2}v$ and $\btheta = \Sigma^{1/2}\btheta$, we have 
		\begin{align*}
			v^\T \EE[\rII_\btheta]   =     p \EE\left[
			(v^\T X_j) e^{X_j^\T \btheta}
			\right]
			=  p   (v^\T \Sigma \btheta) e^{ \btheta^\T \Sigma\btheta /2}
		\end{align*}
		and 
		\begin{align*}
			v^\T \EE[\rI_\btheta] v  
			&=   p \EE\left[
			(X_j^\T v)^2 e^{X_j^\T \btheta}
			\right] = p  \left(
			v^\T \Sigma v  +  (v^\T \Sigma \btheta)^2 
			\right)  e^{\btheta^\T \Sigma\btheta / 2}.
		\end{align*} 
		Since  
		\[
		\EE[N_\btheta] = p \EE[e^{X_j^\T \btheta}] = pe^{\btheta^\T \Sigma\btheta /2},
		\]
		the other claims follow immediately from
		\begin{align*}
			\|\EE[\rII_\btheta]\|_2 = \sup_{v \in \bS^{L-1}} v^\T \EE[\rII_\btheta]
		\end{align*} 
        and 
        \[
            \lambda_1 \left(\EE[\rI_\btheta]\right)  = \sup_{v \in \bS^{L-1}} v^\T \EE[\rI_\btheta] v,\quad \lambda_L \left(\EE[\rI_\btheta]\right)  =  \inf_{v \in \bS^{L-1}} v^\T \EE[\rI_\btheta] v.
        \]
		The proof is complete.
	\end{proof}

     The following lemma follows immediately from Lemmas \ref{lem_I_op}, \ref{lem_II} \& \ref{lem_N}. 
	
	\begin{lemma}\label{lem_deviation_gauss}
		Grant \cref{ass_E_gauss}.
      Let $\btheta \in \RR^L$ with $\|\btheta\|_2 \le B$, $s\ge 2$, $\delta>0$ and $\epsilon>0$.
  For $p\ge p_0(B,s,\os,\delta,\epsilon)$,
  the following holds  with probability at least $1-4p^{1-s}$:
		\begin{enumerate}
			\item[(a)] 
			\[	
			(1-\epsilon) \EE[N_\btheta] \le N_\btheta \le (1+\epsilon)\EE[N_\btheta].
			\]
			\item[(b)]  
			\[
			\|\rII_\btheta\|_2 \le (1+\epsilon)   p  \os^2 B e^{\btheta^\T \Sigma \btheta / 2}
			\]
			\item[(c)] 
			\[
			(1-\epsilon) \lambda_L\left(\EE[\rI_\btheta]\right) \le \lambda_L \left(
			\rI_\btheta 
			\right) \le \lambda_1 \left(
			\rI_\btheta 
			\right) \le (1+\epsilon)  \lambda_1\left(\EE[\rI_\btheta]\right).
			\]
		\end{enumerate}
	\end{lemma}
	\begin{proof}
		Since $\EE[N_\btheta] \ge p$ from \cref{lem_expectations}, by invoking \cref{lem_N}, the first result follows as 
		\[
		N_\btheta \ge \EE[N_\btheta]  - \left|
		N_\btheta - \EE[N_\btheta]
		\right| \ge   \left(
		1 -  C \sqrt{\log(p) \over p^{1-\delta}}
		\right) \EE[N_\btheta]
		\]
		with probability $1-4p^{1-s}$. 
		
		Part (b) follows by  invoking \cref{lem_II} and  
		\cref{lem_expectations}. 
		
		Finally, by Weyl's inequality, we have 
		\[
		\left|\lambda_k(\rI_\btheta) - \lambda_k(\EE[\rI_\btheta]) \right| \le \| \rI_\btheta - \EE[\rI_\btheta]\|_\op.
		\]
		The last statement follows by invoking \cref{lem_I_op} and noting that 
		$
		\lambda_L(\EE[\rI_\btheta]) \ge  p \lambda_L(\Sigma)
		$ 
        from \cref{lem_expectations}.
	\end{proof}

	The following lemma is a key result that provides deviation inequality of $\|H_\btheta - \bar H_\btheta\|_\op$ where  
    \begin{equation}\label{def_bar_H_theta}
        \bar H_\btheta = {\EE[\rI_\btheta] \over \EE[N_\btheta]} -{\EE[\rII_\btheta] \EE[\rII_\btheta]^\T \over (\EE[N_\btheta])^2}.
    \end{equation}

	\begin{lemma}\label{lem_hess}
    Under \cref{ass_E_gauss}, one has $\bar H_\btheta = \Sigma$. 
    Let $\btheta \in \RR^L$ with $\|\btheta\|_2 \le B$ and fix any $s\ge 2$, $\delta>0$ and $\epsilon>0$.
  For $p\ge p_0(B,s,\os,\delta,\epsilon)$,
 %
		\begin{equation}\label{dev_H_diff_op}
			\PP\left\{
			\|H_\btheta - \bar H_\btheta\|_\op \lesssim  \os^2\sqrt{L + \log(p) \over p^{1-\delta}}
			\right\} \ge    1-4 p^{1-s}.
		\end{equation}
		Moreover, with the same probability as above, we have
		\[
		(1-\epsilon)\us^2 \le	\lambda_L(H_\btheta) \le \lambda_1(H_\btheta) \le  (1+\epsilon)\os^2.
		\]
	\end{lemma}
	
	\begin{proof}
        The first claim follows from for any $v\in \RR^L$,
        \begin{align*}
			v^\T \bar H_\btheta v &= {
				\EE[v^\T \rI_\btheta v]
				\over 	\EE[N_\btheta] } - {\EE[v^\T \rII_\btheta]\EE[\rII_\btheta^\T v] \over 	(\EE[N_\btheta])^2}\\ 
			&=     
			v^\T \Sigma v +   (v^\T \Sigma\btheta)^2    - (v^\T \Sigma\btheta)^2 &&\text{by  \cref{lem_expectations}} \\
			&= v^\T \Sigma v.
		\end{align*} 
        
		By adding and subtracting terms, we first have 
		\begin{align*}
			\|H_\btheta - \bar H_\btheta\|_\op &\le {\|\rI_\btheta - \EE[\rI_\btheta]\|_\op \over N_\btheta} + {\|\EE[\rI_\btheta]\|_\op \over  N_\btheta}{ \left|
				N_\btheta - \EE[N_\btheta]
				\right| \over \EE[N_\btheta]}\\
			&\quad + {\|\rII_\btheta(\rII_\btheta  - \EE[\rII_\btheta ])^\T\|_\op \over N_\btheta^2} +  {\|(\rII_\btheta  - \EE[\rII_\btheta ])(\EE[\rII_\btheta])^\T\|_\op \over N_\btheta^2}\\
			&\quad + {\|\EE[\rII_\btheta](\EE[\rII_\btheta])^\T\|_\op \over  N_\btheta^2} {(N_\btheta + \EE[N_\btheta] ) \left|
				N_\btheta - \EE[N_\btheta] 	\right| \over  (\EE[N_\btheta])^2}\\
			&\le  {\|\rI_\btheta - \EE[\rI_\btheta]\|_\op \over N_\btheta} + {\|\rII_\btheta\|_2 + \|\EE[\rII_\btheta]\|_2 \over N_\btheta} {\|\rII_\btheta  - \EE[\rII_\btheta ]\|_2 \over N_\btheta}\\
			&\quad  + \left[{\|\EE[\rI_\btheta]\|_\op \over  N_\btheta}+{\|\EE[\rII_\btheta]\|_2^2 \over  N_\btheta^2} {(N_\btheta + \EE[N_\btheta] )  \over  \EE[N_\btheta]}\right] { \left|
				N_\btheta - \EE[N_\btheta]
				\right| \over \EE[N_\btheta]}
		\end{align*}
		Note that \cref{lem_deviation_gauss} ensures that 
		\begin{equation}\label{lb_N_theta}
			N_\btheta \gtrsim \EE[N_\btheta] \ge p, \qquad \|\rII_\btheta\|_2 \lesssim p \os^2B e^{\btheta^\T \Sigma \btheta /2}
		\end{equation}
		with probability at least $1-4p^{1-s}$. 
  In conjunction with \cref{lem_expectations}, by invoking \cref{lem_I_op}, \cref{lem_II} and \cref{lem_N} with $\sigma^2 = \lambda_1(\Sigma)$, we conclude that 
		\begin{align*}
			\PP\left\{\|H_\btheta - \bar H_\btheta\|_\op\lesssim \os^2\sqrt{L  + \log(p) \over p^{1-\delta}} 
			\right\}	\ge 1-4 p^{1-s}, 
		\end{align*} 
		completing the proof of \eqref{dev_H_diff_op}. 
		
		The last result then follows by the definition of eigenvalues, \eqref{dev_H_diff_op} and Weyl's inequality. 
	\end{proof}

     \subsubsection{Concentration inequalities related with $H_\theta$ under sub-Gaussianity}

    The following lemma bounds the moments of $\EE[N_\btheta]$, $\EE[\rII_\btheta]$ and $\EE[\rI_\btheta]$ under \cref{random-points}. 

    \begin{lemma}\label{lem_expectations_subG} 
		Grant \cref{random-points}. 
         For any $\btheta \in \RR^L$, we have 
		\begin{align}\label{bds_Exp_Ntheta}
			&p \le \EE[N_\btheta] \le  pe^{\os^2\|\btheta\|_2/2}
        \end{align}
    and 
    \begin{align*}
        {\|\EE[\rII_\btheta]\|_2 \over \EE[N_\btheta]} \lesssim  \os + \os^2 \|\btheta\|_2,\qquad {\|\EE[\rI_\btheta]\|_\op \over \EE[N_\btheta]} \lesssim  \os^2 + \os^4 \|\btheta\|_2^2.
		\end{align*}
	\end{lemma}
	\begin{proof}
        The upper bound of $\EE[N_\btheta]$ is easy to see and the lower bounds follows by Jensen's inequality 
        \[
            \EE[N_\btheta] = p \EE[e^{X^\T \btheta}]  \ge p e^{\EE[X^\T \btheta]} = p.
        \]
        Regarding the other two results, fix any $v\in \bS^{L-1}$. For arbitrary $t> 0$, by using   \cref{ass:mu}, we have 
        \begin{align*}
            \EE[v^\T \rI_\btheta v]& =  \EE\left[(v^\T X)^2 e^{X^\T \btheta} 1\{|X^\T v| > t \}\right]  + \EE\left[(v^\T X)^2 e^{X^\T \btheta} 1\{|X^\T v| \le  t\}\right]\\
            &\le \sqrt{\EE[(v^\T X)^4 e^{2X^\T \btheta}}~ \sqrt{\PP(|X^\T v| > t)} +  t^2\EE[e^{X^\T \btheta}]\\
            &\le \sqrt{\EE[(v^\T X)^4 e^{2X^\T \btheta}}~ e^{-{t^2 \over 4\os^2 }}+t^2\EE[e^{X^\T \btheta}]
        \end{align*}
        so that, by choosing $t^2 = 8\os^2 \|\btheta\|_2^2$, 
        \begin{align*}
        {\EE[v^\T \rI_\btheta v]  \over \EE[N_\btheta]} 
        &\le {\sqrt{\EE[(v^\T X)^4 e^{2X^\T \btheta}} \over \EE[e^{X^\T \btheta}]} ~ e^{-{t^2 \over 4\os^2 }}+t^2 \\
        &\lesssim  {\os^2 e^{2\os^2\|\btheta\|_2^2 - {t^2 \over 4\os^2}} \over \EE[e^{X^\T \btheta}]} + t^2 &&\text{by \cref{lem_moment_bds}}\\
        &\le \os^2 + 8\os^4\|\btheta\|_2^2 &&\text{by \eqref{bds_Exp_Ntheta}}.
        \end{align*}
        Furthermore, we have  
		\begin{align*}
			{v^\T \EE[\rII_\btheta]\over \EE[N_\btheta]}   =     { \EE[
			(v^\T X) e^{X^\T \btheta}
			]\over \EE[e^{X^\T \btheta}] } \le \sqrt{\EE[(v^\T X)^2 e^{X^\T \btheta}] \over \EE[e^{X^\T \btheta}]} \lesssim \os + \os^2 \|\btheta\|_2.
		\end{align*}
        Since the above bounds hold for all $v$, the proof is complete. 
	\end{proof}

    Similar as \cref{lem_hess}, we have the following result under \cref{random-points}. 
     \begin{lemma}\label{lem_hess_subG}
    Grant \cref{random-points}. Let $\btheta \in \RR^L$ with $\|\btheta\|_2 \le B$, $s\ge 2$, $\delta>0$ and $\epsilon>0$. Assume 
    $
        \lambda_L(\bar H_\btheta) \ge \us^2.
    $
     Then for $p\ge p_0(B,s,\os, \us,\delta,\epsilon)$, \begin{equation}\label{dev_H_diff_op_subG}
    			\PP\left\{
    			\|H_\btheta - \bar H_\btheta\|_\op \lesssim  \os^2\sqrt{L + \log(p) \over p^{1-\delta}}
    			\right\} \ge    1-4 p^{1-s}.
    		\end{equation}
    		Moreover, for any $\epsilon>0$,  
      with the same probability as above and some constant $C>1$, we have
    		\[
    		(1-\epsilon)\us^2 \le	\lambda_L(H_\btheta) \le \lambda_1(H_\btheta) \le  (1+\epsilon)C\os^2.
    		\]
	\end{lemma}
    \begin{proof}
        The proof of \eqref{dev_H_diff_op_subG} is the same as that of \cref{lem_hess} except that \eqref{lb_N_theta} is replaced by 
        \[
           p \EE[N_\btheta] \lesssim  N_\btheta \lesssim  p \EE[N_\btheta]
        \]
        and 
        \[ 
            {\|\rII_\btheta\|_2 \over \EE[N_\btheta]} \le  {\|\rII_\btheta - \EE[\rII_\btheta]\|_2 \over \EE[N_\btheta]} +  {\EE[\|\rII_\btheta\|_2]\over \EE[N_\btheta]}  \le  (1+\epsilon)   (\os + \os^2 B)
        \]
        by using \cref{lem_expectations_subG}.
        The second statement follows from Weyl's inequality and noting that \cref{lem_expectations_subG} implies 
        \[
            \lambda_1(\bar H_\btheta) \le  {\|\EE[\rI_\btheta]\|_\op \over \EE[N_\btheta]} \lesssim  \os^2 + \os^4 B^2.
        \]
        This completes the proof.
    \end{proof}

\section{Results for  MoM estimation: \cref{sec_MoM}}\label{app_sec_MoM}

\subsection{Parameter recovery from the latent moments}\label{app_sec_MoM_latent}
We give the classical result for recovering the parameters of $\rho^*$ from its moments. It collects existing results on population-level parameter recovery via latent moments, stated explicitly in \cite{Lin89} and implicitly in \cite{Lindsay93}.

\begin{lemma}\label{Lindsay}
    For any  $\bomega^*\in \Omega^*$
satisfying  \eqref{cond_Delta1} and $\min_k \alpha_k^*>0$, the system of equations given by (\ref{mom}) and (\ref{cross-mom}) has  a unique solution which equals to $\bomega^*$, up to label switching. Moreover, the solution can be found explicitly and is given by the expressions below. 
\begin{enumerate} 
	\item The first coordinates $\theta_{11}^*,\theta_{12}^*, \ldots,  \theta_{1K}^*$ are the unique $K$ roots of the degree $K$ polynomial $P(x)$, in one variable, given by 
	\begin{equation}\label{firstcoordinate}
		P(x) :=  \mbox{\rm det}  \begin{pmatrix}
			1 &  m_1 & \dots &  m_{K} \\
			m_1 &  m_2 & \dots &  m_{K+1} \\
			\vdots & \vdots & & \vdots \\
			m_{K-1} &  m_K & \dots &  m_{2K-1} \\
			1 & x & \dots & x^K
		\end{pmatrix}.
	\end{equation}
	
	\item For each $k \in [K]$, the remaining  $L-1$  coordinates $\{\theta_{ik}^*\}_{2\le i\le L}$ 
	are uniquely given by 
	\begin{equation}\label{rest-coordinates}
		\theta_{ik}^* = \begin{pmatrix}
			m_{01; i} \\ \vdots \\  m_{(K-1)1; i}
		\end{pmatrix}^\top   \begin{pmatrix}
			1 &  m_1 & \dots &  m_{K-1} \\
			m_1 &  m_2 & \dots &  m_{K} \\
			\vdots & \vdots & & \vdots \\
			m_{K-1} &  m_K & \dots &  m_{2K-2} 
		\end{pmatrix}^{-1}    \begin{pmatrix}
			1 \\  \theta_{1k}^* \\ \vdots \\  (\theta_{1k}^{*})^{K-1}
		\end{pmatrix}.
	\end{equation}
	
	\item The mixture weight vector  $\balpha^*$   is uniquely given  by 
	\begin{equation}\label{weights}
		{\bm \alpha}^*  = \begin{pmatrix}
			1 & 1 & \ldots & 1 \\
			\theta_{11}^*& \theta_{12}^*&  \ldots & \theta_{1K}^* \\
			\vdots & \vdots & & \vdots \\
			(\theta_{11}^{*})^{K-1}& (\theta_{12}^{*})^{K-1}&  \ldots & (\theta_{1K}^{*})^{K-1} 
		\end{pmatrix}^{-1}\begin{pmatrix} 1 \\ m_1 \\ \vdots \\ m_{K -1}\end{pmatrix}.
	\end{equation}
\end{enumerate} 
\end{lemma}
\begin{proof}
 Results (\ref{firstcoordinate}) and (\ref{weights}) above can be found in \cite{Lin89},  whereas (\ref{rest-coordinates}) is implicit in \cite{Lindsay93}, and we derive its explicit form here.

The first and third result of Lemma \ref{Lindsay} have been known for several decades, in the theory on univariate mixtures. 
Consider the first coordinate $\theta_{11}^*, \ldots, \theta_{1K}^*$, respectively, of the $K$ parameter vectors $\btheta_1^*, \ldots, \btheta_K^*$ in $\RR^L$.  By assumption, they are distinct and, in the  notation of \cref{sec_MoM}, they are the $K$ support points of the one-dimensional distribution of $Z_1$, the first coordinate of the latent vector $Z \sim \rho^*$. Recall that $m_1, \ldots, m_{2K-1}$ are, by definition, moments of $Z_1$. Then by Theorem 2C in \cite{Lin89} (population version),  the polynomial equation $P(x) = 0$ has $K$ distinct roots, and they are equal to $\theta_{11}^*, \ldots, \theta_{1K}^*$. 

Next, one forms the system of equations $m_r = \sum_{k = 1}^{K}\alpha_k\theta_{1k}^{*r}$, for $ 0 \leq r \leq K - 1$, which for given $m_r$,  and  for $\theta_{1k}^*$ found above, is linear in  $\alpha_1, \ldots, \alpha_K$.  Since its  coefficient matrix is a Vandermonde matrix henceinvertible,   the system has the unique solution $\balpha^*$ given by (\ref{weights}). \\
\cite{Lindsay93} gave the road map to extending the univariate result to the multivariate case and we make it explicit here, in our notation.  Consider the matrix of moments  
\[ 
 \bm M :=  \begin{pmatrix}
            1 &  m_1 & \dots &  m_{K-1} \\
            m_1 &  m_2 & \dots &  m_{K} \\
            \vdots & \vdots & & \vdots \\
             m_{K-1} &  m_K & \dots &  m_{2K-2} 
 \end{pmatrix}  
\]
By Theorem 2A of  \cite{Lin89},  this matrix is non-singular.  Consider now the following $( K + 1) \times (K + 1)$ matrix 
\[ U(t) =\begin{pmatrix} \bm M  & {\bm a} \\
    {\bm b}^\top & t \end{pmatrix}, \]
for some generic vectors $\bm a, \bm b \in \RR^K$, and a scalar $t \in \RR$.

On the one hand, we have the following facts. Using the formula for block matrix determinants, we have 
\[ \det(U(t)) = \det({\bm M}) \det(t - {\bm b}^\top {\bm M}^{-1}{\bm a}).\]
Since $\det({\bm M}) > 0$,  the unique solution to $\det(U(t)) = 0$ is given by  
\begin{equation}\label{t} t =  {\bm b}^\top {\bm M}^{-1}{\bm a}.\end{equation} 
On the other hand, we also have the following. Since $\det({\bm M}) > 0$, then  $\text{rank}(U(t)) \geq K$, with maximal possible rank $K + 1$.  We now choose $\bm a, \bm b$ and $t$ such that $\text{rank}(U(t)) =  K$, and thus such that $\det(U(t)) = 0$. 
 
 The choices ${\bm a}  := (\theta_{1k}^{*r})_{r = 0}^{K-1}$, ${\bm b} := (m_{r, i})_{r=0}^{K-1}$,  $t = \theta_{ik}^*$ indeed cause this quantity to vanish, since then the $K+1$ columns of $U(\theta_{ik}^*)$ are spanned by the $K$ vectors $(
            1 , \theta_{1k}^* , \dots , (\theta_{1k}^{*})^{(K-1)} , \theta_{ik}^*)^\top$, $k = 1, \dots, K$. Combining this with (\ref{t}) gives the stated expression (\ref{rest-coordinates})
\end{proof}

\subsection{Selection of the primary axis via  the subspace estimation}\label{sec_mm_proj}
 Recall that in \cref{sec_MoM}  we have chosen the primary axis as $\be_1$, which leads to  $\Delta_1$ in \eqref{cond_Delta1}.  As we explain
 in \cref{rem_rate} below, finding a good primary axis $v$ relative to which  $\Delta_1$  is large is crucial for  the success of  the MoM estimation technique. 
In rare cases, the statistician may have \emph{a priori} knowledge of a good direction $v$ for which a lower bound  on $$\Delta(v^\T \btheta_1^*, \dots, v^\top \btheta_K^*) := \min_{k\ne k'}|v^\T(\btheta_k^*-\btheta_{k'}^*)|$$  is sufficiently large. 
 
In general, to obtain results that hold uniformly over the parameter space, one could choose $v$ randomly on the sphere $\bS^{L-1}$. Recall that 
$
    \Delta  :=   \min_{k\ne k'} \|\btheta_k^* - \btheta_{k'}^*\|_2.
$
A simple probabilistic argument (see, \cref{lem_unif_sphere}) gives that, for any $t \in (0,1)$ and any $v$ uniformly drawn from $\bS^{L-1}$, 
\begin{equation}\label{lb_separation}
    \PP\left\{
        \Delta(v^\T  \btheta_1^*, \dots, v^\top   \btheta_K^*) \ge {t \Delta    \over K^2 \sqrt{L}}  
    \right\} \ge 1-t.
\end{equation}
Fix such $v$, and let $R = (v, w_2, \ldots, w_L)$ be an $L\times L$ rotation matrix  with orthonormal columns. 
 Following an identical argument to that of the  proof of \cref{prop:theta_gap}, applied to the re-scaled targets $R^\T  \btheta_1^*,\ldots, R^\T  \btheta_K^*$, the constant $D$ scales as multiples of $(\Delta/(K^2\sqrt{L}))^{-cK}$, for some $c> 0$. 
 Thus, it  scales as $K^{\cO(K)}$ if $L < K$,  while it scales as $ L^{\cO(K)}$ otherwise.
   Fortunately, it is possible to eliminate the dependency of $D$ on $L$ altogether, by choosing a direction $v$ from the lower-dimensional subspace spanned by $\btheta_1^*,\ldots,\btheta_K^*$, which will allow us to improve upon (\ref{lb_separation}).
 Recall that the subspace of $\btheta_k^*$'s are contained in $\Gamma$ given by \eqref{def_Gamma}. Denote by $\wh V\in \bO_{L\times K}$ the first $K$ eigenvectors of its estimator $\wh\Gamma$ in \eqref{def_Gamma_hat}. In view of \cref{prop_eigensp}, we propose to choose the projection vector $v$ as 
\begin{equation}\label{def_proj}
    v = {\wh V\wh V^\T u \over \|\wh V\wh V^\T u\|_2}
\end{equation}
where the vector $u\in \RR^{L}$ contains i.i.d. entries of $\cN(0,1)$. The following lemma gives a lower bound on the desired minimum pairwise  separation relative to this  choice for  $v$. It is worth mentioning that   our analysis does not require any spectral condition on $\Gamma$. 

\begin{lemma}\label{lem_proj}
    Grant conditions in \cref{prop_eigensp}.   Then 
    for any $t \in (0,1)$ and any $v$ drawn as \eqref{def_proj}, $\mu$-almost surely for all sufficiently large $p$ satisfying $p\ge CL^2\log\log(p) / (\ua^2\Delta^4)$ and $\pi$-almost surely for all sufficiently large $N$ satisfying $N\ge CL^2\log\log(N) / (\ua^2\Delta^4)$,  the following holds with probability (with respect to the sampling probability of $u$ in \eqref{def_proj}) at least $1-t$:
    \[ 
        \Delta(v^\T  \btheta_1^*, \ldots, v^\T  \btheta_K^*) \ge {t \Delta  \over 2K^2 \sqrt{K}} .
    \]
\end{lemma}

Compared to  \eqref{lb_separation},  the dimension reduction in \eqref{def_proj}   eliminates the dependency on $L$ in the constant $D$ of \cref{prop:theta_gap}. 

      \begin{proof}
        Pick any $k\ne k'$. We first bound from below 
        $$|v^\T  \btheta_k^* - v^\T \btheta_{k'}^*| = {|(\btheta_k^* - \btheta_{k'}^*)^\T \wh V \wh V^\T u| \over
        \|\wh V \wh V^\T u\|_2} = {|(\btheta_k^* - \btheta_{k'}^*)^\T \wh V (\wh V^\T u)| \over
        \|\wh V^\T u\|_2}.
        $$
        Since, conditioning on $\wh V$, $\wh V^\T u \sim \cN_K(0, \bI_K)$ so that $\wh V^\T u / \|\wh V^\T u\|_2$ is uniformly distributed over $\bS^{K-1}$, invoking \cref{lem_unif_sphere} gives 
        that for all $t\ge 0$,
        \[
            \PP\left\{
                |v^\T   \btheta_k^* - v^\T  \btheta_{k'}^*| <  \|\wh V  (\btheta_k^* - \btheta_{k'}^*)\|_2 ~ t
            \right\} < t\sqrt{K}.
        \]
        To bound from above $\|\wh V^\T  (\btheta_k^* - \btheta_{k'}^*)\|_2$, recall that $\wh V \in \bO_{L\times K}$ denotes the left leading eigenvectors of $\wh\Gamma$. It then follows that 
        \begin{align*}
            \|\wh V^\T  (\btheta_k^* - \btheta_{k'}^*)\|_2^2 &  = (\btheta_k^* - \btheta_{k'}^*)^\T    \wh V \wh V^\T   (\btheta_k^* - \btheta_{k'}^*)\\
            & = \| \btheta_k^* -  \btheta_{k'}^*\|_2^2 - (\btheta_k^* - \btheta_{k'}^*)^\T  (\bI_L - \wh V \wh V^\T) (\btheta_k^* - \btheta_{k'}^*)\\
            &\ge   \Delta^2 - 2\btheta_k^{*\T}   (\bI_L - \wh V \wh V^\T)  \btheta_k^* - 2\btheta_{k'}^{*\T}  (\bI_L - \wh V \wh V^\T) \btheta_{k'}^*.
        \end{align*}
        Notice that 
        \begin{align*}
        \btheta_k^{*\T}   (\bI_L - \wh V \wh V^\T)  \btheta_k^* & = 
        \sup_{u\in \bS^{L-1}} u^\T (\bI_L - \wh V \wh V^\T)   \btheta_k^* \btheta_k^{*\T}   (\bI_L - \wh V \wh V^\T) u \\
        & \le {1\over \ua} \sup_{u\in \bS^{L-1}} u^\T (\bI_L - \wh V \wh V^\T) \sum_{a=1}^K \alpha_a^*   \btheta_a^* \btheta_a^{*\T}   (\bI_L - \wh V \wh V^\T) u \\
        &\le  {1\over \ua} \sup_{u\in \bS^{L-1}} u^\T (\bI_L - \wh V \wh V^\T) ~ \wh \Gamma~  (\bI_L - \wh V \wh V^\T) u \\
        &\qquad + {1\over \ua} \sup_{u\in \bS^{L-1}} u^\T (\bI_L - \wh V \wh V^\T) (\wh  \Gamma -\Gamma) (\bI_L - \wh V \wh V^\T) u\\
        &\le {1\over \ua}  \left( \lambda_{K+1}(\wh \Gamma) + \|\wh \Gamma - \Gamma\|_\op\right) \\ 
        &\le  {2\over \ua} \|\wh \Gamma - \Gamma\|_\op.
        \end{align*}
        The last step uses Weyl's inequality and $\lambda_{K+1}(\Gamma) = 0$. We write $\lambda_1(Q) \ge \lambda_2(Q) \ge \cdots \ge \lambda_d(Q)$ as the non-increasing eigenvalues of any $d\times d$ symmetric matrix $Q$. 

        Therefore, invoking \cref{prop_eigensp} implies
        \begin{equation}\label{cond_snr_dr}
         \Delta^2 \ua \ge 8 \|\wh \Gamma - \Gamma\|_\op,
        \end{equation}
        under the stated almost-sure asymptotics,
        we obtain that for all $t\ge 0$,
        \[
                 \PP\left\{
                |v^\T \btheta_k^* - v^\T  \btheta_{k'}^*| <   {t\Delta \over 2\sqrt{K}}  
            \right\} < t.
        \]
        By taking the union bounds over $k,k'\in [K]$ with $k\ne k'$, the proof is complete. 
    \end{proof}

\begin{remark}\label{rem_axis} 
    In practice we recommend to take multiple random projection vectors $v_1, \ldots, v_n$,  and select the one that yields the largest separation. Since their corresponding separations $\Delta(v_i^\T \btheta_1^*, \ldots, v_i^\T \btheta_K^*)$ are unknown, we propose to use the following criterion. For any $i\in [n]$, compute the moment vector $\wh\bmvec(v_i)$ as in \eqref{m-hat} with $h_r$ given by \eqref{eq:hr_def}, for each $v_i$,
    and its denoised version $\wt \bmvec(v_i)$ as in \eqref{def_dmm}, form the moment matrix of $\wt \bmvec(v_i)$ as 
    \[
      \wt \bM(v_i) :=    \begin{pmatrix}
		1 &  \wt m_1(v_i) & \dots &  \wt m_{K-1}(v_i) \\
		\wt m_1(v_i) &  \wt m_2(v_i) & \dots &  \wt m_{K}(v_i) \\
		\vdots & \vdots & & \vdots \\
		\wt m_{K-1}(v_i) &  \wt m_K(v_i) & \dots &  \wt m_{2K-2} (v_i)
	\end{pmatrix},
    \]
    and choose $v_{i^*}$  with $i_*$ selected as
    $
        i_* = \argmax_{i \in [n]} ~ \det(\wt \bM(v_i)).
    $
    The intuition lies in the important result in \citet[Theorem A2]{Lin89}  that 
    $
        \det(\bM(v_i)) =  (K!)^{-1} \prod_{1\le k<k'\le K} (
            v_i^\T  \btheta_k^* -v_i^\T \btheta_{k'}^*
        )^2
    $
    so that the selected $v_{i^*}$ approximately maximizes $\det(\bM(v_i))$, thereby leading to the largest separation among $v_i^\T \btheta_1^*, \ldots,v_i^\T \btheta_K^*.$ 
\end{remark}

\subsection{Remarks mentioned in \cref{sec_MoM}}\label{app_sec_mom_rem}

This section collects some remarks mentioned in \cref{sec_MoM} together with relevant proofs.

\begin{remark}[The importance of random features] \label{randX}
We highlight the importance played by the randomness of  $X_1, \ldots, X_p \sim \mu$ in our argument. It is enough to consider $m_r$ for some $r\in \NN$. We did show above that $m_r(\bomega^*)   \approx \bar{m}_{r} (\bomega^*)$, for $h_r$ defined by (\ref{eq:hr_def}), by using  a law of large numbers-type argument. It is natural to ask if we could use a different construction that would, instead, give exact equality.  Specifically, we ask the following question:  Given {\it generic,  non-random}  $x_1, \ldots, x_p$, does there exist a function  $s_r : \RR^L \to \RR$ such that $m_r  =  \bar{m}_{r}$? We now show 
that, unfortunately, no such function can exist, even for $r = 1$. 
To see why this claim
holds,  it is enough to consider $v = \be_1$, and suppose that there existed such a function $s_1$. To lighten notation in this argument, we let $\btheta_k := \btheta_k^*$,  for all $k$.  Using (\ref{mix}), and the definition of $m_1$ in (\ref{mom}),  if  equality held throughout in $m_1(\bomega^*)=\bar m_1(\bomega^*)$, 
then with $A_{{\btheta}_k}(x_j)$ denoting  $A_{{\btheta}_k}(x_j \mid x_1,\ldots,x_p)$, since $ x_1,\ldots,x_p$ are non-random, we have 
\[  \sum_{k = 1}^{K} \alpha_k\left[ \sum_{j = 1}^{p}A_{{\btheta}_k}(x_j) s_1(x_j) \right]  = \sum_{k=1}^{K} \alpha_k\theta_{1k}. \]
Let us write $\beta_j := s_1(x_{j})$ for $j \in [p]$.
Under the softmax parametrization (\ref{softmax}), these quantities therefore satisfy
\begin{align*}  \frac{ \sum_{j=1}^{p}\exp({x_{j}^\T{\btheta_k}}) \beta_j   } {   \sum_{\ell = 1}^{p} \exp({x_{\ell}^\T{\btheta_k} })} &= \theta_{1k}, \quad \quad\forall {\btheta_k} \in \RR^L\,,\end{align*}
or, differentiating in ${\btheta_k}$,
\begin{align}\label{eq:impossible}
  &  \frac{ \sum_{j=1}^{p}\exp ({x_{j }^\T{\btheta_k} } )x_{j } \beta_j  } {   \sum_{\ell = 1}^{p} \exp({x_{\ell }^\T{\btheta_k}})} - \frac{ \sum_{j=1}^{p}
    \exp( {x_{j }^\T{\btheta_k}}) \beta_j  } {   \sum_{\ell = 1}^{p} \exp({x_{\ell }^\T{\btheta_k}})}\frac{ \sum_{j=1}^{p}\exp({x_{j }^\T{\btheta_k}}) x_{j } } {   \sum_{\ell = 1}^{p} \exp({x_{\ell }^\T{\btheta_k}})}  = \be_1, 
      \quad
    \forall {\btheta_k} \in \RR^L.
\end{align}
Now, let $\mathcal C$ be the convex hull of ${ x}_{ 1}, \dots, { x}_{ p}$.
This is a nonempty polytope in $\RR^L$.
Assume without loss of generality that $x_1$ is an extreme point of $\mathcal C$, and let $\bm{a} \in \RR^L$ be any vector in the interior of the normal cone of $\mathcal C$ at $x_1$.
For any real numbers $\lambda_1, \dots, \lambda_p$, it holds that
\begin{equation*}
    \lim_{t \to \infty} \frac{ \sum_{j=1}^{p} \exp\left( {x_{j }^\T{(t \bm a)}} \right)\lambda_j  } {   \sum_{\ell = 1}^{p} \exp\left({x_{\ell }^\T{(t \bm a)}} \right)} = \frac{\sum_{j: x_j = x_1} \lambda_j}{|{j: x_j = x_1}|}\,.
\end{equation*}
Therefore, choosing ${\btheta_k} = t x_1$ in~\eqref{eq:impossible} and taking the limit $t \to \infty$ on both sides yields
\begin{equation*}
   \bm{0}  = x_{1 } \frac{\sum_{j: x_j = x_1} \beta_j}{|{j: x_j = x_1}|} - x_{1 } \frac{\sum_{j: x_j = x_1} \beta_j}{|{j: x_j = x_1}|} = \be_1\,,
\end{equation*}
    a contradiction. \qed
\end{remark}

The following remark provides more discussion on the exponential dependence of $D$ on $K$ in \cref{prop:theta_gap}. 

\begin{remark}\label{rem_rate}  
    Although we have stated \cref{prop:theta_gap} without explicit constants, the dependence on $K$, $B$, $\Delta_1$, and $\underline{\alpha}$ can be extracted from the proof.
	In particular, see, for instance,  \eqref{badink}, the  dependence on $\Delta_1$ is poor:  the constant  $D$ can be shown to scale as $\Delta_1^{-c K}$ for some absolute constant $c$.
 While it is possible that the exponent can be improved, the exponential dependence of this constant on $K$ cannot be entirely avoided, even for univariate mixtures ($L = 1$). 
     This follows from the fact that, when $L = 1$,  for any $K \geq 2$ and sufficiently small $\Delta_1> 0$, there exist a pair of $K$-atomic probability measures $\rho$ and $\rho'$ on $[-1, 1]$ with support $\{\btheta_1, \dots, \btheta_K\}$ and $\{\btheta'_1, \dots, \btheta'_K\}$, all of which are separated by at least $\Delta_1$, and such that
     \begin{equation}\label{eq:sep_lb}
         \|\btheta'_k - \btheta_k\|_2 \geq C_K \Delta_1^{-2K+2} \|\bm m- \bm m'\|_2, \quad \quad \text{for all $k \in [K]$}
     \end{equation}
     where $\bm m$ and $\bm m'$ are the vectors of the first $2K-1$ moments of $\rho$ and $\rho'$, respectively.
     We prove  this fact below.
     This example shows that any deterministic bound on the distance between the atoms in terms of the moment difference for $K$-atomic distributions with well separated atoms must involve a prefactor of the same type as appears in (\ref{eq:sep_lb}).
     Since \cref{prop:theta_gap} is a bound of this type, we conclude that the $\Delta_1^{-cK}$ scaling of $D$ is essentially tight.
\end{remark} 

\begin{proof}[Proof of \cref{rem_rate}]
    Fix $K \geq 2$. \citet[Lemma 18]{WuYan20} implies that there exist two $K$-atomic distributions $\nu$ and $\nu'$ on $[-1, 1]$ whose first $2K-2$ moments match; moreover, these distributions are supported on the maxima and minima, respectively, of $P^* - f^*$, where $f^*$ and $P^*$ are solutions to a particular saddle point problem involving uniform polynomial approximation of Lipschitz functions on $[-1, 1]$.
    In particular, the atoms of $\nu$ and $\nu'$ are all separated from each other by some $c_K > 0$, and, since each distribution is supported on $[-1, 1]$ and the first $2K-2$ moments match, the moment vectors satisfy
    \begin{equation*}
        \|\bm m(\nu) - \bm m(\nu')\|_2 \leq 2\,.
    \end{equation*}
    Now, denote by $\rho$ and $\rho'$ the image of $\nu$ and $\nu'$ under the dilation $x \mapsto \tfrac{\Delta}{c_K} x$.
    Note that the atoms of $\rho$ and $\rho'$ are now all separated from each other by at least $\Delta$; moreover, since $\rho$ and $\rho'$ differ only in their $(2K-1)$th moment,  
    \begin{equation*}
        \| \bm m(\rho) - \bm m (\rho')\|_2 \leq 2 \left(\frac{\Delta}{c_K}\right)^{2K-1} \,.
    \end{equation*}
    Letting $\{\btheta_1, \dots, \btheta_K\}$ and $\{\btheta_1', \dots, \btheta_K'\}$ denote the support of $\rho$ and $\rho'$ respectively, we obtain
    \begin{equation*}
        \min_{k \in [K]} \|\btheta_k - \btheta_k' \|_2 \geq \Delta  \geq \tfrac 12 c_K^{2K-1} \Delta^{-2K+2}  \|\bm m - \bm m'\|_2\,,
    \end{equation*}
    as desired.
\end{proof}

\begin{remark}\label{non-param}
The parametric-type rates of \cref{thm_mom_est}  hold when the mixture atoms are well separated. As pointed out by \citet{WuYan20}, since $\wh {\bm m}$ after projection in \eqref{def_dmm} belongs to $\cM$,   the $K$-atomic measure $\wh \rho_1$ defined by
$	\wh \rho_1 = \sum_{k=1}^K \wh \alpha_k \delta_{ \wh \theta_{1k}}$
is a valid probability distribution on $[-B, B]$  whose moments satisfy $M_r(\wh \rho_1) = \wh m_r$ for $1\le r \le 2K-1$.
The measure $\wh \rho_1$ therefore estimates the univariate measure $\rho_1^* := \sum_{k=1}^K \alpha_k^* \delta_{\theta_{1k}^{*}}$, which is the projection of the mixing measure $\rho^*$ onto its first coordinate, and whose moments satisfy $M_r(\rho_1^*) = m_r$.
 
In  particular, the proof of \cref{prop:theta_gap}  reveals that its conclusion holds when each atom of  $\wh \rho_1$ and $\rho_1^*$  is at least $\Delta_1/2$ away  from all but $\ell'= 1$  other atom (itself). However, if $\ell' > 1$, Proposition 4 in \cite{WuYan20} shows that we cannot expect a  parametric rate in the estimation of $\btheta_k$, even in one dimension, as display (\ref{expK}) in the proof then becomes
   \begin{equation}\label{expK-nonparam}
W_1(\wh \rho_1, \rho_1^*) \leq 2K \left({2K 4^{2K - 1} 2^{2K - \ell' -1} \over \Delta_1^{2K - \ell'- 1}}\right)^{1 \over \ell'}\|\wh  {\bm m}  - {\bm m} \|_2^{1\over \ell'},
 \end{equation}
a rate that will be inherited by $|\theta_{1k}^{*} - \wh \theta_{1k}|$, for each $k$,  via (\ref{perm}). In the worst case, when $\ell' = 2K-1$, we obtain 
$    W_1(\wh \rho_1, \rho_1) \lesssim K \|\wh \bmvec - \bmvec\|_2^{1/(2K-1)},$
by Proposition 1 in \cite{WuYan20}. Thus, although consistent estimation of the softmax mixture parameters will continue to hold when the atoms are distinct, but not well separated,  neither the estimation of $\btheta_k^*$ nor that of $\balpha^*$ can be expected to follow a parametric decay rate. This is confirmed by our simulation results in  Section~\ref{sec_sims}. 
\end{remark}

\subsection{Subspace estimation under elliptically symmetric distributions}
\label{sec:subspace-elliptical}
In this section we explain how to recover the subspace $\text{span}(\btheta_1,\ldots,\btheta_K)$ without knowing $\mu$ explicitly, 
from the sample matrices $n^{-1}\sum_{i=1}^nY_iY_i^\top$ and $ p^{-1} \sum_{i=1}^p X_i X_i^\T$ only. We assume that \cref{ass_alpha,separation,ass_theta,random-points} hold and the additional key requirement is that $\mu$ has
a centered elliptically symmetric density 
\[
f_{\mu}(x)
=
c\,|\Sigma |^{-1/2}
g\!\left(x^\top \Sigma ^{-1}x\right),
\qquad x\in\mathbb R^L
\]
where $\Sigma \succ0$ is the scatter matrix and $g\ge0$ is the radial
generator.
Since $\mu$ is sub-Gaussian,  the moment generating function
of $X$ is finite.
For elliptically symmetric distributions, the following
standard facts \cite[Chapter~2]{FangKotzNg1990} then apply:
\begin{align}\label{basic-ellip}
\Cov(X)=\kappa \Sigma \quad \text{ 
 and } \quad
\EE\!\left[\exp({\btheta^\top X})\right]=
\varphi(\btheta^\top \Sigma \btheta),
\end{align}
 for some $\kappa>0$ and 
scalar function $\varphi$, which satisfies
\begin{align}
    \label{diag-ellip}
\text{$\varphi(q)>0$ and  $\varphi''(q)>0$ for all $q$}
\end{align}
The key identity, proved below,
\begin{align}
\label{KEY-ellip}
\frac{\EE[XX^\top \exp({X^\top\btheta})]}
{\EE[ \exp({X^\top\btheta})]}
&= 2\frac{\varphi'(q)}{\varphi(q)}\Sigma +
4\frac{\varphi''(q)}{\varphi(q)} \Sigma \btheta\btheta^\top \Sigma 
\end{align} with $q:=\btheta^\top \Sigma \btheta$,
shows that the Gaussian second-moment construction \eqref{def_Sigma_Y} 
has a
natural elliptical analogue.
Indeed, we expect,  for  large $N$,
\[
\frac1n\sum_{i=1}^n Y_iY_i^\top
=
\sum_{j=1}^p \wh\pi_j X_jX_j^\top
\]
 to be close to its conditional expectation
\[
\sum_{j=1}^p \pi_j X_jX_j^\top
= \sum_{k=1}^K \alpha_k \frac{ \sum_{j=1}^p X_jX_j^\top \exp(\btheta_k^\top X_j) }{ \sum_{j=1}^p \exp(\btheta_k^\top X_j)},
\]
which, in turn, for large $p$,   is naturally
approximated by  
\[
\Lambda:= \sum_{k=1}^K\alpha_k\frac{\EE[XX^\top\exp(X^\top\btheta_k)]}{\EE[\exp(X^\top\btheta_k)]}.
\]
 Let
\[
q_k:=\btheta_k^\top \Sigma \btheta_k,
\qquad
a:=2\sum_{k=1}^K\alpha_k\frac{\varphi'(q_k)}{\varphi(q_k)},
\qquad
B_k= 4\alpha_1\frac{\varphi''(q_k)}{\varphi(q_k)}
\]
and define $B=\diag(B_1,\ldots,B_K)$.
Using  identity \eqref{KEY-ellip} at $\btheta=\btheta_k$ in the definition of $\Lambda$ gives 
\begin{align}\label{algebra-ellip}
\Lambda=a\Sigma +\Sigma \btheta B\btheta^\top \Sigma 
\end{align}
after a little algebra given below. Since
$\alpha_k>0$, 
$\varphi(q_k)>0$, and $\varphi''(q_k)>0$,
every diagonal
entry of $B$ is strictly positive. Now let
$C:=\Cov(X)=\kappa \Sigma $ and \(\btheta :=[\btheta_1,\ldots,\btheta_K]\).
Then
\[
\Lambda=\frac{a}{\kappa}C+\frac{1}{\kappa^2}C\btheta B\btheta^\top C
\] and its whitened version takes the form
\[
W:=C^{-1/2}\Lambda C^{-1/2}=\frac{a}{\kappa}I_L+\frac{1}{\kappa^2}C^{1/2}\btheta B\btheta^\top C^{1/2}.
\]
The first term is a scalar multiple of the identity and therefore does not affect the eigenvectors.  Because $B\succ0$ and
$\rank(\btheta)=K$, the second term is positive semidefinite of rank $K$.  Therefore the $K$ leading eigenvectors
$u_1,\ldots,u_K$ of $W$ are precisely the leading eigenvectors of
\( C^{1/2}\btheta B\btheta^\top C^{1/2} \)
and 
\[
\textrm{span}(\btheta_1,\ldots,\btheta_K)=\textrm{span}(C^{-1/2}u_1,\ldots,C^{-1/2}u_K).
\]
We propose the following procedure:
\begin{itemize}
\item
Estimate $\Lambda$, $C$ and $W$ by
\[
\wh\Lambda=\frac1n\sum_{i=1}^nY_iY_i^\top,\qquad\wh C=\frac1p\sum_{j=1}^pX_jX_j^\top,\qquad
\wh W=\wh C^{-1/2}\wh\Lambda\,\wh C^{-1/2},
\] respectively.
\item Compute the $K$ leading eigenvectors $\wh u_1,\ldots,\wh u_K$ of
$\wh W$.
 \item Estimate 
 the parameter subspace  by
\(
\textrm{span}(\wh C^{-1/2}\wh u_1,\ldots,\wh C^{-1/2}\wh u_K).
\)
\end{itemize}
This  strategy recovers the
span of $\btheta_1,\ldots,\btheta_K$ without requiring knowledge of the
generator $g$, the radial function $\varphi$, or the proportionality
constant $\kappa$.  These unknown quantities affect the eigenvalues of
$W$, but not its informative eigenspace.

In case $\mu=\cN_L(0,\Sigma)$, we find that $a=\kappa=1$ and $B=\diag(\alpha_1,\ldots,\alpha_K)$. Moreover,
under the same conditions as in \cref{prop_eigensp}
in addition to the elliptical distributional assumption,
the same conclusion holds for 
$\| \wh W - W\|_\op$. The proof is almost identical and is therefore omitted.

\subsubsection*{Proof of \eqref{basic-ellip} and \eqref{diag-ellip}}

We refer to \citet[Chapter~2]{FangKotzNg1990}.  
Since $Z:=\Sigma ^{-1/2}X$ has density $c g(\|z\|^2)$,
it is  spherically symmetric: $QZ\overset{d}=Z$ for every orthogonal matrix $Q$.
This implies $\Cov(Z)=Q\,\Cov(Z)\,Q^\top$ for every orthogonal $Q$ and this forces $\Cov(Z)=\kappa I_L$ for some $\kappa > 0$. Hence $\Cov(X)=\Sigma ^{1/2}\Cov(Z)\Sigma ^{1/2}=\kappa \Sigma$, which  
 proves the first part of \eqref{basic-ellip}.
 
The spherical symmetry of $Z$ ensures as well that 
   $M_X(\btheta)=\EE\!\left[\exp({(\Sigma ^{1/2}\btheta)^\top Z})\right]$
 is invariant under orthogonal transformations and therefore depends on $\Sigma ^{1/2}\btheta$ only through $q= \|\Sigma ^{1/2}\btheta\|^2=
\btheta^\top \Sigma \btheta$. Hence 
$M_X(\btheta)= \varphi(q)$ for some scalar function $\varphi$.
This proves the second part of \eqref{basic-ellip}.

We will now prove $\varphi(q)>0$ and $\varphi''(q)>0$.
Let $v:= \Sigma^{1/2} \btheta$ so that $q=\|v \|^2$ and $M_X(\btheta) = \EE\!\left[\exp({v^\top Z})\right]$, and let $Q$ be an orthogonal matrix with $Qv= \sqrt{q} e_1$. This means that $v^\top Z= (Qv)^\top QZ \overset{d}= \sqrt{q} e_1^\top Z  \overset{d}= \sqrt{q} e_1^\top (-Z )$. Hence,
\[
\varphi(q)
= \EE\!\left[\exp({v^\top Z}) \right] = \EE[\exp[{\sqrt q\,Z_1}] = \EE[\cosh(\sqrt q\,Z_1)]
\]
 for $Z_1:= e_1^\top Z$.
A Taylor expansion yields \[
\varphi(q)=\EE[\cosh(\sqrt q\,Z_1)]=\sum_{m=0}^{\infty}\frac{\EE[Z_1^{2m}]}{(2m)!}\,q^m \ge 1
\]
and  
\[
\varphi''(q)=\sum_{m=2}^{\infty}\frac{m(m-1)\EE[Z_1^{2m}]}{(2m)!}\, q^{m-2} \ge \frac{\EE[Z_1^4]}{12} >0.\]
This proves \eqref{diag-ellip}. \qed

\subsubsection*{Proof of \eqref{KEY-ellip}  and  \eqref{algebra-ellip}}
We observe that
$M_X(\btheta)=\EE[\exp{(X^\top\btheta)}]=\varphi(q)$, with
$q=q(\btheta) :=\btheta^\top \Sigma \btheta$, and $\EE[XX^\top \exp({X^\top\btheta})] =\nabla_{\btheta}^2 M_X(\btheta)$. Hence,
\[
\begin{aligned}
\EE[XX^\top \exp({X^\top\btheta})] 
&= \nabla_{\btheta} ^2 \varphi(q) \\
&=
\nabla_{\btheta}\left(2\varphi'(q)\Sigma \btheta\right) \\
&=
2\varphi'(q)\Sigma +2\Sigma \btheta\left(\nabla_{\btheta}\varphi'(q)\right)^\top\\
& =2\varphi'(q)\Sigma +
2\Sigma \btheta\left( 2\varphi''(q)\Sigma \btheta\right)^\top\\
&=
2\varphi'(q)\Sigma +4\varphi''(q)\Sigma \btheta\btheta^\top \Sigma .
\end{aligned}\]
This proves \eqref{KEY-ellip}.
The derivation of the representation \eqref{algebra-ellip} of $\Lambda$ follows from 
\eqref{KEY-ellip}:  
\begin{align*}
\Lambda
&:=\sum_{k=1}^K\alpha_k\frac{\EE[XX^\top \exp({X^\top\btheta_k})]}{\EE[\exp({X^\top\btheta_k})]}\\
&=\sum_{k=1}^K\alpha_k\left[2\frac{\varphi'(q_k)}{\varphi(q_k)}\Sigma +4\frac{\varphi''(q_k)}{\varphi(q_k)}\Sigma \btheta_k\btheta_k^\top \Sigma \right]  &&\text{with $q_k:=\btheta_k^\top \Sigma \btheta_k$} \\
&=2\left(\sum_{k=1}^K\alpha_k\frac{\varphi'(q_k)}{\varphi(q_k)}\right)\Sigma  +\Sigma \left[\sum_{k=1}^K4\alpha_k\frac{\varphi''(q_k)}{\varphi(q_k)}\btheta_k\btheta_k^\top\right]\Sigma \\
&= a\Sigma +\Sigma \btheta B\btheta^\top \Sigma 
\end{align*}
with $a:=2\sum_{k=1}^K\alpha_k{\varphi'(q_k)}/{\varphi(q_k)}$ and   $B={\rm diag}(B_{11},\ldots,B_{KK})$ with  $B_{kk}:=4\alpha_k{\varphi''(q_k)}/{\varphi(q_k)}>0$, $k\in [K]$. \qed


\subsection{Proof of \cref{crux}}\label{app_proof_crux}

Let $(v,w_2,\ldots, w_L)$ be an orthonormal basis of $\RR^L$. 
We first prove the following lemma for such basis. Then \cref{req_h_hi} of Proposition \ref{crux} follows from \cref{crux-lemma} by choosing the canonical basis. \cref{cond-mom} is an immediate consequence of the strong law of large numbers (SLLN). 

Similar to \eqref{eq:hr_def} and \eqref{eq:hr1_def}, define  
\begin{equation}\label{eq:hr_def_v}
	h_r(x; v) := (-1)^r f_\mu(x)^{-1} \frac{\rd^r}{\rd t^r} f_\mu(x + tv)|_{t = 0}\,.
\end{equation}
and
\begin{align}\label{eq:hr1_def_w}\nonumber
	h_{r1;i}(x; v,w_i) &:= \frac{1}{r+1} \nabla_v h_{r+1} (x)^\T w_i\\
    &= (-1)^{r+1} f_\mu(x)^{-1}  \frac{\rd^{r+1}}{\rd t^r \rd s} f_\mu(x + tv + sw_i)|_{t, s = 0}, \, 
\end{align}
for $i \in \{2, \ldots, L\}$.

\begin{lemma}\label{crux-lemma}
Let $X \sim \mu$ with $\mu$ satisfying Assumption \ref{ass:mu}. Then,  for any $\btheta \in \RR^L$,
\begin{align}
		\EE_\mu\left[  h_r(X;v) \exp(X ^\T \btheta)\right] \over \EE_\mu\left[ \exp(X^\T \btheta) \right]  &= (v^\T \btheta)^r\, \label{Approx-E-gen} \\
		\EE_\mu\left[  h_{r1;i}(X;v,w_i) \exp(X^\T \btheta)\right] \over \EE_\mu\left[ \exp(X^\T \btheta) \right] &= (v^\T \btheta)^r (w_i^\T \btheta)\,,  \label{Approx-E2-gen}
\end{align}
for $i \in \{2, \ldots, L\}$.
\end{lemma}

\begin{proof}
	It suffices to prove the first claim, since the second follows by differentiating both sides of \eqref{Approx-E-gen} with respect to $v$ and applying dominated convergence.
	
	Write $g_{r}(X; v, t) = (-1)^r f_\mu(X)^{-1}\frac{\rd^r}{\rd t^r} f_\mu(X + tv) \exp(X^\T \btheta)$.
	We will show by induction that
	\begin{equation}\label{eq:hypo}
		\EE_\mu\left[g_r(X; v, t)\right] = \left(v^\T \btheta\right)^r \EE_\mu[\exp((X-tv)^\T \btheta)]
	\end{equation}
	for all $t \in \RR$ and $v \in \RR^L$, and conclude by taking $t = 0$.
	
	When $r = 0$, we have
	\begin{align*}
		\EE_\mu\left[g_0(X; v, t)\right] & = \int f_\mu(x + tv) \exp(x^\T \btheta) \dd x \\
		& = \int f_\mu(x) \exp((x-tv)^\T \btheta) \dd x \\
		& = \EE_\mu[\exp((X-tv)^\T \btheta)]\,.
	\end{align*}
	
	Now assume \eqref{eq:hypo} holds for a natural number $r$.
	The assumption that the partial derivatives of $f_\mu$ decay super-exponentially implies that we can apply dominated convergence to obtain
	\begin{align*}
		\EE_\mu\left[g_{r+1}(X; v, t) \right] & = - \EE_\mu\left[\frac{\rd}{\rd t}g_{r}(X; v, t) \right] \\
		& = - \frac{\rd}{\rd t}  \EE_\mu\left[g_{r}(X; v, t) \right] \\
		& = - \frac{\rd}{\rd t} (v^\T \btheta)^r \EE_\mu[\exp((X-tv)^\T \btheta)] \\
		& = (v^\T \btheta)^{r+1} \EE_\mu[\exp((X-tv)^\T \btheta)]\,.
	\end{align*}
    The proof is complete.
\end{proof}

\subsection{Proof of \cref{prop:theta_gap}}\label{app_proof_prop:theta_gap}

\begin{proof} The proof is divided into two parts.  

\noindent {\bf Proof of $(i)$.} We first  show  that the bound holds for the first coordinate. By re-scaling, we may assume $B = 1$. To this end, we use  existing results   in  ~\citet{WuYan20}. 
 To begin with, we recall here   \ref{cond_Delta1}, and fix $\Delta_1$ and $\ua$. Define $$\epsilon : = {\Delta_1\ua \over 4}$$ 
 and write 
 $\rho_1^* = \sum_k\alpha_k^*\delta_{\theta_{1k}^*}$ and $\wt\rho_1 = \sum_k\bar\alpha_k\delta_{\bar \theta_{1k}}.$
 By Proposition 1 in \cite{WuYan20}, 
 there exists $c' = c'(K)$ such that, if $\|\wt {\bm m}  - {\bm m} \|_2 \leq c'$, then $W(\rho_1^*, \wt \rho_1) \leq \epsilon$. 
 We will show that our result holds by considering, separately,  $\|\wt {\bm m}  - {\bm m} \|_2 \leq c'$ and $\|\wt  {\bm m}  - {\bm m} \|_2 >  c'$. We begin with the former.  

 We show that $\|\wt  {\bm m}  - {\bm m} \|_2 \leq c'$, for  $c'$ above,  implies that:
 \begin{itemize}
     \item [(i)] There exists a permutation $\varrho$ of  $K$  integers such that 
\begin{equation}\label{perm}
  | \theta_{1k} - \bar \theta_{1\varrho(k)}| \leq W_1(\wt \rho_1, \rho_1^*)/\ua, \ \ \mbox{for each} \ k \in [K]. 
\end{equation}
    \item [(ii)] There exists a constant $C$ depending on $K$ and $\Delta_1$ such that 
\begin{equation}\label{WY}
 W_1(\wt \rho_1, \rho_1^*) \leq C  \|\wt  {\bm m}  - {\bm m} \|_2.
\end{equation}
 \end{itemize}
\noindent The claimed result, in this case,  will then follow by combining (\ref{perm}) and (\ref{WY}).

By the definition of the Wasserstein distance, and for $\Pi$ denoting a distribution with marginals $\wt \rho_1$ and $\rho_1^*$, we have 
		\begin{align}\label{inter}
			W_1(\wt \rho_1, \rho_1^*) &= \inf_{\Pi} \sum_{k,k'} \Pi_{kk'} |\theta_{1k}^* - \bar \theta_{1k'}| \nonumber \\
			&\ge  \sum_{k=1}^K \alpha_k^*  \min_{k'\in [K]} | \theta_{1k}^* -\bar  \theta_{1k'}| \nonumber \\
			& \ge \ua \max_{k\in [K]}\min_{k'\in [K]} | \theta_{1k}^* -\bar  \theta_{1k'}|,
		\end{align}
	and so
 \[ \min_{k'\in [K]} | \theta_{1k}^* -\bar  \theta_{1k'}| \leq {W_1(\wt \rho_1, \rho_1^*) \over  \ua}, \quad \mbox{for each} \ k \in [K]. \]
Then,  there must exist a permutation $\varrho$ such that \eqref{perm} holds. 
Otherwise, suppose there exists some $ \varrho(k) = \varrho(k')$ for some $k\ne k'$ such that 
		\[
		|\theta_{1k}^* - \bar \theta_{1\varrho(k)}| \le W_1(\wt \rho_1, \rho_1^*) / \ua,\qquad |\theta_{1k'}^* - \bar \theta_{1\varrho(k')}| \le W_1(\wt \rho_1, \rho_1^*) / \ua.
		\]
		This   however leads to the  contradiction 
		\[
		\Delta_1  \le |\theta_{1k}^*- \theta_{1k'}^*| \le 	|\theta_{1k}^* - \bar \theta_{1\varrho(k)}| + |\theta_{1k'}^* - \bar \theta_{1\varrho(k')}| \le  {2W_1(\wt \rho_1, \rho_1^*) \over  \ua} \le   {2\epsilon \over  \ua} \leq {\Delta_1 \over 2},
		\]
	where the penultimate inequality uses $W(\rho_1^*, \wt \rho_1) \leq \epsilon$ from  $\|\wt  {\bm m}  - {\bm m} \|_2 \leq c'$ and the last inequality follows by the definition  of $\epsilon$.	 This proves (\ref{perm}).  
 


 To show (\ref{WY}), without loss of generality, we assume $\varrho$ is the identity permutation. We first notice that 
 \begin{equation} \label{hat-dif} 
    \min_{k \neq k'}|\bar \theta_{1k} - \bar \theta_{1k'}| \geq \min_{k \neq k'}|\theta_{1k}^* -  \theta_{1k'}^*| - 2\max_k |\theta_{1k} - \bar \theta_{1k}|  \ge \Delta_1 - {2\epsilon \over \ua} \ge  {\Delta_1 \over 2}
\end{equation} 
and, similarly, 
\[
\min_{k \neq k'}|\theta_{1k}^* - \bar \theta_{1k'}| \geq \min_{k \neq k'}|\theta_{1k} -  \theta_{1k'}| - \max_k |\theta_{1k} - \bar \theta_{1k}|  \ge \Delta_1 - {2\epsilon \over \ua} \ge  {\Delta_1 \over 2}
\]
 Thus,  the atoms of $\wt \rho_1$ and $\rho_1^*$ are all  separated by at least $\Delta_1 /2$. This places us in the setting of Proposition 4 in \cite{WuYan20},  which we apply (relative to their notation) with $\gamma = \Delta_1/2$, $l = 2K$ and $\ell' = 1$ yielding the bound 
 \begin{equation}\label{expK}
 W_1(\wt \rho_1, \rho_1^*) \leq {4K^2 4^{2K - 1} \over \Delta_1^{2K - 2}}\|\wt  {\bm m}  - {\bm m} \|_2,
 \end{equation}
which completes the proof of (\ref{WY}) and  thus, for all $k \in [K]$  
\[ |\bar \theta_{1k} - \theta_{1\varrho(k)}^*| \leq C'\|\wt  {\bm m}  - {\bm m} \|_2,  \] 
 by taking 
	\begin{equation} \label{badink} 
    C' := {4K^2 4^{2K - 1} \over \ua \Delta_1^{2K - 2}}.
    \end{equation} 
On the other hand, when $\|\wt {\bm m} - {\bm m}\|_2 > c'$, we have 
\[ | \bar \theta_{1k} - \theta_{1k}^*| \leq 2 < \frac{2}{c'}\|\wt {\bm m} - {\bm m}\|_2 \leq D_1\|\wt {\bm m} - {\bm m}\|_2,\]
for $D_1 := \max\{2/c', C'\}$, and where the first inequality holds since,  for each $k\in [K]$, we have  $\bar \theta_{1k}, \theta_{1k} \in [-1, 1]$. 

Therefore, for all $k \in [K]$, there exist a constant $D_1$ as above such that 
\begin{equation}\label{theta1-det}
 |\bar \theta_{1k} - \theta_{1k}^*| \leq D_1\|\wt {\bm m} - {\bm m}\|_2
\end{equation}

 We  next fix $i\in \{2, \ldots, L\}$ and $k\in [K]$  and show that  estimation error of the remaining coordinates  $\theta_{ik}^*$ has upper bound similar to  (\ref{theta1-det}), for a different constant $D_2$.  Recall that
  \begin{eqnarray*}
     && \bmvec_{1;i} = \left(  m_{01;i},  \ldots,   m_{(K-1)1; i} \right)^\T\qquad  \bar \bmvec_{1;i}=  \left( \bar m_{01;i},  \ldots,  \bar  m_{(K-1)1; i} \right)^\T 
 \end{eqnarray*}  
 and let 
    \begin{eqnarray*} 
    \xi := \left( 1,  \theta_{1k}, \ldots ,   \theta_{1k}^{K-1}\right)^\T,  \qquad  \bar \xi:=\left( 1,  \bar \theta_{1k}, \ldots ,  \bar \theta_{1k}^{K-1}\right)^\T\,.
  \end{eqnarray*}
Finally, define the operator $\mathsf{clip}_B$ by
\begin{equation*}
	\mathsf{clip}_B(x) = \begin{cases}
		-B & \text{if $x < -B$} \\
		x & \text{if $|x| \leq B$} \\
		B & \text{if $x > B$.}
	\end{cases}
\end{equation*}
Using the definition of $\bar \theta_{ik}$ and~\eqref{rest-coordinates}, we can therefore write
\begin{align}
    \bar \theta_{ik} - \theta_{ik}^* &= \mathsf{clip}_B(\bar\bmvec_{1;i}^{\T} \wt M^{\dagger} \bar\xi) - \bmvec_{1;i}^{\T} M^{-1} \xi\,,  \nonumber
\end{align}
where the  $K \times K$ matrix $\wt M$ is obtained from 
\[ 
    \wt M :=  \begin{pmatrix}
	1 &  \wt m_1 & \dots &  \wt m_{K-1} \\
	\wt m_1 &  \wt m_2 & \dots &  \wt m_{K} \\
	\vdots & \vdots & & \vdots \\
	\wt m_{K-1} & \wt  m_K & \dots &  \wt m_{2K-2} 
\end{pmatrix}.  
\]

We consider two cases.
As in the preceding argument, there exists a constant $c'$ depending on $K$, $\Delta_1$, $B$, and $\ua$ such that if $\|\wt {\bm m} - \bm m\|_2 \leq c'$, then the measure $\wt \rho_1$ corresponding to $\wt {\bm m}$ has $K$ atoms, each separated by at least $\Delta_1/4$.

Under this scenario, \citet[Theorem 2A]{Lin89} implies that $\wt M$ is invertible,
and we obtain
\begin{align}\label{rest-rate}
	|\bar \theta_{ik} - \theta_{ik}^*| &= |\mathsf{clip}_B(\bar\bmvec_{1;i}^{\T} \wt M^{-1} \bar\xi) - \bmvec_{1;i}^{\T} M^{-1} \xi|  \nonumber \\
	& \leq |\bar\bmvec_{1;i}^{\T} \wt M^{-1} \bar\xi - \bmvec_{1;i}^{\T} M^{-1} \xi| \nonumber \\
	& \leq |\bar\bmvec_{1;i}^{\T} (\wt M^{-1} - M^{-1} )\bar \xi| + |(\bar \bmvec_{1;i} - \bmvec_{1;i})^\T M^{-1} \xi| + |\bmvec_{1;i}^{\T} M^{-1} (\bar \xi -\xi)|
\end{align}

By \citet[Theorem 2A]{Lin89}, $M$ is invertible, and $\| M^{-1}\|_{\rm op}$ is bounded by a constant depending on $K$, $\Delta_1$, $B$, and $\ua$.
Next, recall that  we work under the assumption that $\| \btheta_k^*\|_2 \leq B$, for all $k$, for some constant $B$ and thus, as $K$ is fixed, both $\| \bmvec_{1;i}\|_2$ and $\|  \xi\|_2$ are of order $\cO(1)$.

Then, an application of the Cauchy-Schwarz inequality and~\eqref{theta1-det} shows that the last two terms in (\ref{rest-rate}) are bounded by a constant multiple of $\|\bar \bmvec_{1;i} - \bmvec_{1;i}\|_2 + \|\wt {\bm m} - \bm m\|_2$.

For the first term in~\eqref{rest-rate}, we note that 
\begin{equation}\label{last-b}
|\bar\bmvec_{1;i}^{\T} (\wt M^{-1} - M^{-1} )\bar \xi|  \leq \|\bar \bmvec_{1;i}\|_2 \|\bar \xi\|_2 \|\wt M^{-1} - M^{-1}\|_{\rm op}\,.
\end{equation}
The norms $\|\bar \bmvec_{1;i}\|_2$ and $\|\bar \xi\|_2$ are both bounded by a constant depending on $B$ and $K$.
Furthermore,
\begin{align*}
	\| \wt M^{-1} - M^{-1} \|_{\rm op} &=   \| M^{-1} ( M  -\wt  M) \wt M^{-1} \|_{\rm op} \le \| M^{-1}\|_{\rm op} \| \wt M^{-1}\|_{\rm op} \|    M  -\wt  M \|_{\rm op}\,.
\end{align*}
As noted above, the assumption $\|\wt {\bm m} - \bm m\|_2 \leq c'$ implies that $\wt \rho_1$ has $K$ atoms separated by at least $\Delta_1/4$; this implies that $\|\wt M^{-1}\|_{\rm op}$ is also bounded by a constant depending on $K$, $\Delta_1$, $B$, and $\ua$.
We obtain, for a constant $C'$ different than above
\begin{equation*}
	\| \wt M^{-1} - M^{-1} \|_{\rm op} \leq C' \|\wt {\bm m} - \bm m\|_2\,.
\end{equation*}
All together, we obtain that when $\|\wt {\bm m} - \bm m\|_2 \leq c'$, we have the bound
\begin{equation}
	|\bar \theta_{ik} - \theta_{ik}^*| \leq C' (\|\bar \bmvec_{1;i} - \bmvec_{1;i}\|_2 + \|\wt {\bm m} - \bm m\|_2)\,.
\end{equation}

On the other hand, if $\|\wt {\bm m} - \bm m\|_2 > c'$, then the same argument as was given above shows that
\begin{equation}
	|\bar \theta_{ik} - \theta_{ik}^*| \leq 2B < D_2(\|\bar \bmvec_{1;i} - \bmvec_{1;i}\|_2 + \|\wt {\bm m} - \bm m\|_2)\,,
\end{equation}
where $D_2 := \max\{C', 2B/c'\}$.

Finally, to establish the desired bound on $\bar {\bm \alpha}$, we use a very similar argument.
Let $T$ be the Vandermonde matrix appearing on the right side of \eqref{weights}, and $\bar T$ its empirical counterpart in~\eqref{weights-empirical}.
If $\|\wt {\bm m} - \bm m\|_2 \leq c'$, then $T$ and $\bar T$ are both invertible, with smallest singular value bounded away from zero.
We obtain, for some other constant $C'$
\begin{equation}
	\|\bar {\bm \alpha} - \bm \alpha^*\|_2 \lesssim \|T - \bar T\|_{\rm op} + \|\wt {\bm m} - \bm m\|_2 \leq C'\|\wt {\bm m} - \bm m\|_2\,.
\end{equation}
When $\|\wt {\bm m} - \bm m\|_2 > c'$, we use the trivial bound
\begin{equation}
	\|\bar {\bm \alpha} - \bm \alpha^*\|_2 \leq 2 < D_3\|\wt {\bm m} - \bm m\|_2
\end{equation}
for $D_3 = \max\{C', 2/c'\}$. Since $\|\wt{\bm  m} -  {\bm m} \|_2  
   \le  \| \bar {\bm m} - {\bm m}\|_2$
        from \eqref{def_dmm},
taking $D = D_1\vee  D_2 \vee D_3$  completes the argument.

{\bf Proof of $(ii)$.} In view of $(i)$, it remains to bound $|\bar m_r-m_r|$ and $|\bar m_{r1;i}-m_{r1;i}|$ for $r\le 2K$ and $2\le i\le L$. For each $\bar m_r$, note that 
\[
    \bar m_r -  m_r  =  \sum_{k=1}^K \alpha_k^* \left( {{1\over p}\sum_{j=1}^p h_r(X_j) \exp(X_j^\T \btheta_k^*) \over {1\over p}\sum_{\ell = 1}^p \exp(X_\ell^\T \btheta_k^*)}-{\EE_\mu[h_r(X) \exp(X^\T \btheta_k^*)] \over  \EE_\mu[\exp(X^\T \btheta_k^*)]}\right).
\] 
By applying the law of the iterated logarithm to both the numerator and the denominator within the parentheses, their differences from the corresponding population quantities are each of order $\cO(\sqrt{\log\log p/p})$, $\mu$-a.s. Since, by Jensen’s inequality, $\EE_\mu[\exp(X^\T \btheta_k^*)] \ge \exp(\EE_\mu[X^\T \btheta_k^*]) = 1$, we obtain $|\bar m_r-m_r| = \cO(\sqrt{\log\log p/p})$ $\mu$-a.s.. Similar arguments can be used for bounding $|\bar m_{r1;i} - m_{r1;i}|$. Applying the same argument to the remaining finitely many moments and intersecting the corresponding probability-one events completes the proof. 
\end{proof}

\subsection{Proof of  \cref{thm_mom_est}}\label{app_sec_proof_thm_mom_est}
 \begin{proof} 
 From \cref{prop:theta_gap}, it suffices to show that for $\mu$-almost every path $X_1,\ldots, X_p$ and for all $p$ sufficiently large, 
 \[
     \max_{r<2K} | \wh m_r - \bar m_r | + \max_{r<K, 2\le i\le L} | \wh m_{r1;i} - \bar m_{r1:i} | = \cO(\sqrt{\log\log N/N})
 \] 
 $\pi$-almost surely.  We only prove for $\max_{r<2K}| \wh m_r - \bar m_r |$ as similar arguments can be used to bound $\max_{r<K, 2\le i\le L} | \wh m_{r1;i} - \bar m_{r1:i} |$. 
For every integer $r<2K$, note that 
\[
    \wh m_r - \bar m_r = {1\over N}\sum_{i=1}^N \Bigl(h_r(Y_i) - \EE_{\pi^*}[h_r(Y_i) \mid \bX]\Bigr) =: {1\over N}\sum_{i=1}^N S_i.
\]
and
\[
    \var_{\pi^*}[S_i^2] \le \EE_{\pi^*}[h_r(Y)^2 \mid \bX] = \sum_{k=1}^K\alpha_k^* {{1\over p}\sum_{j=1}^p h_r(X_j)^2\exp(X_j^\T\btheta_k^*) \over {1\over p}\sum_{\ell=1}^p  \exp(X_\ell^\T\btheta_k^*)}.
\]
By the strong law of large number and \cref{ass:mu_moment}, $\var_{\pi^*}[S_i^2]$ is finite  $\mu$-almost surely for all $p$ sufficiently large. As a result, another application of the law of iterated logarithm conditioning on $\bX$ yields 
\begin{eqnarray*}	
  |\wh m_r  - \bar m_r|  = \cO(\sqrt{\log\log N/N}),\qquad \pi-\text{almost surely.}
\end{eqnarray*}
Since there are finite $r< 2K$, the proof is complete.
\end{proof}

 \subsection{Proof of \cref{prop_eigensp}}\label{app_proof_prop_eigensp}
    \begin{proof}
        Pick any $i,j\in [L]$. 
        First the argument for proving \cref{crux} and the Law of Iterated Logarithm (LIL) ensure  that $|\bar \Gamma_{ij} - \Gamma_{ij}| = \cO(\sqrt{\log\log p/p})$, $\mu$-almost surely. It remains to bound 
        $$
            |\wh \Gamma_{ij} - \bar \Gamma_{ij}  |  =   {1\over N}\left|\sum_{\ell = 1}^N [W_\ell]_{ij} \right| ,
        $$
        where we write 
        \[
             W_\ell := { \nabla^2 \mu(Y_\ell)\over \mu (Y_\ell)} - \EE\left[{ \nabla^2 \mu(Y_\ell)\over \mu (Y_\ell)} \mid \bX\right] \in \RR^{L\times L}.
        \]
        Reasoning similarly as in the proof of \cref{thm_mom_est}, by another application of LIL conditioning on $\mu$-almost every $\bX$ for $p$ sufficiently large,  one has 
        $ |\wh \Gamma_{ij} - \bar \Gamma_{ij}  | = \cO(\sqrt{\log\log N/N})$
        $\pi$-almost surely. For finite number of $i,j\in [L]$, the result follows by $\|\wh\Gamma - \Gamma\|_\op \le \sum_{i,j=1}^L |\wh \Gamma_{ij}-\Gamma_{ij}|\le \sum_{i,j=1}^L |\wh \Gamma_{ij}-\bar\Gamma_{ij}| + \sum_{i,j=1}^L |\bar \Gamma_{ij}-\Gamma_{ij}|$. 
    \end{proof}


    \section{Proofs of the results in \cref{sec_synthesis}}\label{app_proof_sec_syn}
    
    \subsection{Proof of \cref{lem_rand_init}}\label{app_sec_proof_lem_rand_init}

    We first give a proof of the following lemma for completeness.

    \begin{lemma}
        Let $\btheta^*$ be any fixed vector in $\bS^{L-1}$ with $L\ge 2$.
        For any $\varepsilon > 0$ and $\delta\in (0,\sqrt{2})$, with probability at least $1-\epsilon$, there exists at least one vector $\bar v$ in independent draws $\{v_1,\ldots, v_m\}$ from $\bS^{L-1}$ such that 
        \begin{equation}\label{init_cap_bound}
             \|\bar v - \btheta^*\|_2 \le \delta
        \end{equation}
        provided that 
        $$
            m \ge  C\sqrt{L}
            \left[
                \delta^2\left(1-\frac{\delta^2}{4}\right)
            \right]^{-(L-1)/2}
            \log(1/\varepsilon),
        $$
        for some absolute constant $C>0$.
    \end{lemma}
   \begin{proof}
        Fix $\btheta^*\in \bS^{L-1}$. Let $v$ be uniformly distributed on $\bS^{L-1}$. By rotational invariance, we may take $\btheta^*=e_1$. Since
        $v= g/\|g\|_2$ in distribution, with $g\sim\cN_L(0,\bI_L)$, 
        we have
        \[
            (v^\T\btheta^*)^2
            =
            \frac{g_1^2}{\sum_{j=1}^L g_j^2}\sim
            \operatorname{Beta}\left(\frac12,\frac{L-1}{2}\right).
        \]
        The last step uses the fact that  $g_1^2\sim\chi_1^2$ and $\sum_{j=2}^L g_j^2\sim\chi_{L-1}^2$ are independent. 
        By symmetry of the distribution of $v^\T\btheta^*$ and a change of variables, the density of $v^\T\btheta^*$ is
        \[
            f_L(z)
            =
            c_L(1-z^2)^{(L-3)/2},
            \qquad
            c_L
            =
            \frac{\Gamma(L/2)}
                 {\sqrt{\pi}\Gamma((L-1)/2)},
            \qquad z\in(-1,1).
        \]
        For $L\ge 3$, one has
        \begin{align*}
            \PP\left\{v^\T\btheta^*\ge t\right\}
            &=
            c_L\int_t^1(1-z^2)^{(L-3)/2}\,dz\\
            &\ge
            c_L(1+t)^{(L-3)/2}
            \int_t^1(1-z)^{(L-3)/2}\,dz\\
            &=
            \frac{2c_L}{(L-1)(1+t)}
            (1-t^2)^{(L-1)/2}\\
            &\ge
            \frac{c_L}{L-1}
            (1-t^2)^{(L-1)/2}.
        \end{align*}
        For $L=2$, since $c_2=1/\pi$,
        \[
            \PP\left\{v^\T\btheta^*\ge t\right\}
            =
            \frac{\arccos(t)}{\pi}
            \ge
            \frac{\sqrt{1-t^2}}{\pi}
            =
            \frac{c_2}{L-1}(1-t^2)^{(L-1)/2},
        \]
        where we used $\arccos(t)\ge\sqrt{1-t^2}$ for $t\in[0,1)$.
        Thus, for every $L\ge2$,
        \[
            \PP\left\{v^\T\btheta^*\ge t\right\}
            \ge
            \frac{c_L}{L-1}(1-t^2)^{(L-1)/2}.
        \]
        Since $v_1,\ldots,v_m$ are i.i.d., it follows  that, for any $\varepsilon\in(0,1)$,
        \begin{align*}
            \PP\left\{
                v_i^\T\btheta^*<t,~
                \text{for all }i\in[m]
            \right\}
            &=
            \left(
                1-\frac{c_L}{L-1}
                (1-t^2)^{(L-1)/2}
            \right)^m\\
            &\le
            \exp\left\{
                -\frac{mc_L}{L-1}
                (1-t^2)^{(L-1)/2}
            \right\}
            &&\text{by }1-x\le e^{-x}\\
            &\le
            \varepsilon,
        \end{align*}
        provided that
        \[
            m
            \ge
            \frac{L-1}{c_L}
            (1-t^2)^{-(L-1)/2}
            \log(1/\varepsilon).
        \]
        Recall that for any $v,\btheta^*\in\bS^{L-1}$,
        \[
            \|v-\btheta^*\|_2\le\delta
            \quad\Longleftrightarrow\quad
            v^\T\btheta^*\ge 1-\frac{\delta^2}{2}.
        \]
        Taking $t=1-\delta^2/2$ and noting that
        \[
            1-t^2
            =
            \delta^2\left(1-\frac{\delta^2}{4}\right),
        \]
        we conclude that, with probability at least $1-\varepsilon$, there exists some
        $\bar v\in\{v_1,\ldots,v_m\}$ such that
        $\|\bar v-\btheta^*\|_2\le\delta$, provided that
        \[
            m \ge
            \frac{L-1}{c_L}
            \left[
                \delta^2\left(1-\frac{\delta^2}{4}\right)
            \right]^{-(L-1)/2}
            \log(1/\varepsilon).
        \]
        Finally, we have the fact that $c_L\asymp \sqrt{L}$.
    \end{proof}

    \begin{proof}[Proof of \cref{lem_rand_init}]
        Pick any $k\in [K]$ with  $\btheta_k^*\in \bS^{L-1}$.
        For any $i\in [m]$ with $v_i$ defined in \eqref{def_proj}, one has 
        \begin{align*}
            \|v_i - \btheta_k^*\|_2^2  &= 2 - 2 v_i^\T \btheta_k^*\\
            &= 2 -  {2u_i^\T \wh V \wh V^\T  \btheta_k^* \over \|\wh V^\T u_i\|_2}\\
            & = 2 -  {2u_i^\T \wh V \wh V^\T  \btheta_k^* \over \|\wh V^\T u_i\|_2\|\wh V^\T \btheta_k^*\|_2} +  {2u_i^\T \wh V \wh V^\T  \btheta_k^* \over \|\wh V^\T u_i\|_2\|\wh V^\T \btheta_k^*\|_2}(1 - \|\wh V^\T \btheta_k^*\|_2)\\
            &\le    \left\| {  \wh V^\T u_i\over \|\wh V^\T u_i\|_2} - {\wh V^\T \btheta_k^*\over \|\wh V^\T \btheta_k^*\|_2}\right\|_2^2  +2 \left(1 - \|\wh V^\T \btheta_k^*\|_2\right).
        \end{align*} 
       Using the arguments in the proof of \cref{prop_eigensp}, one has 
       $$
        1-   \|\wh V^\T \btheta_k^*\|_2^2 = \btheta_k^{*\T}(\bI_L-\wh V\wh V^\T)\btheta_k^* \le {2\over \ua}\|\wh \Gamma - \Gamma\|_\op \le {\delta_0^2\over 4}.
        $$  
       By applying \eqref{init_cap_bound} to   the first term  with $\delta^2 = \delta_0^2/2$ and $L=K$, the number of draws to $\delta_0$-cover $\btheta_k^*$, with probability at least $1-\varepsilon$, is 
       $$C\sqrt{K}\left[
                {\delta_0^2\over 2}\left(1-\frac{\delta_0^2}{8}\right)
            \right]^{-(K-1)/2}
            \log(1/\varepsilon).$$ 
        The claim follows by the union bound.
    \end{proof}


%
%

\subsection{Proof of \cref{prop:weak}}
\label{app-weak-limit-SMM}
For $x\in\RR^L$,  let
\[
F_p(x;\bomega)
:= \PP\{ Y_p\le x\}= 
 \sum_{k=1}^K
\alpha_k
\frac{p^{-1}\sum_{j=1}^p \exp( {X_j^\T\btheta_k})\mathbf 1\{X_j\le x\}}{p^{-1}\sum_{j=1}^p \exp({X_j^\T\btheta_k})},
\] 
where $X_j\le x$ denotes coordinate-wise inequality.
 For each $q\in\mathbb Q^L$ and $k\in[K]$, the strong law of large numbers gives
\[ \frac1p\sum_{j=1}^p \exp({X_j^\T\btheta_k})\mathbf 1\{X_j\le q\}
\longrightarrow
\int_{(-\infty,q]} \exp({x^\T\btheta_k})\, d\mu(x)
\]
and
\[
\frac1p\sum_{j=1}^p \exp({X_j^\T\btheta_k})\longrightarrow \exp (\psi (\btheta_k))>0
\]
almost surely, as $p\to\infty$. Since $K$ is fixed and $\mathbb Q^L$ is countable, these
convergences hold simultaneously for all $k\in[K]$ and
$q\in\mathbb Q^L$ on a single probability-one event. Hence, on that
event,
\[
F_p(q;\bomega)
\longrightarrow
\sum_{k=1}^K\alpha_k\int_{(-\infty,q]} \exp({x^\T\btheta_k} - \psi(\btheta_k) f_\mu(x)\,dx 
=: F(q;\bomega),
\qquad
q\in\mathbb Q^L
\]as $p\to \infty$.
 For arbitrary $F(\cdot;\bomega)$-continuity point $x\in\RR^L$, choose
$q_n^-,q_n^+\in\mathbb Q^L$ with
$q_n^-\uparrow x$ and $q_n^+\downarrow x$ coordinatewise. Monotonicity
gives
$
F_p(q_n^-;\bomega)
\le
F_p(x;\bomega)
\le
F_p(q_n^+;\bomega )
$.
Letting first $p\to\infty$ and then $n\to\infty$ yields
$F_p(x;\bomega) \longrightarrow F(x;\bomega)$ for every $x\in\RR^L$.
This establishes the weak convergence $Y_p\implies Y$.
\qed 


\subsection{Proof of \cref{prop:iden}}
\label{app-asym-iden}
We first show that the limiting mixture $P(\cdot; \bomega):= \sum_{k=1}^K \alpha_k \mu(\cdot;\btheta_k)$   is identifiable. 
Since
\begin{equation*} 
f(x;\btheta)
:=
\exp\!\left\{x^\top\btheta-\psi(\btheta)\right\}f_{\mu}(x).
\end{equation*}
is the tilted exponential density of $f_\mu$, the limit measure $P(\cdot; \bomega)
$ has density $\sum_{k=1}^K\alpha_kf(x;\btheta_k)$.
By \cite{YakowitzSpragins1968}, we need to show that the exponential tilt densities $f(x;\btheta_1),\ldots,f(x,\btheta_K)$ are linearly independent, whenever $\btheta_1,\ldots\btheta_K$ are distinct. Since 
\[ 
\sum_{k=1}^K c_kf(x;\btheta_k) =f_\mu(x) \sum_{k=1}^K c_k \exp\left\{x^\top \btheta_k-\psi(\btheta_k)\right\}
\]
and $f_\mu$ is continuous implies $f_\mu>0$ on some open set $U\ne \emptyset$,
it suffices to show that 
 the functions $\exp(x^\top  \btheta_1), \dots, \exp(x^ \top \btheta_K)$ are linearly independent on the open set $U$. Suppose
 \[\sum_{k=1}^K c_k \exp(x^\top \btheta_k)=0 \qquad\text{ for all } x\in U \]
 Let $x_0\in U$ and $v\in[0,1]^L$ be arbitrary. By assumption, the function 
 \[g(t)= \sum_{k=1}^K c_k\exp((x_0 + tv)^\top\btheta_k)\]
	vanishes identically (for $t$ small enough so that $x_0+tv\in U$). Differentiating $n$ times and evaluating at
	$t=0$ gives
	\[
		\sum_{k=1}^K c_k \exp( x_0^\T \btheta_k) (v^\top \btheta_k)^n=0,
		\qquad n\geq0.
	\]
	Taking $n=0,\dots,K-1$ and writing $\gamma:=\left(c_1\exp( x_0^\T \btheta_1),\ldots,c_K\exp( x_0^\T \btheta_K)\right)^\T$ yields
	\[
		\begin{pmatrix}
		1 & \cdots & 1\\
		v^\top\btheta_1 & \cdots & v^\top\btheta_K\\
		\vdots & & \vdots\\
		(v^\top\btheta_1)^{K-1} & \cdots &
		(v^\top\btheta_K)^{K-1}
		\end{pmatrix}
		\gamma=0.
	\]
	If $v \sim \text{Unif}([0,1]^L)$, 
    then $v ^\top \btheta_1, \dots, v ^\top\btheta_K$ are distinct with probability $1$, and the above matrix equation then implies that $\gamma = 0$ since the Vandermonde matrix is invertible. Since $\exp( x_0^\T \btheta_k)>0$ for $k\in [K]$, it follows that $c_1=\cdots=c_K=0$.
	This implies that the limiting measure is identifiable, up to label permutations \cite{YakowitzSpragins1968}.\\

Now we are in the position  to prove \cref{prop:iden}.
If $\bomega^*\sim\bomega^\dagger$, then the two parameter configurations
differ only by a relabeling of their mixture components and
$
\pi(\cdot;\bomega^*)
=
\pi(\cdot;\bomega^\dagger)
$
for every $p$ and every realization $X_{1},\ldots,X_{p}$.

For the converse, suppose that
$
\bomega^*\not\sim\bomega^\dagger.
$
By \cref{prop:weak}, for
$\mu^{\otimes\mathbb N}$-almost every sequence,
\[
\pi_p(\cdot; {\bomega^*}) \Rightarrow P(\cdot;{\omega^*}),
\qquad
\pi_p(\cdot;{\bomega^\dagger}) \Rightarrow P(\cdot;{\bomega^\dagger}),
\]
The identifiability for finite mixtures of exponential tilts implies
$P(\cdot;{\bomega^*})\ne P(\cdot;{\bomega^\dagger})$, because otherwise $\bomega^*$ and $\bomega^\dagger$ would agree up to
label permutation.
Let $d(P,Q)$  be any metric that  metrizes weak convergence on $\RR^L$. (For instance, take the bounded-Lipschitz metric.)  On the same probability-one event,
using  the triangle inequality, we find
\begin{align*}
&\left| d \left(\pi_p ( \cdot;{\bomega^*}),\pi_p(\cdot;{\bomega^\dagger}) \right)
- d (P(\cdot;{\bomega^*}),P(\cdot;{\bomega^\dagger}))
\right|\\
&\qquad
\le
d\left(\pi_p(\cdot;{\bomega^*}),P(\cdot;{\bomega^*}) \right)
+
d\left(\pi_p(\cdot;{\bomega^\dagger}),P(\cdot;{\bomega^\dagger}) \right)
\longrightarrow0
\end{align*} and we find
\[
d \left(\pi_p(\cdot;{\bomega^*}),\pi_p(\cdot;{\bomega^\dagger})\right)
\longrightarrow
d (P(\cdot;{\bomega^*}),P(\cdot;{\bomega^\dagger}) ) >0.
\]
It follows that, for almost every   sequence $\{ X_i\}_{i\ge1}$, there exists
$p_0<\infty$ such that
\[
\pi_p(\cdot;\bomega^*)
\ne
\pi_p(\cdot;\bomega^\dagger)
\qquad
\text{for every }p\ge p_0.
\]
Thus equality of the finite-$p$ softmax measures for all sufficiently
large $p$ is impossible when
$\bomega^*\not\sim\bomega^\dagger$, proving the reverse implication.
\qed 
 
\subsection{Proof of \cref{prop:iden2}} 
\label{app-new-asym-iden}
We provide an alternative proof of \cref{prop:iden} under a stronger set of assumptions. 
It allows us to quantify {\em for $p$ large enough}  in the proof, see the definition of the threshold $p_0$ in  \eqref{p_0}.
We first state the identifiability result under the new set of assumptions on $\mu$.

\begin{theorem} \label{prop:iden2}
Suppose that $\mu$ satisfies \eqref{ass:mu}.
Let
$\bomega^*=(\alpha^*,\btheta_1^*,\ldots,\btheta_K^*)$ and 
$\bomega^\dagger = (\alpha^\dagger,\btheta_1^\dagger,\ldots,\btheta_K^\dagger)
$ be two fixed parameter configurations satisfying Assumptions \ref{ass_alpha} and \ref{separation}.

Then,
for $\mu^{\otimes\mathbb N}$-almost every sequence $(X_i)_{i\ge1}$,
$$
\bomega^*\sim\bomega^\dagger \quad\Longleftrightarrow\quad \pi(\cdot;\bomega^*)
= \pi(\cdot;\bomega^\dagger) \quad
\text{for all sufficiently large }p.
$$
Here $\sim$ denotes equality up to label permutation.
\end{theorem}

\begin{proof}[Proof of \cref{prop:iden}]
The quantities in (\ref{req_h_hi}) are the expectations of
$h_r(X)$ $h_{r1;i}(X)$, defined in (\ref{eq:hr_def}) and (\ref{eq:hr1_def}), under the tilted density $f(x;\btheta):=
\exp\!\left\{x^\top\btheta-\psi(\btheta)\right\}\mu(x)$.
It follows that ${m}_r$ and $m_{r1:i}$ are
the expected values of $h_r(X)$ and $h_{r1;i}(X)$, respectively, under the
  mixture density $$
\sum_{k=1}^K\alpha_kf(x;\btheta_k)
=
\mu(x)\sum_{k=1}^K\alpha_k \exp\!\left\{x^\top\btheta_k-\psi(\btheta_k)\right\}.
$$
\cref{Lindsay} implies that this 
 distribution  is identifiable up to label
permutations.

Fix the pair
$(\bomega^*,\bomega^\dagger)$   in the proposition.  After a
single relabeling and a common orthogonal change of coordinates, we
may assume without loss of generality that the first coordinates of
the component parameters are pairwise distinct within each
configuration and that, if the two atom configurations differ, they
differ in the first coordinate.
Set
\[
\delta
:=
\max\left\{
\|\alpha^*-\alpha^\dagger\|_2,\,
\max_{k\in[K]}
|\btheta_{k1}^*-\btheta_{k1}^\dagger|
\right\}>0.
\]
Let ${\bm m}(\bomega)$
be the vector of moments defined in (\ref{mom}) and,   
${\bm m}_{1;i}(\bomega)$ be
the vector of  mixed-moments in (\ref{cross-mom}).  
Let $\bar{\bm m}_p(\bomega)$ and
$\bar{\bmvec}_{p,1;i}(\bomega)$ denote the corresponding 
moment vectors in \eqref{cond-mom}, with expectations under   $\pi(\cdot;\bomega)$. 
Apply the first-coordinate and mixing-weight bounds in the proof of
\cref{prop:theta_gap}, with $\bomega^*$ playing the role of the true parameter and with the exact
moment vector $m(\bomega^\dagger)$ in place of the approximating moment vector. By \cref{prop:theta_gap}, we find 
\begin{align}
\label{eq:ordinary-moment-gap}
\Delta_m
:=
\sqrt{L\left\|{\bm m}(\bomega^*)-{\bm m}(\bomega^\dagger)\right\|_2^2
+
\sum_{i=2}^L
\left\|\bmvec_{1;i}(\bomega^*)-\bmvec_{1;i}(\bomega^\dagger)\right\|_2^2 }
 \ge\frac{\delta}{D}.
\end{align}
where  $D>1$ is the constant  in \cref{prop:theta_gap}. 
Define
\[
\varepsilon_p^2(\bomega)
:=
L\left\|\bar{\bm m}_p(\bomega)-{\bm m}(\bomega)\right\|_2^2
+
\sum_{i=2}^L
\left\|\bar{\bmvec}_{p,1;i}(\bomega)-\bmvec_{1;i}(\bomega)\right\|_2^2
\]
By the strong law of large numbers, there exists a set $\mathcal A\subset(\mathbb R^L)^{\mathbb N}$ with
$\mu^{\otimes\mathbb N}(\mathcal A)=1$ such that, for every
$(x_i)_{i\ge1}\in\mathcal A$, one may take
\begin{align} \label{p_0}
p_0=\min\left\{q\ge1: \sup_{p\ge q}
\left[ \varepsilon_p(\bomega^*) +\varepsilon_p(\bomega^\dagger )
\right] < \frac{\delta}{2D} \right\}.
\end{align}
This $p_0$ is finite under \cref{ass:mu}, and the asserted equivalence holds for every
$p\ge p_0$.
Indeed, 
suppose that, for  $p\ge p_0$,
$\pi_p(\cdot;\bomega^*)=\pi_p(\cdot;\bomega^\dagger)$. Hence
$\bar m_p(\bomega^*)=\bar m_p(\bomega^\dagger)$. 
But the triangle inequality yields
$\Delta_m \le \varepsilon_p(\bomega^*) + \varepsilon_p(\bomega^\dagger) 
<
{\delta}/(2D)$, which 
contradicts ~\eqref{eq:ordinary-moment-gap}.  Thus
$\pi_p(\cdot;\bomega^*)
\ne
\pi_p(\cdot;\bomega^\dagger)$ 
for all $p\ge p_0$.
\end{proof}

We observe that  $p_0$ depends both on the realized support
sequence $x_1,x_2,\ldots$ and on the fixed parameter pair $\bomega^*$ and $\bomega^\dagger$.  
The latter dependence
enters through  $D$.  In particular, if
$\Delta_1$ denotes the minimum separation of the first coordinates,
the proof of \cref{prop:theta_gap} shows that $D$ deteriorates polynomially
in $\Delta_1^{-1}$ with exponent proportional to $K$; the
first-coordinate reconstruction bound contains explicitly the factor
$\Delta_1^{-(2K-2)}$.

	\section{Auxiliary lemmas}\label{app_auxiliary}

    The following lemmas contain some basic results on moments related with (sub-)Gaussian random variables.
	
	\begin{lemma}\label{lem_gauss}
		Let $Z \sim N(0, \sigma^2)$. Then for any $t\in \RR$, 
		\[
		\EE\left[Z e^{Zt}\right ] = \sigma^2 t e^{\sigma^2 t^2/2},\qquad 
		\EE\left[Z^2 e^{Zt}\right] = \sigma^2 \left(1 + \sigma^2 t^2\right) e^{\sigma^2 t^2 /2}
		\]
	\end{lemma}
	\begin{proof}
		The proof follows from the Gaussian density and integration by parts. 
	\end{proof}
	
	\begin{lemma}\label{lem_moments}
		Let $ Z \sim \cN_L(0, \sigma^2 \bI_L)$. For any vectors $u, \btheta \in \RR^L$, we have 
		\begin{align*}
			& \EE\left[
			(Z^\T u)  e^{Z^\T \btheta}
			\right] =  \sigma^2 (u^\T \btheta) e^{\sigma^2 \|\btheta\|_2^2 /2}\\
			& \EE\left[
			(Z^\T u)^2  e^{Z^\T \btheta}
			\right] =  \sigma^2 \left(
			\|u\|_2^2 + \sigma^2 (u^\T \btheta)^2
			\right) e^{\sigma^2 \|\btheta\|_2^2 /2}.
		\end{align*}
	\end{lemma}
	\begin{proof}
		To prove the first claim, let $Q$ be an $L\times L$ orthogonal matrix such that 
		\begin{equation}\label{def_Q_mat}
			Q u = \|u\|_2 ~ \be_1 . 
		\end{equation}
		Write $\bar\btheta = Q\btheta$ with $\bar \btheta = (\bar \btheta_1, \bar \btheta_{-1}^\T)^\T$, and similarly $Z = (Z_1, Z_{-1}^\T)^\T$.
		By the rotational invariance of spherical Gaussian, we have 
		\begin{align*}
			\EE\left[
			(Z^\T u)  e^{Z^\T \btheta}
			\right]  &= \|u\|_2 \EE\left[
			(Z^\T \be_1)  e^{Z^\T \bar\btheta }
			\right]  \\
			&= \|u\|_2 \EE\left[ Z_1 e^{Z_1 \bar{\btheta}_1}
			\right] \EE \left[
			e^{Z_{-1}^\T \bar \btheta_{-1}}
			\right] &&\text{by independence between $Z_1$ and $Z_{-1}$}\\
			&= \|u\|_2 \sigma^2 \bar \btheta_1 e^{\sigma^2 \bar \btheta_1^2 /2} e^{\sigma^2 \|\bar \btheta_{-1}\|_2^2 /2} &&\text{by \cref{lem_gauss}}\\
			&= \|u\|_2 \sigma^2 \bar \btheta_1 e^{\sigma^2 \|\btheta\|_2^2 /2}.
		\end{align*}
		The claim follows by noting that $\bar \btheta_1 = \btheta^\T Q^\T \be_1 =  \btheta^\T u / \|u\|_2$ from \eqref{def_Q_mat}.
		
		Regarding the second claim, by similar arguments, we have 
		\begin{align*}
			\EE\left[
			(Z^\T u)^2  e^{Z^\T \btheta}
			\right]  &= \|u\|_2^2 ~  \EE\left[
			(Z^\T \be_1)^2  e^{Z^\T \bar\btheta}
			\right] \\
			&=  \|u\|_2^2 ~ \EE\left[
			Z_1^2  e^{Z_1  \bar\btheta_1}
			\right] \EE \left[
			e^{Z_{-1}^\T \bar \btheta_{-1}}
			\right] \\
			&=  \sigma^2 \left(
			\|u\|_2^2  + \|u\|_2^2  \sigma^2 \bar\btheta_1^2
			\right) e^{\sigma^2 \|\btheta\|_2^2 /2} &&\text{by \cref{lem_gauss}},
		\end{align*}
		completing the proof.
	\end{proof}
	
	\begin{lemma}\label{lem_moment_bds}
		Let $Z \in \RR^L$ be a zero-mean, sub-Gaussian random vector with sub-Gaussian constant $\sigma^2$.  Then for any $u \in \bS^{L-1}$ and $\btheta \in \RR^L$, one has
		\[
		\EE\left[
		(Z^\T u)^4 e^{2 Z^\T \btheta}
		\right] \lesssim
		\sigma^4 e^{4\sigma^2\|\btheta\|_2^2}.
		\]
	\end{lemma}
	\begin{proof}
		The proof follows by the Cauchy-Schwarz inequality and the sub-Gaussianity of $Z$. 
	\end{proof}

  The following lemma states upper bounds of the quadratic form of a sub-Gaussian random vector \citep{Hsu2012}. 
	
	\begin{lemma} \label{lem_quad}
		Let $\xi\in \RR^d$ be a  sub-Gaussian random vector with parameter $\gamma_\xi$. Then, for all symmetric positive semi-definite matrices $H$, and all $t\ge 0$, 
		\[
		\PP\left\{
		\xi^\T H\xi > \gamma_\xi^2\left(
		\sqrt{{\rm tr}(H)}+ \sqrt{2 t \|H\|_{\rm op} }
		\right)^2
		\right\} \le e^{-t}.
		\] 
	\end{lemma}



  The next lemma states an anti-concentration inequality of $v^\T \btheta$ for any $v$ uniformly drawn from $\bS^{d-1}$. See, for instance, the proof of Lemma 3.1 in \cite{doss2023optimal}.

  \begin{lemma}\label{lem_unif_sphere}
    Let $\btheta \in \RR^d$ be any fixed vector. For  any $v$ uniformly drawn from $\bS^{d-1}$, one has that, for all $t\ge 0$,
    \[
        \PP\left\{
            |v^\T \btheta| < t \|\btheta\|_2
        \right\} < t\sqrt{d}.
    \]
  \end{lemma}

\section{Additional simulation studies}\label{app_sec_sim}

We report some additional simulation results in this section. \cref{app_sec_sim_plots} reports the estimation error $\text{Err}_\balpha$ of all methods in \cref{sec_sims}. \cref{app_sec_sim_rate} examines the algorithmic convergence of the EM algorithm and the effect of the stepsize. \cref{app_sec_sim_parametric} verifies the parametric rate of MoM. 

\subsection{Results of estimating $\alpha^*$ in the setting of \cref{sec_sims}}\label{app_sec_sim_plots}

\begin{figure}[ht]
    \centering
    \includegraphics[width=\linewidth]{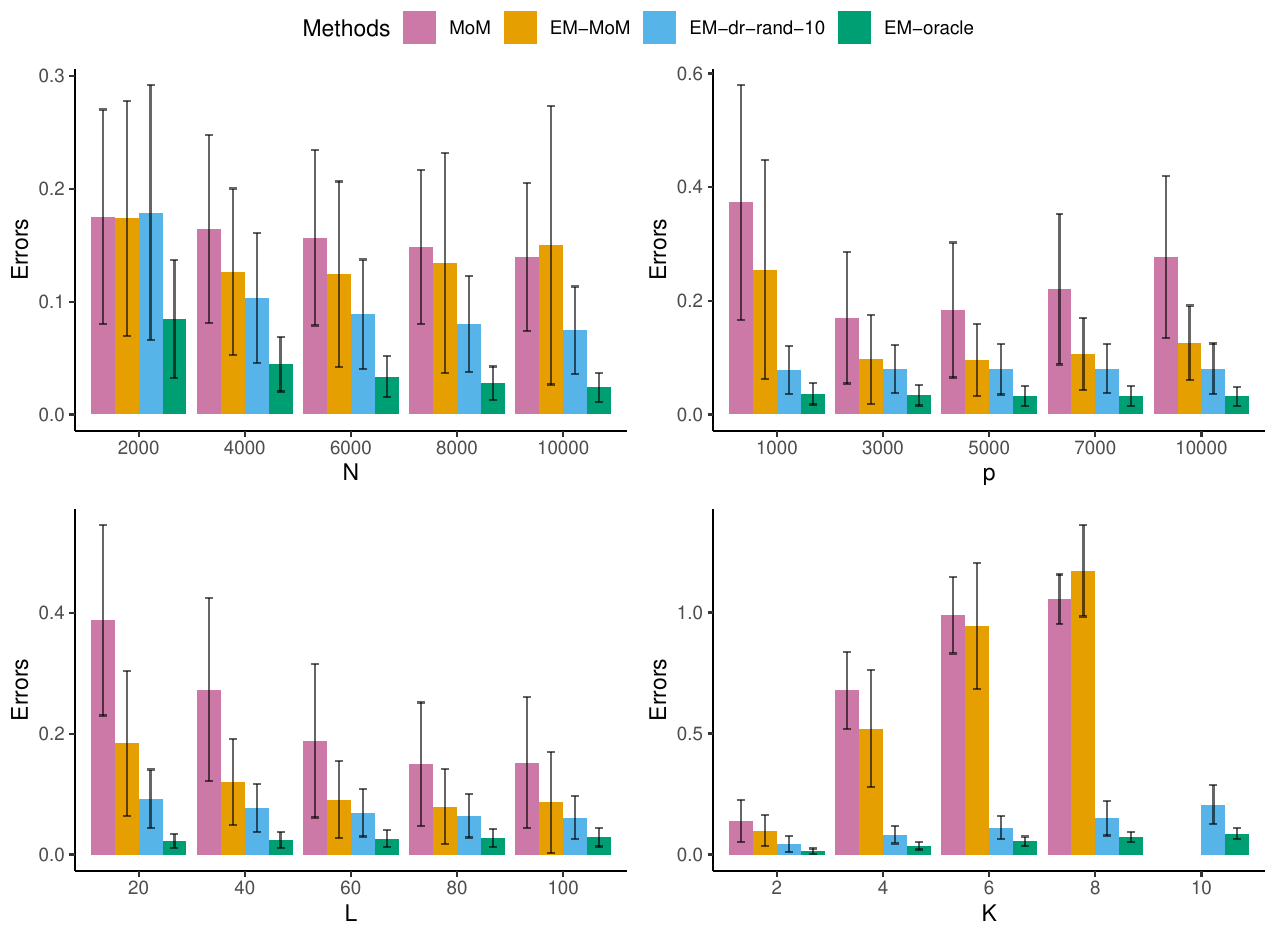}
    \caption{The averaged $\text{Err}_{\balpha}$ in different settings}
    \label{fig_weights}
\end{figure}

\cref{fig_weights} shows the errors $\text{Err}_\balpha$ of all methods in \cref{sec_sims}.
For estimating the mixing weights $\balpha^*$, both MoM and EM-MoM exhibit greater fluctuations in their errors due to the method's sensitivity to the choice of random projection in \cref{sec_mm_proj}. For large $K$, the errors in estimating $\balpha^*$ are substantially larger for these methods compared to other EM estimators, and are more sensitive than the corresponding errors in estimating $\btheta_k^*$. EM-dr-rand-10 and EM-oracle perform better for larger $N$ and smaller $K$, whereas their performance remains similar as $p$ and $L$ vary.

\subsection{Algorithmic convergence rate of the EM algorithm and the effect of step size}\label{app_sec_sim_rate}

\begin{figure}[ht]
        \centering
        \includegraphics[width=.45\linewidth]{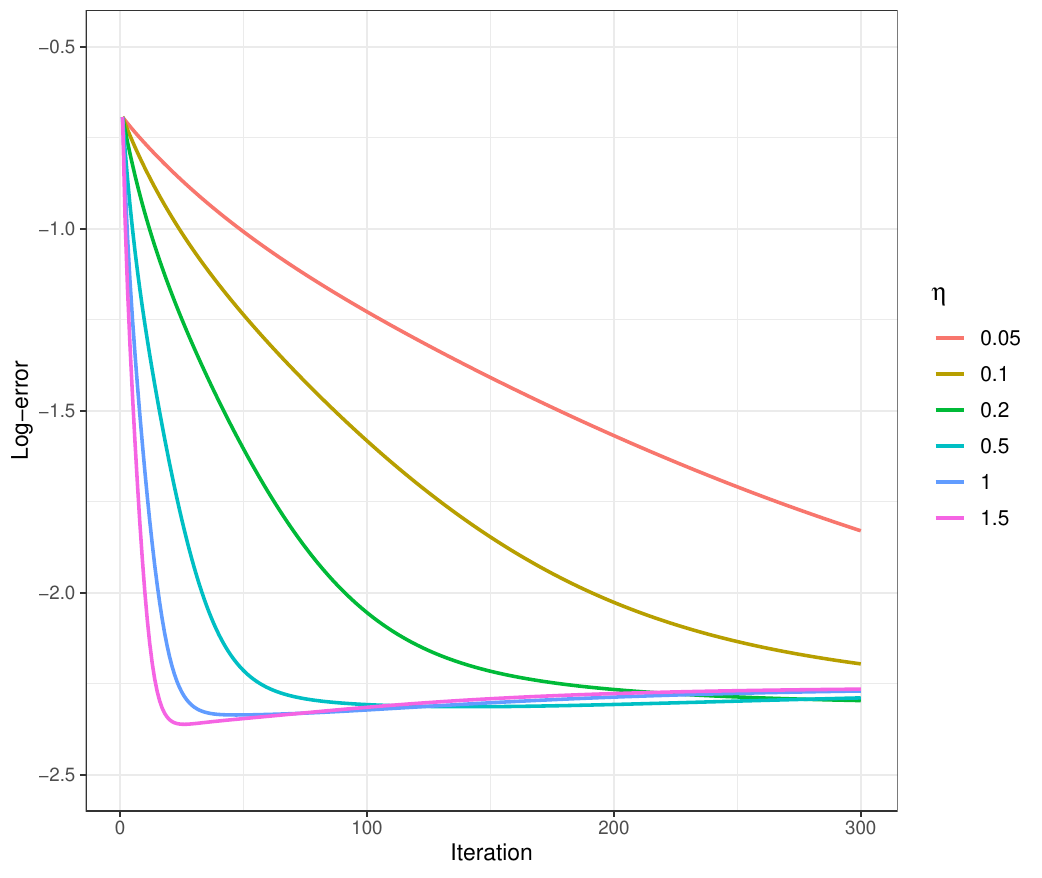}\hspace{5mm}\includegraphics[width=.45\linewidth]{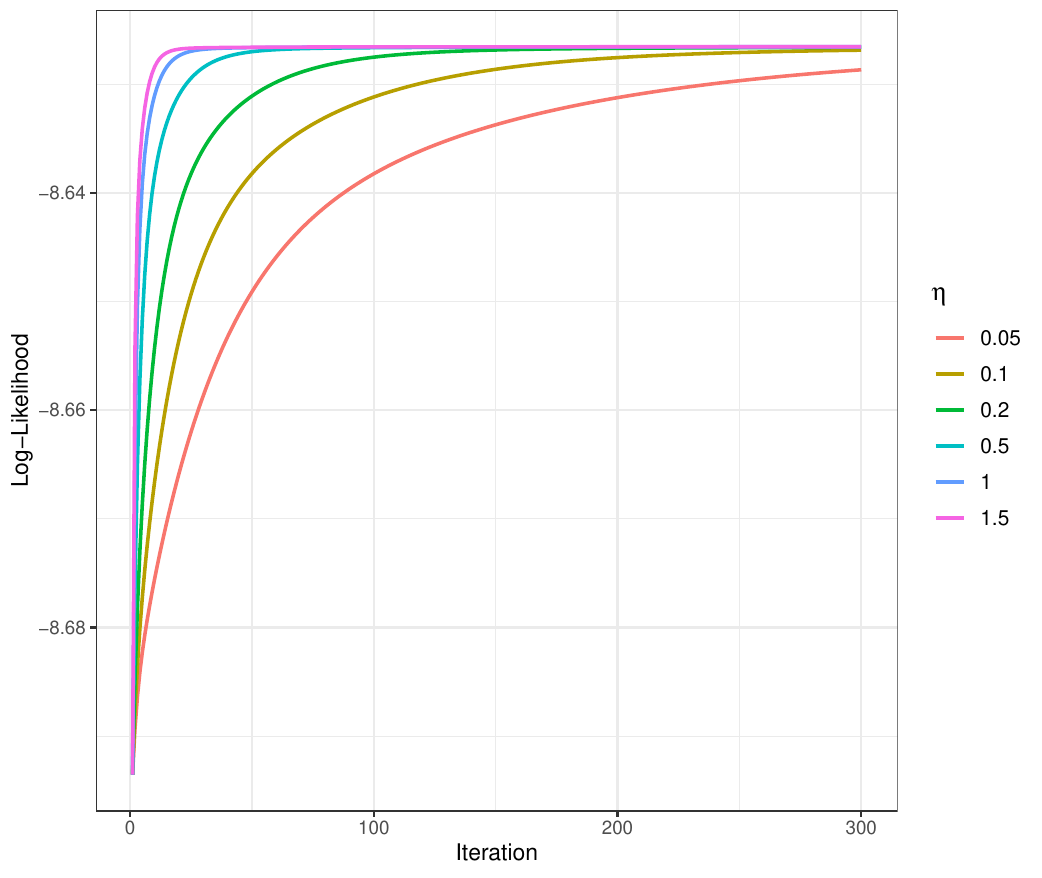}
        \caption{$\log(\text{Err}_{\btheta})$ and log-likelihood per iteration for EM-$\eta$.}
        \label{fig_conv_rate}
    \end{figure}

In this section we verify the algorithmic convergence of the EM algorithm as well as the effect of the step size $\eta_k$. We consider the data generating mechanism in \cref{sec_sims} with $N = 10000$, $p=7000$, $K = 3$ and $L = 10$. The EM algorithm is initialized with $\balpha^{(0)} =  1/K$ and $\btheta_k^{(0)} = \btheta_k^* + (0.5) v_k$ with $v_1,\ldots,v_K$ being uniformly sampled from the group of matrices with orthonormal columns. We consider EM-$\eta$ with $\eta_k := \eta\in \{0.05, 0.1, 0.2, 0.5, 1, 1.5\}$ denoting the stepsize. \cref{fig_conv_rate} shows the averaged log-transformed $\text{Err}_{\btheta}$ as well as the log-likelihood over the first 300 iterations.
Clearly, the smaller $\eta$ is, the slower the algorithm converges. For relatively large $\eta$, the log-error decreases approximately linearly during the initial iterations, indicating geometric decay of the estimation error until it reaches a neighbourhood of the ground truth. This aligns with the linear convergence rate in \cref{thm_EM_fix}. The log-likelihood exhibits a similar dependence on the stepsize, with larger stepsizes leading to substantially faster convergence to essentially the same limiting value. When the stepsize is too large, we also observe a slight increase in the estimation error after it reaches the desired precision, despite the log-likelihood remaining stable.

\subsection{Parametric rate of MoM}\label{app_sec_sim_parametric}

\begin{figure}[ht]
    \centering
    \includegraphics[width=.9\linewidth]{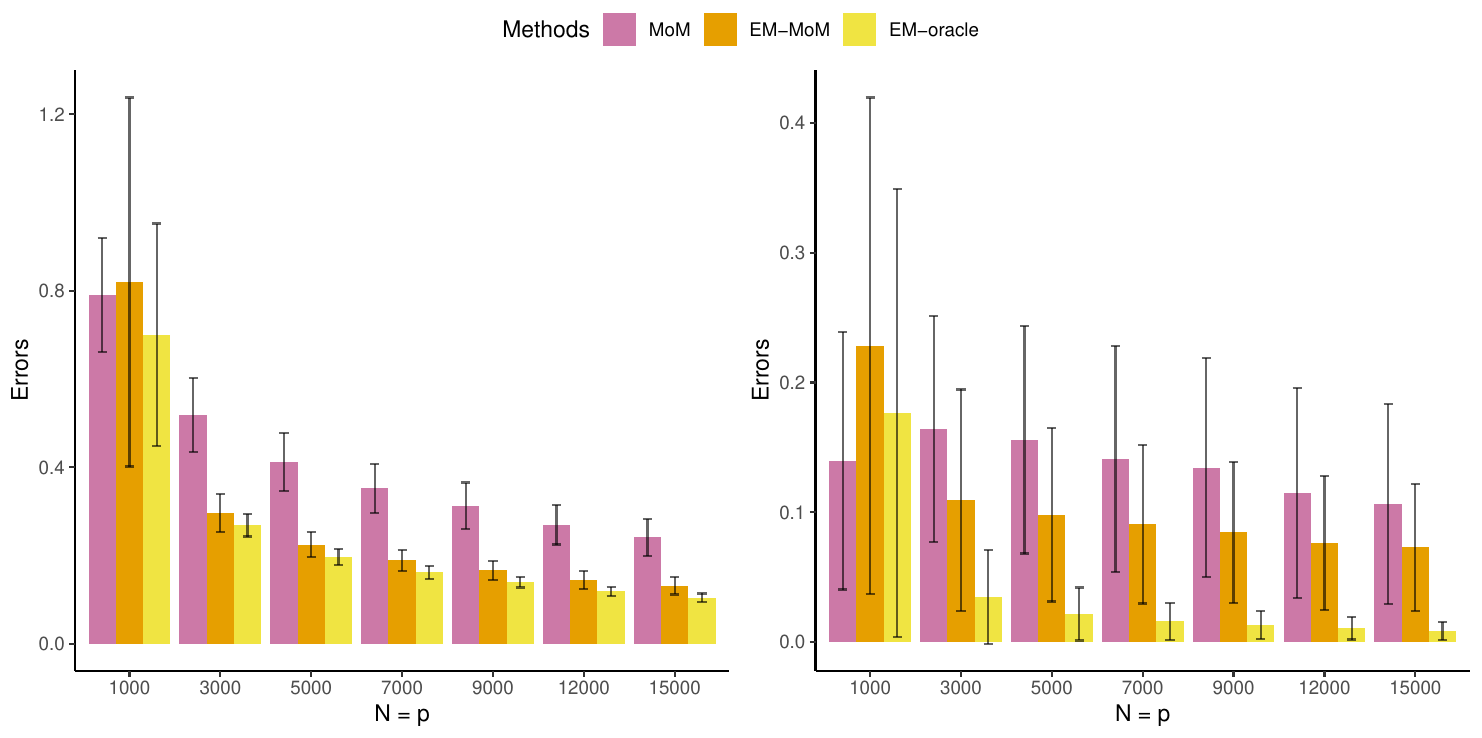}
    \vspace{0mm}
    \caption{$K = 2$: The averaged $\text{Err}_{\btheta}$ (left)  and $\text{Err}_{\balpha}$ (right) in different settings}
    \label{fig_param}
\end{figure}

We conduct a separate simulation study below to verify that MoM can indeed enjoy a parametric rate,  by taking  $K =2$ and ensuring that  the atoms have the theoretically prescribed separation. We let $L = 50$ and vary $p = N \in \{1, 3, 5, 7, 9, 12, 15\}\times 10^3$. \cref{fig_param} depicts the estimation errors of MoM, EM-MoM and EM-oracle. We observe the same phenomenon as above except that $\text{Err}_{\btheta}$ of MoM decays in the faster parametric rate as $N$ and $p$ increase. The large fluctuation of $\text{Err}_{\balpha}$ for MoM  can be explained by the sensitivity of the method to the selection of the random projection in \cref{sec_mm_proj}.  

\section{Real data analysis}\label{sec_real_data}

We illustrate the proposed method using the Dunnhumby Complete Journey grocery transaction data.\footnote{The data is available at \href{https://www.dunnhumby.com/source-files/}{https://www.dunnhumby.com/source-files/}.} We focus on the soft-drink category, which contains a large and heterogeneous collection of products with substantial variation in package format, manufacturer, price, and promotional activity. We merge the transaction records with the product information using the product identifier and retain transactions classified under the `SOFT DRINKS’ commodity. Transactions with nonpositive purchase quantity or sales value are removed. We divide the observation horizon into three periods: weeks 1–52 are used exclusively to construct product-level covariates, weeks 53–80 are used for model estimation, and weeks 81–102 are reserved for out-of-sample evaluation. We retain products that have at least 30 purchase records during weeks 1–52 and appear during the estimation period. Products whose subcommodity descriptions correspond to juice, tea, energy drinks, or coffee are also excluded to maintain a relatively homogeneous product category and prevent the latent components from primarily distinguishing fundamentally different types of beverages. The resulting choice set contains $p=305$ products.

For each product, we construct covariates using only information from weeks 1--52. The continuous features include the mean and standard deviation of unit price and purchase quantity; the mean magnitude and frequency of retail and coupon discounts; and quarterly purchase shares capturing historical seasonality, where the share for each quarter is defined as the fraction of a product’s purchase records during weeks 1–52 occurring in that quarter. Since the four quarterly shares sum to one for each product, the fourth-quarter share is omitted as the reference category. We additionally extract package characteristics from the product descriptions, including pack count and total package volume in ounces. Missing package-size information is imputed using medians first within the same subcommodity and pack-count group, then within the same subcommodity, and finally across all retained products. Manufacturer and subcommodity are included as categorical product characteristics through dummy variables, with 15 manufacturer indicators and 12 subcommodity indicators, resulting in \(L=40\) features in total. All continuous covariates are standardized before estimation.

To construct the response, for each product \(j\), we define \(Y_j\) as the number of purchase records involving product \(j\) during the estimation period. Thus, \(Y_j\) measures how frequently product \(j\) appears in purchases rather than the total number of physical units sold: purchasing one or multiple units of the same product in a basket contributes one count to \(Y_j\). Purchase quantity is instead incorporated into the product covariates through its historical mean and standard deviation, allowing quantity-related purchasing patterns to inform product characteristics without weighting the response by the number of units purchased. We end up with $N= 30154$ for the estimation period and $N_{\rm test} = 22626$ for the validation period.

We estimate the mixture model for a range of values of \(K \in \{1,2,\ldots, 10\}\) using weeks 53--80 and evaluate each fitted model on the held-out observations from weeks 81--102 using the average held-out log-likelihood
$ 
\sum_{j=1}^{p}
(Y_{j,\mathrm{test}}/N_{\mathrm{test}})\log\widehat{\pi}_j$
where \(\widehat{\pi}_j = \sum_{k=1}^K \wh \alpha_k A_{\wh \btheta_k}(x_j)\) denotes the fitted probability of product \(j\) using \(K\) softmax components. Due to the existence of categorical features, we fit the model by using the EM algorithm with 1000 initializations. The held-out criterion is optimized at \(K=8\). However, two of the eight estimated mixing weights are zero, indicating that the fitted model effectively uses only six components. A similar phenomenon occurs for \(K=9\), for which three estimated mixing weights are zero. Moreover, since the held-out average log-likelihood for \(K=6\) is close to that obtained with \(K=8\) and $K=9$, we choose \(K=6\) in the end for a more parsimonious representation of the latent structure. 

For \(K=6\), the six fitted components $\wh\btheta_1,\ldots,\wh\btheta_6$, with their estimated mixing weights $\wh \balpha = 
(0.171,0.140,0.130,0.224,0.229,0.106)$, reveal heterogeneity along several interpretable dimensions, most notably package format, manufacturer, promotional activity, and historical price and purchase variability.

Component 1, with weight \(0.171\), represents a clear 2-liter bottled-drink profile. The 2-liter carbonated-bottle indicator has the largest positive product-type coefficient, and the highest-probability products under this component are overwhelmingly 2-liter bottles. The component is also positively associated with the average magnitude of retail and coupon discounts and with average purchase quantity, but strongly negatively associated with coupon frequency, suggesting a profile associated with relatively large rather than frequent coupon discounts. Component 6, with weight \(0.106\), exhibits a different package-format profile, placing relatively high probabilities on 4/6-pack canned beverages and positive weight on pack count and total package volume.

Components 2 and 3 demonstrate that package size alone does not explain the estimated heterogeneity. Component 2, with weight \(0.140\), favors products with larger total volume and pack count, particularly multipack and single-serve bottled beverages. It also loads positively on price and purchase-quantity variability and coupon frequency, while average price, average purchase quantity, and retail-discount frequency enter negatively. Component 3, with weight \(0.130\), is instead sharply concentrated on 12/18/15-pack cans. Its largest positive continuous coefficients are the standard deviations of purchase quantity and price, whereas 2-liter bottles, total package volume, 20/24-pack cans, and average purchase quantity receive large negative coefficients. Retail-discount magnitude enters positively while retail-discount frequency enters negatively, suggesting relatively substantial but less frequent discounts. The strong positive coefficients on historical price and purchase variability may also reflect episodic demand or promotional cycles.

The two largest components (4 and 5) exhibit particularly strong manufacturer and promotion effects. Component 4, with weight \(0.224\), has a large positive coefficient for Manufacturer 69 and favors single-serve carbonated bottles. Retail-discount frequency is strongly positive, whereas average retail discount, total volume, pack count, and average purchase quantity are negative, suggesting a smaller-format profile associated with relatively frequent retail promotions. In contrast, Component 5, with weight \(0.229\), is highly concentrated on Manufacturer 1208 products, particularly 12/18/15-pack cans and 2-liter bottles. Its two highest-probability products account for approximately \(16\%\) and \(13\%\) of the component-specific probability, respectively. Price and purchase-quantity variability and the 2-liter indicator have large positive coefficients, whereas 20/24-pack cans, pack count, average purchase quantity, and retail-discount frequency have large negative coefficients. Component 5 therefore represents a relatively concentrated, manufacturer-specific demand profile.

A notable feature of the estimates is the reversal of several covariate effects across components. Retail-discount frequency, for example, is strongly positive for Component 4 but negative for Components 2, 3, and 5, while coupon frequency is strongly negative for Component 1 but positive for Components 2 and 5. Manufacturer effects also vary substantially: Manufacturer 1208 is strongly favored by Component 5 but disfavored by other components, while Manufacturer 69 is favored by Component 4 but disfavored by Component 3. These reversals indicate that the relationships between product demand and package format, manufacturer, and promotional activity vary substantially across the latent components.

These findings have several potential retailing implications. In particular, the distinction between promotion frequency and magnitude suggests that promotional activity may be better characterized along both dimensions rather than through a single measure of discounting. The pronounced differentiation among single-serve bottles, 2-liter bottles, and different multipack formats also suggests that package architecture represents an important dimension of demand, while the component-specific manufacturer effects indicate that aggregate manufacturer-level sales may conceal substantial heterogeneity across package formats and promotional regimes. These patterns may therefore be relevant for product assortment, promotion, and inventory-planning decisions.

Overall, the estimated mixture reveals economically interpretable heterogeneity in aggregate soft-drink demand along package format, manufacturer, promotional activity, and historical demand variability. In particular, the cross-component reversals illustrate relationships that would be obscured under a homogeneous softmax specification.

\end{document}

%% file: jnwcommands.tex
\usepackage{amsmath,amssymb,mathtools}
\usepackage{xifthen}
\usepackage{dsfont}

\newcommand{\cB}{\mathcal{B}}

\newcommand{\cE}{\mathcal{E}}

\newcommand{\cM}{\mathcal{M}}
\newcommand{\cN}{\mathcal{N}}
\newcommand{\cO}{\mathcal{O}}

\newcommand{\cX}{\mathcal{X}}

\newcommand{\EE}{\mathbb{E}}

\newcommand{\NN}{\mathbb{N}}

\newcommand{\PP}{\mathbb{P}}

\newcommand{\RR}{\mathbb{R}}

\newcommand*{\kl}[3][]{%
\ifthenelse{\isempty{#1}}{\operatorname{D}(#2\,\|\,#3)}%
{\operatorname{D}(#2\,\|\,#3\mid#1)}%
}

\DeclarePairedDelimiter{\triplenorm}{\vert\kern-0.25ex\vert\kern-0.25ex\vert}{\vert\kern-0.25ex\vert\kern-0.25ex\vert}

\newcommand*{\p}[1]{\mathbb P\left\{#1\right\}}

\DeclareMathOperator{\var}{Var}

\newcommand*{\rd}{\mathrm{d}}
\newcommand*{\dd}{\, \rd}
\DeclareMathOperator*{\argmin}{argmin}
\DeclareMathOperator*{\argmax}{argmax}

\newcommand*{\todo}[1][]{%
\ifthenelse{\equal{#1}{}}{\textcolor{red}{[\textbf{TODO}]}}%
{\textcolor{red}{[\textbf{TODO:} #1]}}%
}